\documentclass[a4paper,fleqn]{cas-dc}

\usepackage[numbers]{natbib}
\usepackage{lscape}
\usepackage{longtable}
\usepackage{caption}
\usepackage{subcaption}
\usepackage{xcolor}
\usepackage{soul}
\usepackage{color}

\usepackage[export]{adjustbox}
\newcolumntype{X}[1]{>{\raggedright\arraybackslash}m{#1}}
\newcolumntype{Y}[1]{>{\centering\arraybackslash}m{#1}}
\newcolumntype{Z}[1]{>{\raggedleft\arraybackslash}m{#1}}
\graphicspath{{./pics/}}

\begin{document}
\let\WriteBookmarks\relax
\def\floatpagepagefraction{1}
\def\textpagefraction{.001}
\shorttitle{Medical Visual Question Answering: A Survey}
\shortauthors{Lin et~al.}

\title[mode = title]{Medical Visual Question Answering: A Survey}                      



\author[1]{Zhihong Lin}[orcid=0000-0001-8499-4097]
\ead{zhihong.lin@monash.edu}
\author[2]{Donghao Zhang}
\ead{donghao.zhang@monash.edu}

\author[3]{Qingyi Tao}
\ead{qtao002@e.ntu.edu.sg}

\author[4]{Danli Shi}
\ead{shidli@mail2.sysu.edu.cn}

\author[5]{Gholamreza Haffari}
\ead{gholamreza.haffari@monash.edu}

\author[6]{Qi Wu}
\ead{qi.wu01@adelaide.edu.au}

\author[7]{Mingguang He}
\ead{mingguang.he@unimelb.edu.au}

\author[2,8,9]{Zongyuan Ge}
\cormark[1]
\ead{zongyuan.ge@monash.edu}

\address[1]{Faculty of Engineering, Monash University, Clayton, VIC, 3800 Australia}
\address[2]{eResearch Center, Monash University, Clayton, VIC, 3800 Australia}
\address[3]{NVIDIA AI Technology Center, 038988, Singapore}
\address[4]{State Key Laboratory of Ophthalmology, Zhongshan Ophthalmic Center, Sun Yat-Sen University, Guangzhou, 510060 China}
\address[5]{Faculty of Information Technology, Monash University, Clayton, 3800, VIC, Australia}
\address[6]{Australian Centre for Robotic Vision, The University of Adelaide, Adelaide, SA 5005, Australia}
\address[7]{Eye Research Australia, Royal Victorian Eye and Ear Hospital, East Melbourne, VIC, 3002 Australia}
\address[8]{Airdoc Research, Melbourne, VIC, 3000 Australia}
\address[9]{Monash-NVIDIA AI Tech Centre, Melbourne, VIC, 3000 Australia}

\cortext[cor1]{Corresponding author}


\begin{abstract}
Medical Visual Question Answering~(VQA) is a combination of medical artificial intelligence and popular VQA challenges. Given a medical image and a clinically relevant question in natural language, the medical VQA system is expected to predict a plausible and convincing answer. Although the general-domain VQA has been extensively studied, the medical VQA still needs specific investigation and exploration due to its task features. 
In the first part of this survey, we collect and discuss the publicly available medical VQA datasets up-to-date about the data source, data quantity, and task feature. In the second part, we review the approaches used in medical VQA tasks. We summarize and discuss their techniques, innovations, and potential improvements. In the last part, we analyze some medical-specific challenges for the field and discuss future research directions. Our goal is to provide comprehensive and helpful information for researchers interested in the medical visual question answering field and encourage them to conduct further research in this field.
\end{abstract}

\begin{highlights}

\item It is the first medical VQA survey paper as the current increasing application demand on medical VQA systems. It describes the history of medical VQA and research directions in the future. 

\item This survey presents an overview of the publicly available medical VQA datasets. The tasks include previous ImageCLEF VQA-Med challenges and other published datasets. This will help researchers to identify suitable research benchmarks and metrics. 

\item This survey gives a comprehensive summary and discussion of the published method papers for the medical VQA. It provides the comparison and discussion for the competition work notes and research papers. These will help researchers to better understand model design principles in common and conduct further research. 

\item It concludes some current challenges and future research directions. The common core of these challenges is the final application in the clinical scenario. The research topics cover dataset design to human-computer interaction. 
    
\end{highlights}


\ExplSyntaxOn
\keys_set:nn { stm / mktitle } { nologo }
\ExplSyntaxOff

\maketitle

\section{Introduction}\label{sec:intro}
Visual Questing Answering~(VQA)~\cite{antol2015vqa} is a multidisciplinary problem that incorporates computer vision~(CV) and natural language processing~(NLP). The VQA system is expected to answer an image-related question according to the image content. Inspired by the VQA research in the general domain, the recent exploration of medical VQA has attracted great interest. The medical VQA system is expected to assist in clinical decision-making and improve patient engagement~\cite{hasan2018overview,kovaleva2020towards}. Unlike other medical AI applications often restricted to pre-defined diseases or organ types, the medical VQA can understand free-form questions in natural language and provide reliable and user-friendly answers.

In recent research, the medical VQA has been assigned to several ``jobs''. The first one is the diagnostic radiologist, who acts as an expert consultant to the referring physician. A workload study~\cite{mcdonald2015effects} shows that the average radiologist has to interpret one CT or MRI image in 3 to 4 seconds. Besides the long queue of imaging studies, a radiologist must also answer an average of 27 phone calls per day from physicians and patients~\cite{cross2020voice}, leading to further inefficiencies and disruptions in the workflow. A medical VQA system can potentially answer the physician's questions and help relieve the burden of the healthcare system and improve medical professionals' efficiency. Another application matching the advantage of VQA is to act as the pathologists who examine body tissues and help other healthcare providers make diagnoses~\cite{he2020pathvqa}. 

In addition to the health professional role, the medical VQA system can also serve as a knowledgeable assistant. For example, the ``second opinion'' from the VQA system can support the clinicians' opinion in interpreting medical images and decrease the risk of misdiagnosis at the same time~\cite{tschandl2020human}.

Ultimately, a mature and complete medical VQA system can directly review patients' images and answer any kind of questions. In some situations, such as fully automated health examinations, where medical professionals may not be available, a VQA system can provide equivalent consultation. After a hospital visit, patients search for further information online. The irregular and misleading information from the search engine might result in inappropriate answers. Alternatively, a medical VQA can be integrated into an online consultation system to provide reliable answers anytime and anywhere.

Medical VQA is technically more challenging than general-domain VQA because of the following factors. Firstly, creating a large-scale medical VQA dataset is challenging because expert annotation is expensive for its high requirement of professional knowledge, and QA pairs can not be synthetically generated directly from images. Secondly, answering questions according to a medical image also demands a specific design of the VQA model. The task also needs to focus on a fine-grained scale because a lesion is microscopic. Hence, segmentation techniques may be required to locate the region of interest precisely. Finally, a question can be very professional, which requires the model to be trained with medical knowledge base rather than a general language database.

Since the first medical VQA challenge was organized in 2018~\cite{hasan2018overview}, an increasing number of organizations and researchers have joined to expand the tasks and propose new datasets and approaches, which have made the medical VQA task an active and inspiring field. To provide a comprehensive retrospect of these efforts, we conduct the first survey (to our best knowledge) for medical VQA. 

In the first part of this survey, we overview the publicly available medical VQA datasets up-to-date.
To collect the most complete information, we did an exhaustive search for the available medical VQA datasets including sources from relevant papers in google scholar, medical image computing conferences, and top-tier journals, and resulted in a total of 10 papers proposing datasets. Two dataset papers are repetitive and different versions of included datasets, and consequently, 8 datasets are analyzed and discussed in this survey. Among them, three datasets are proposed as ImageCLEF\footnote{\underline{https://www.imageclef.org/}} competitions. The selected datasets are diverse in image modality and question categories. The imaging modality of those datasets covers chest X-ray, CT, MRI, and pathology. The questions include close-end questions (such as Yes/No questions) and open-end questions on a variety of topics. We also compare the data sources, the question-answer pairs creation methods, and the metrics for evaluation. These will be inspiring for researchers who are interested in designing new tasks. 

In the second part of this survey, we review the published approaches to medical VQA. We gather the work notes describing the approaches used in the ImageCLEF VQA-Med competitions and collect 32 papers in total. However, the work notes are mainly simple solutions because of the time-limited situation. We also search for technical papers from conferences or journals that aim at the current pain point and we collect 13 papers. The papers' sources are from both the community conferences such as Medical Image Computing and Computer Assisted Intervention (MICCAI) and Association for Computing Machinery Multimedia (ACM-MM) and influential journals such as the IEEE Transactions on Medical Imaging. By reviewing those papers, we find that the current approaches are mostly in a framework of four components: image encoder, language encoder, feature fusion module, and answering module. The review of existing approaches will help researchers to identify the key problem in previous research and the potential hypothesis in future research. 

Finally, we discuss four medical-specific challenges for the field. The medical-domain VQA is a more application-oriented problem compared with the general-domain VQA. It has a real application scenario that will produce practical challenges. In this work, we analyze the clinical requirements to develop practical and useful applications and raise six significant challenges: the question diversity, extra medical information, interpretability, generalizability, large language models, and integration in the medical workflow. The proposed challenges will inspire researchers and develop mature and accurate medical VQA systems to support the clinical decision-making process.

\section{Datasets and performance metrics}
\subsection{Datasets}\label{sec:dataset}

To the best of our knowledge, there are 8 public-available medical VQA datasets up to date: VQA-MED-2018~\cite{hasan2018overview}, VQA-RAD~\cite{lau2018dataset}, VQA-MED-2019~\cite{abacha2019vqa}, RadVisDial~\cite{kovaleva2020towards}, PathVQA~\cite{he2020pathvqa}, VQA-MED-2020~\cite{abacha2020overview}, SLAKE~\cite{liu2021slake}, and VQA-MED-2021~\cite{ImageCLEF-VQA-Med2021} (in chronological order). The dataset's details are summarized in Table~\ref{table1}. In the following paragraphs, we provide an overview of the QA pairs collection.

\begin{table*}[width=\textwidth,pos=htb]
\begin{scriptsize}
    \caption{Overview of the medical VQA datasets and their main characteristics. Visual Genome, VQA 2.0, and OK-VQA are general-domain VQA datasets listed here for comparison. The medical VQA datasets are presented in chronological order.}
    \begin{tabular*}{\tblwidth}{@{\extracolsep{\fill}}p{80pt}p{45pt}p{45pt}p{100pt}p{45pt}p{110pt}} 
    \toprule
    Dataset&\# Images&\# QA pairs&Source of images\par and content&QA\par Creation&Question Category\\
    \midrule
    Visual Genome~\cite{krishna2017visual}&108K&1,773K&YFCC100M~\cite{thomee2016yfcc100m}\par Microsoft COCO~\cite{lin2014microsoft}&Manual&- Object\par - Attibutes\par - Relationships\\
    VQA 2.0~\cite{balanced_vqa_v2}&204K&614K&Microsoft COCO&Manually&
        - Object\par - Color\par - Sport\par - Count\par - etc.\\
    OK-VQA~\cite{marino2019ok}&14,031&14,055&Microsoft COCO&Manual&- External knowledge\\
    \midrule
    VQA-Med-2018~\cite{hasan2018overview}&2,866&6,413&PubMed Central Articles\footnote{\underline{https://www.ncbi.nlm.nih.gov/pmc/}}&Synthetical&- Location\par - Finding\par - Yes/No questions\par - Other questions\\
    
    VQA-RAD~\cite{lau2018dataset}&315&3,515&MedPix database:\par - Head axial single-slice CTs or MRIs\par - Chest X-rays\par  - Abdominal axial CTs&Manual&- Modality\par - Plane\par - Organ System\par - Abnormality\par - Object/Condition Presence\par - Positional Reasoning\par - Color\par - Size\par - Attribute Other\par - Counting\par - Other\\
    
    VQA-Med-2019~\cite{abacha2019vqa}&4,200&15,292&MedPix database:\par - Various in 36 modalities, 16 planes, and 10 organ systems&Synthetical&- Modality\par - Plane\par - Organ system\par - Abnormality\\
    
    RadVisDial~\cite{kovaleva2020towards}\par(Silver-standard)&91,060&455,300&MIMIC-CXR~\cite{johnson2019mimiccxrjpg}:\par - Chest X-ray posterior-anterior~(PA) view&Synthetical&Abnormality\\
    
    RadVisDial~\cite{kovaleva2020towards}\par(Gold-standard)&100&500&MIMIC-CXR~\cite{johnson2019mimiccxrjpg}:\par - Chest X-ray posterior-anterior~(PA) view&Manual&Abnormality\\
    
    PathVQA~\cite{he2020pathvqa}&4,998&32,799&Electronic pathology textbooks\par PEIR Digital Library&Synthetical&- Color\par - Location\par - Appearance\par - Shape\par - etc.\\
    
    VQA-Med-2020~\cite{abacha2020overview}&5,000&5,000&MedPix database&Synthetical&- Abnormality\\
    
    SLAKE~\cite{liu2021slake}&642&14K&Medical Segmentation Decathlon\cite{simpson2019large},\par NIH Chest X-ray\cite{ChestXRay8},\par CHAOS\cite{CHAOSdata2019}:\par - Chest X-rays/CTs\par - Abdomen CTs/MRIs\par - Head CTs/MRIs\par - Neck CTs\par - Pelvic cavity CTs&Manual&- Organ\par - Position\par - Knowledge Graph\par - Abnormality\par - Modality\par - Plane\par - Quality\par - Color\par - Size\par - Shape\\
    VQA-Med-2021~\cite{ImageCLEF-VQA-Med2021}&5,000&5,000&MedPix database&Synthetical&- Abnormality\\
    \bottomrule
    \end{tabular*}
    \label{table1}
\end{scriptsize}
\end{table*}

\subsubsection{VQA-Med-2018} 
VQA-Med-2018~\cite{hasan2018overview} is a dataset proposed in the ImageCLEF 2018\footnote{\underline{https://www.imageclef.org/2018}}, and it is the first publicly available dataset in the medical domain. The QA pairs were generated from captions by a semi-automatic approach. First, a rule-based question generation (QG) system\footnote{\underline{http://www.cs.cmu.edu/\~ark/mheilman/questions/}} automatically generated possible QA pairs by sentence simplification, answer phrase identification, question generation, and candidate questions ranking. Then, two expert human annotators (including one expert in clinical medicine) manually checked all generated QA pairs in two passes. Respectively, one pass ensures semantic correctness, and another ensures clinical relevance to associated medical images.

\subsubsection{VQA-RAD}
VQA-RAD~\cite{lau2018dataset} is a radiology-specific dataset proposed in 2018. The image set is a balanced one containing samples by the head, chest, and abdomen from MedPix\footnote{\underline{https://medpix.nlm.nih.gov/home}}. To investigate the question in a realistic scene, the author presented the images to clinicians to collect unguided questions. The clinicians are required to produce questions in both free-from and template structures. Afterward, the QA pairs are validated and classified manually to analyze the clinical focus. The answer types are either close-ended or open-ended. Although without a large quantity, the VQA-RAD dataset has acquired essential information about what a medical VQA system should be able to answer as an AI radiologist.
\begin{table*}[width=\textwidth,pos=htb]
    \caption{Samples of images and question-answer pairs from the mentioned Datasets. Q = Question, A = Answer. The datasets are presented in chronological order.}
    \begin{tabular*}
    {\tblwidth}{@{\extracolsep{\fill}}m{80pt}Y{80pt}m{80pt}Y{80pt}m{80pt}} 
    \toprule
    Dataset&\multicolumn{4}{c}{Samples}\\
    \midrule
    VQA-Med-2018~\cite{hasan2018overview}
    &\includegraphics[width=80pt,margin=0pt 3pt]{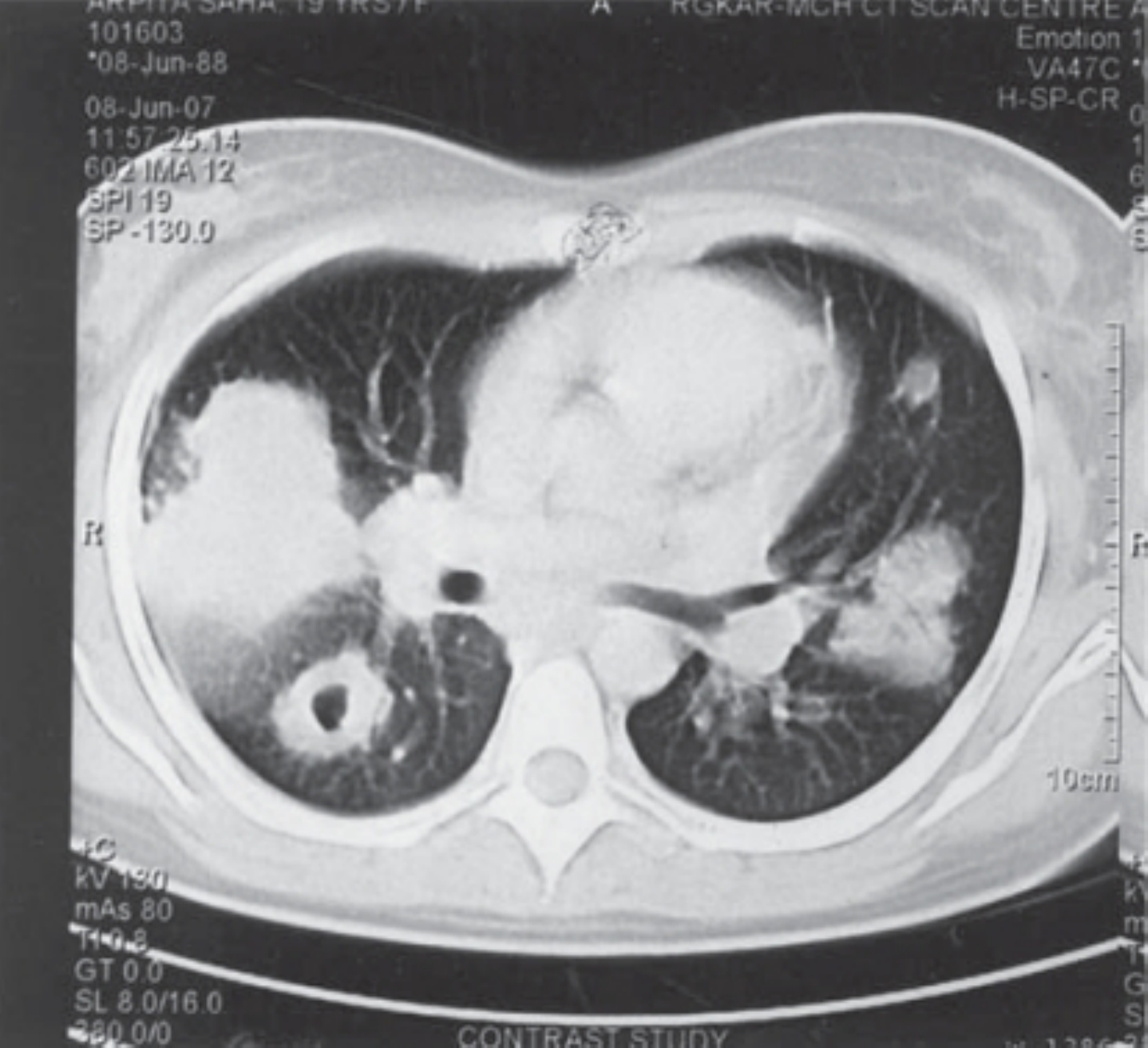}
    &\textbf{Q:} What does the ct scan of thorax show?\par\textbf{A:} bilateral multiple pulmonary nodules
    &\includegraphics[height=80pt,margin=0pt 3pt]{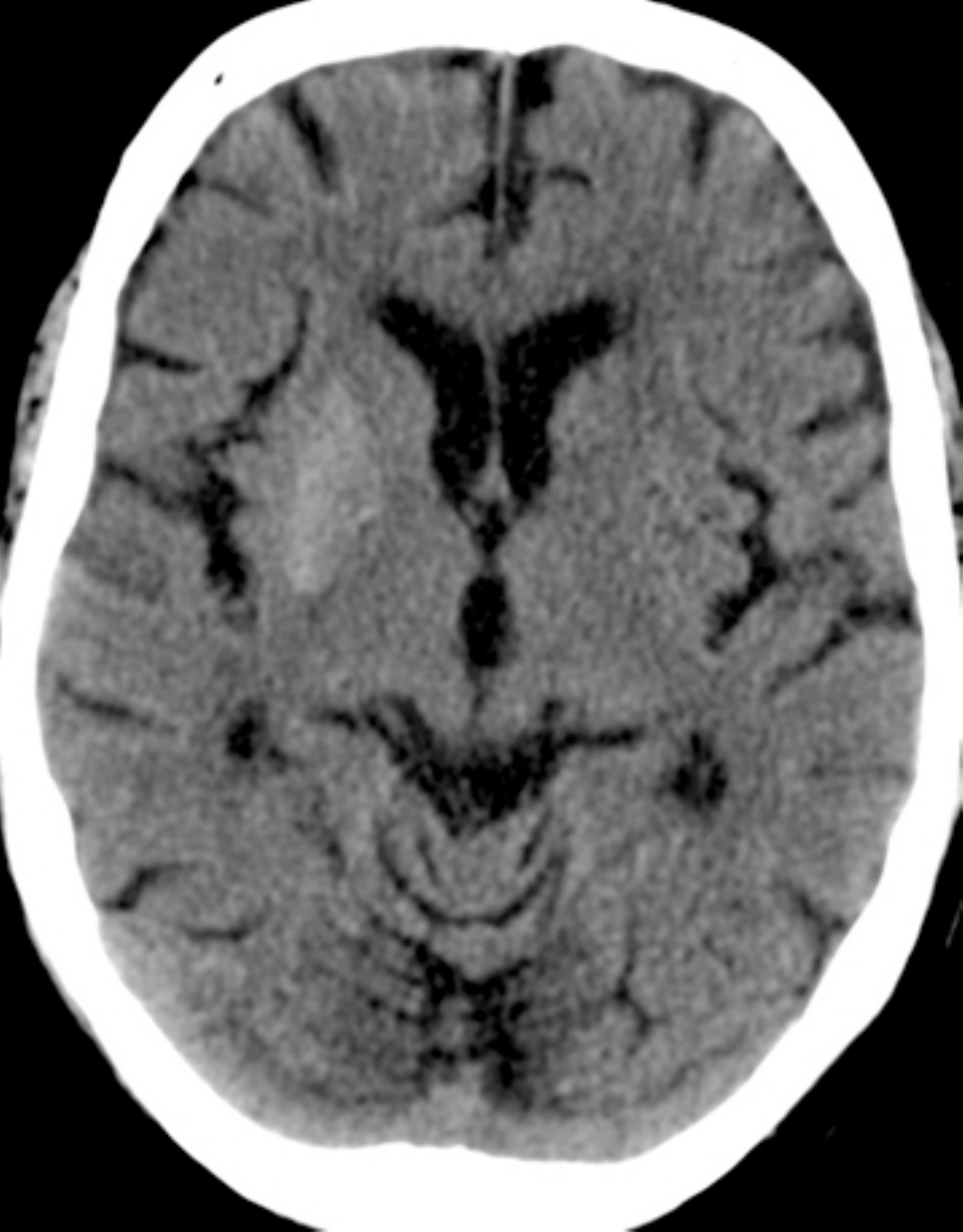}
    &\textbf{Q:} Is the lesion associated with a mass effect?\par\textbf{A:} no\\
    VQA-RAD~\cite{lau2018dataset}
    &\includegraphics[width=80pt,margin=0pt 3pt]{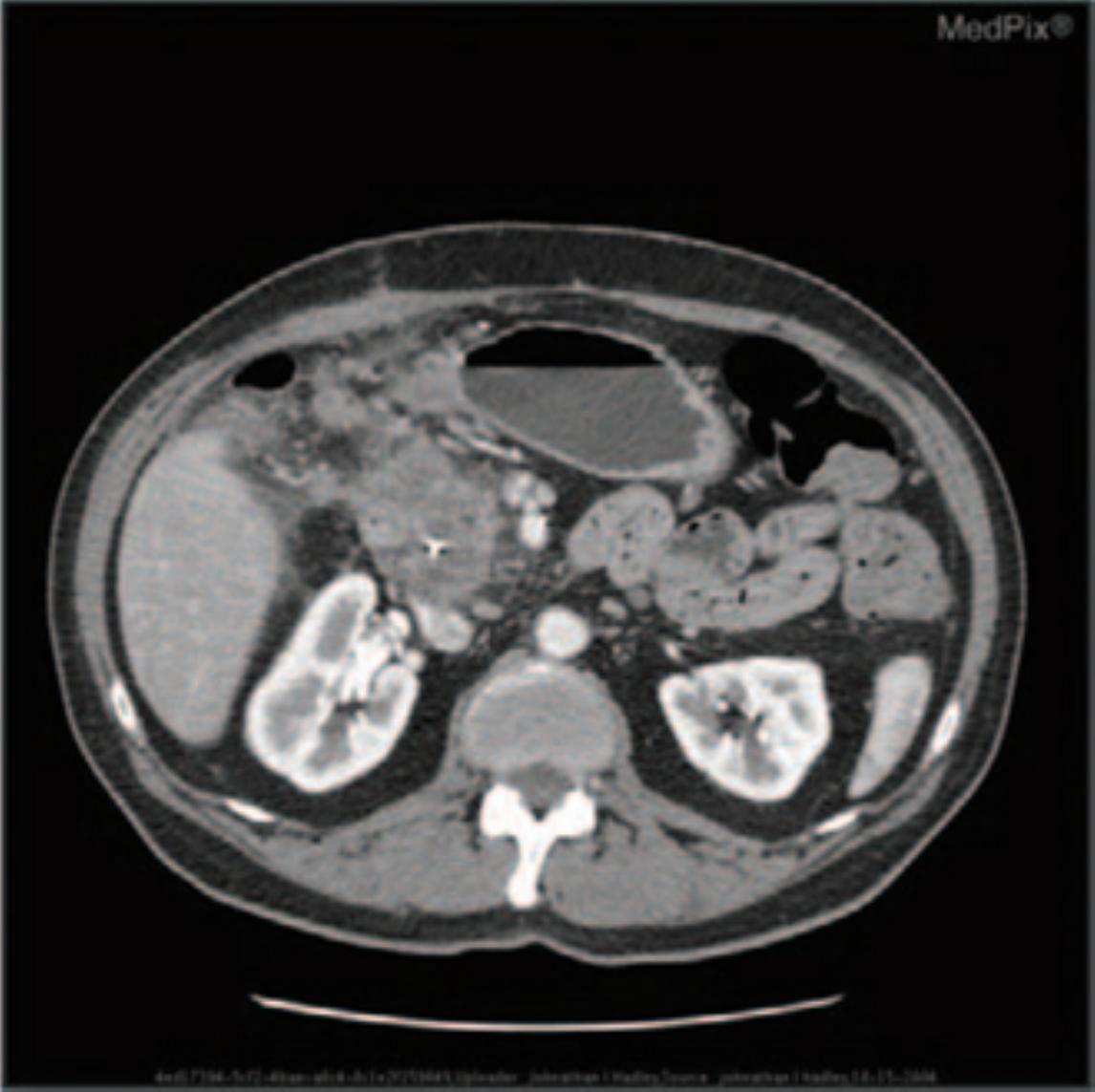}
    &\begin{center}\textbf{Organ System}\end{center}\par\textbf{Q:} What is the organ system?\par\textbf{A:} Gastrointestinal
    &\begin{center}\textbf{Object/Condition Presence}\end{center}\par\begin{flushleft}\textbf{Q:} Is there gastric fullness?\par\textbf{A:} yes\end{flushleft}
    &\begin{center}\textbf{Positional}\end{center}\par\textbf{Q:} What is the location of the mass?\par\textbf{A:} head of the pancreas\\
    VQA-Med-2019~\cite{abacha2019vqa}
    &\includegraphics[width=80pt,margin=0pt 3pt]{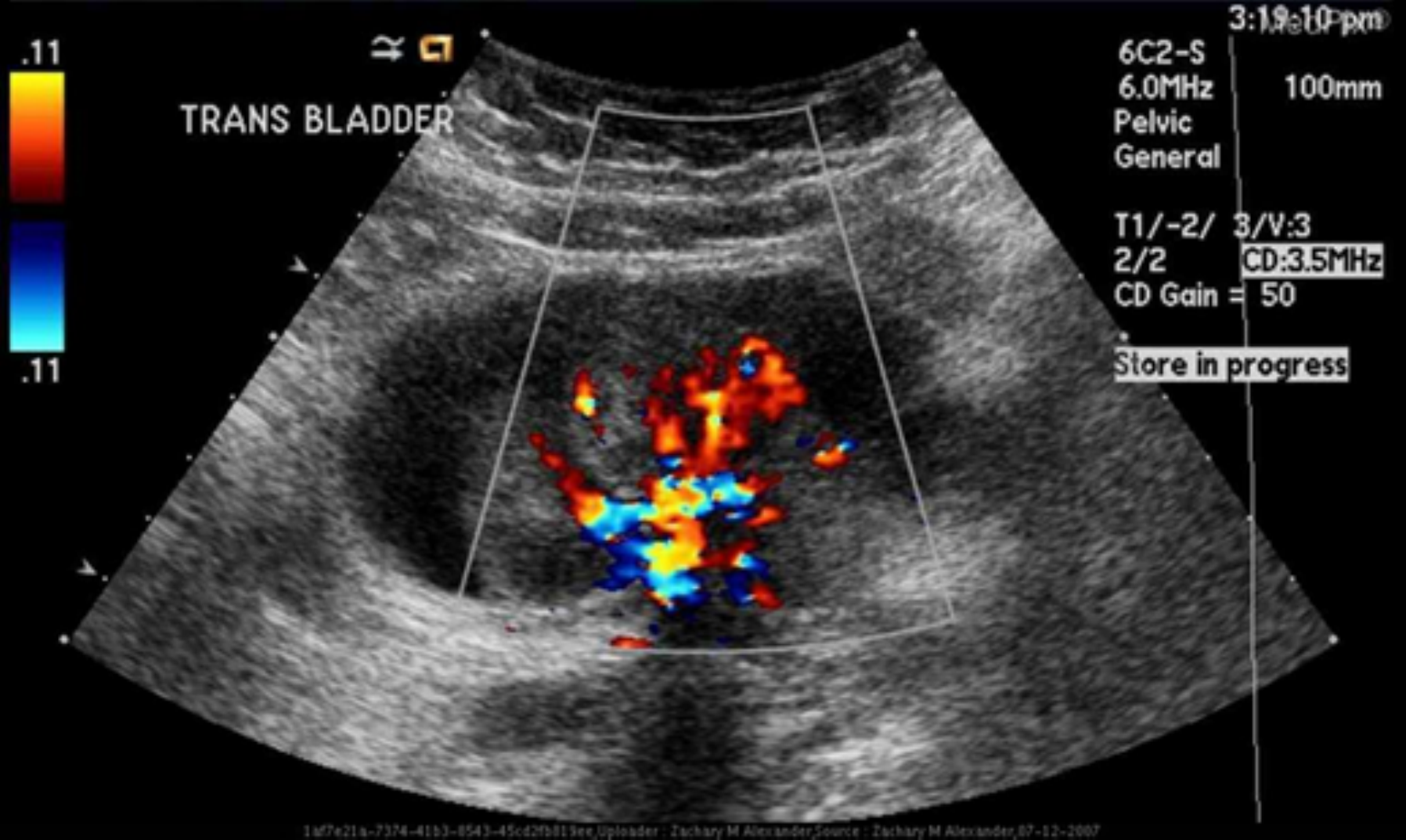}
    &\begin{center}\textbf{Modality}\end{center}\par\textbf{Q:} what imaging method was used?\par\textbf{A:} us-d - doppler ultrasound
    &\includegraphics[width=80pt,margin=0pt 3pt]{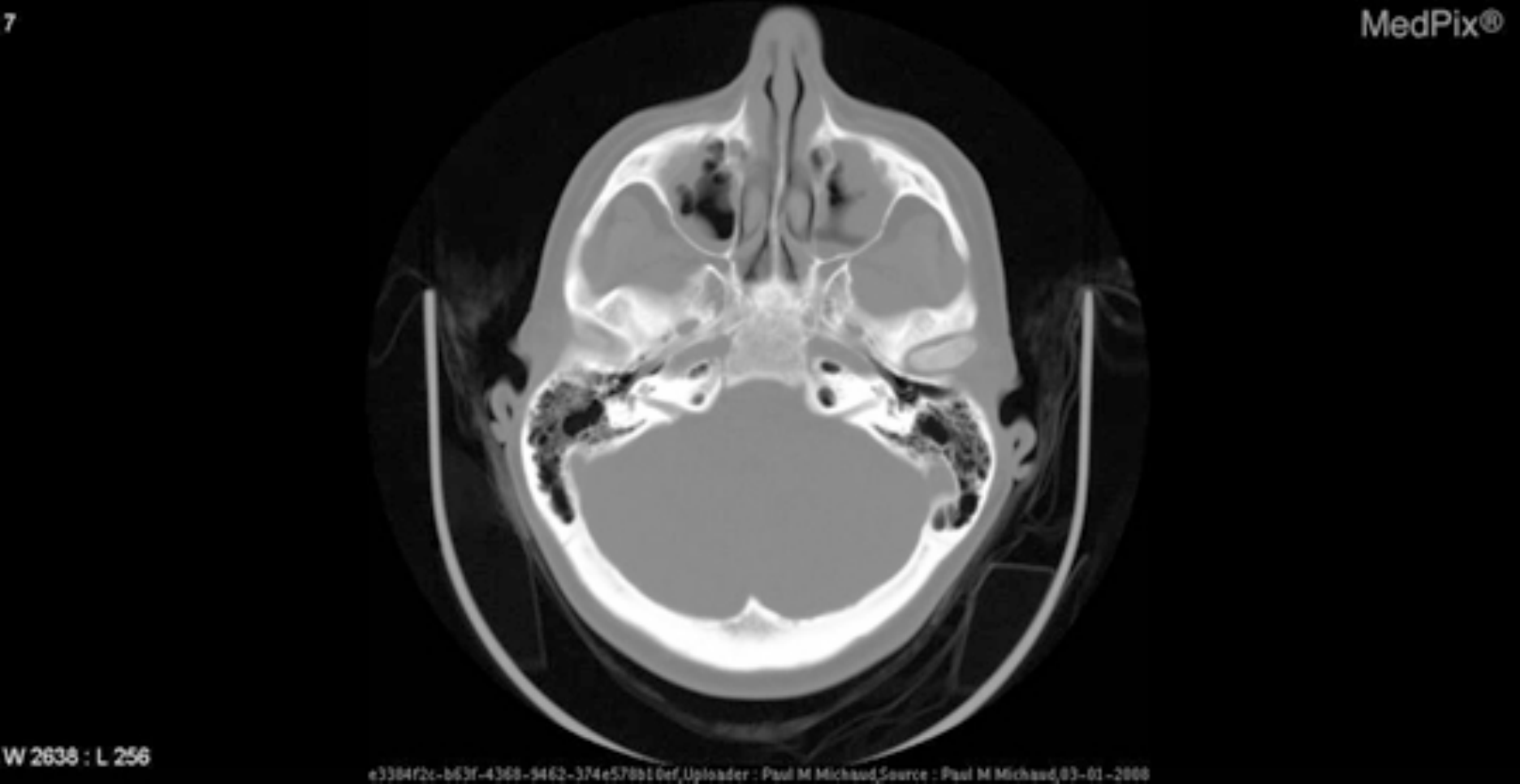}
    &\begin{center}\textbf{Plane}\end{center}\par\textbf{Q:} which plane is the image shown in?\par\textbf{A:} axial\\
    RadVisDial~\cite{kovaleva2020towards}
    &\includegraphics[width=80pt,margin=0pt 3pt]{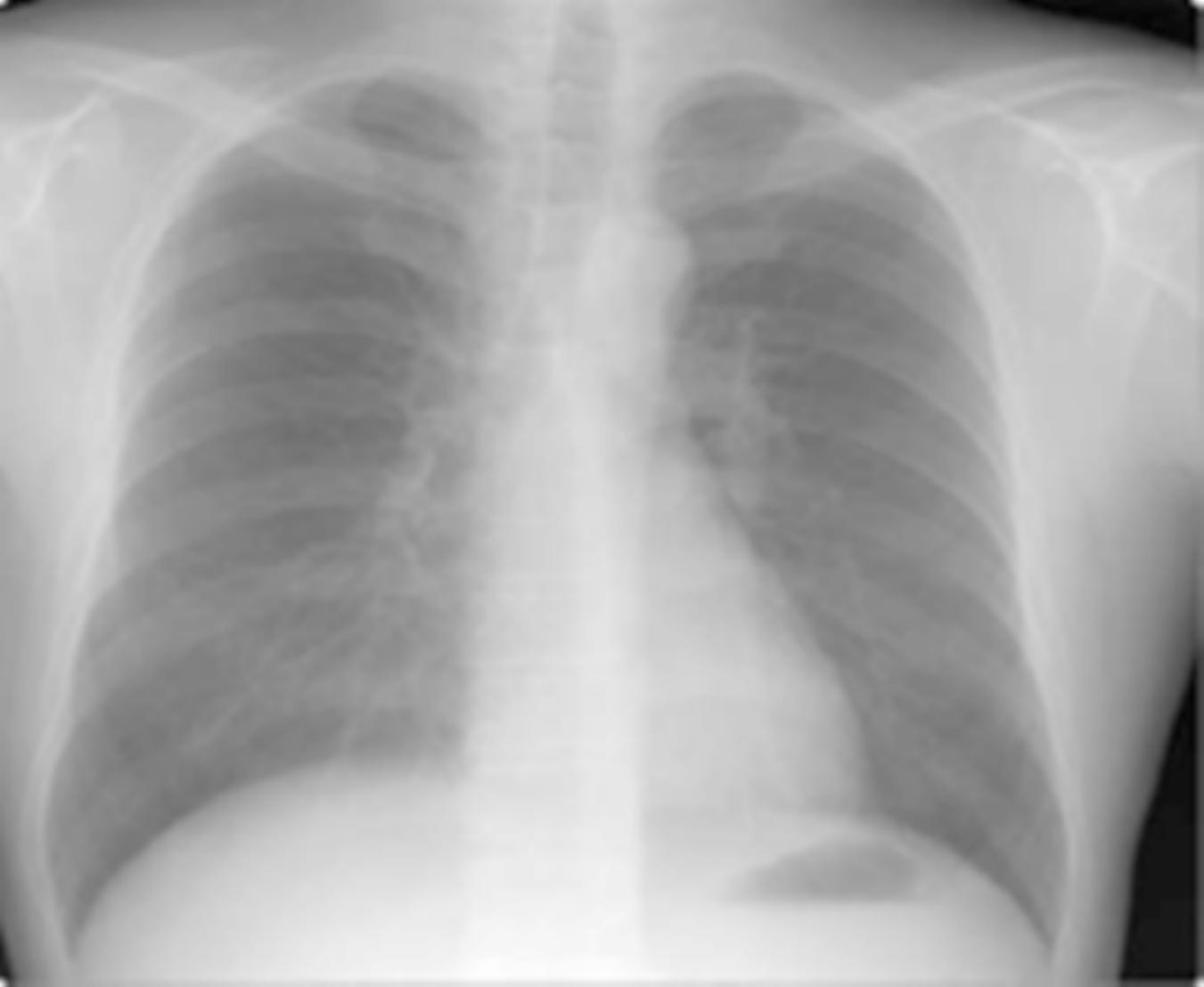}
    &\textbf{Q:} Airspace opacity?\par\textbf{A:} Yes\par
    \textbf{Q:} Fracture?\par\textbf{A:} Not in report\par
    &\begin{flushleft}\textbf{Q:} Lung lesion?\par\textbf{A:} No\par
    \textbf{Q:} Pneumonia?\par\textbf{A:} Yes\par\end{flushleft}
    &\\
    PathVQA~\cite{he2020pathvqa}
    &\includegraphics[height=80pt,margin=0pt 3pt]{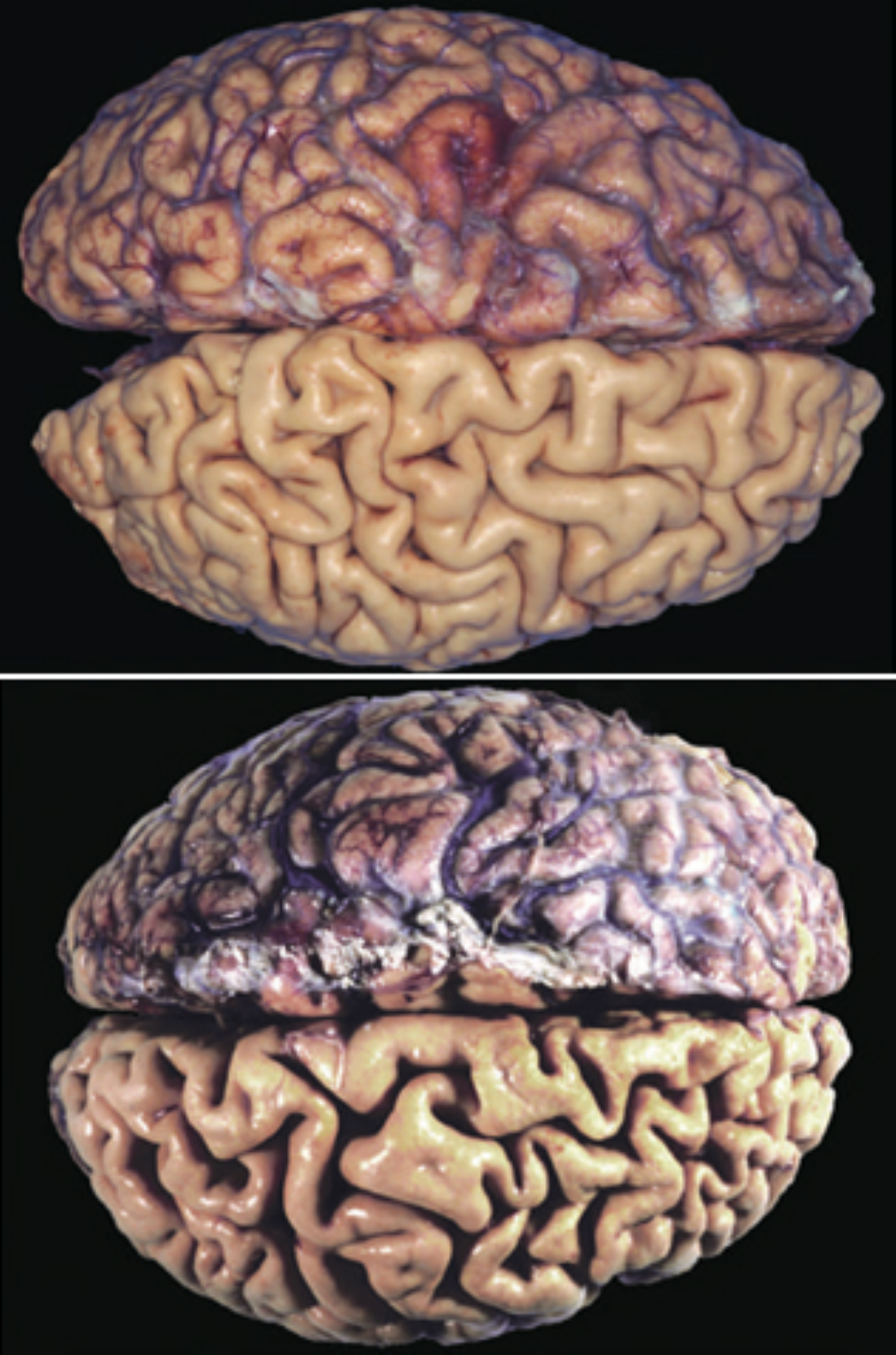}
    &\textbf{Q:} What have been stripped from the bottom half of each specimen to show the surface of the brain?\par\textbf{A:} meninges
    &\includegraphics[width=80pt,margin=0pt 3pt]{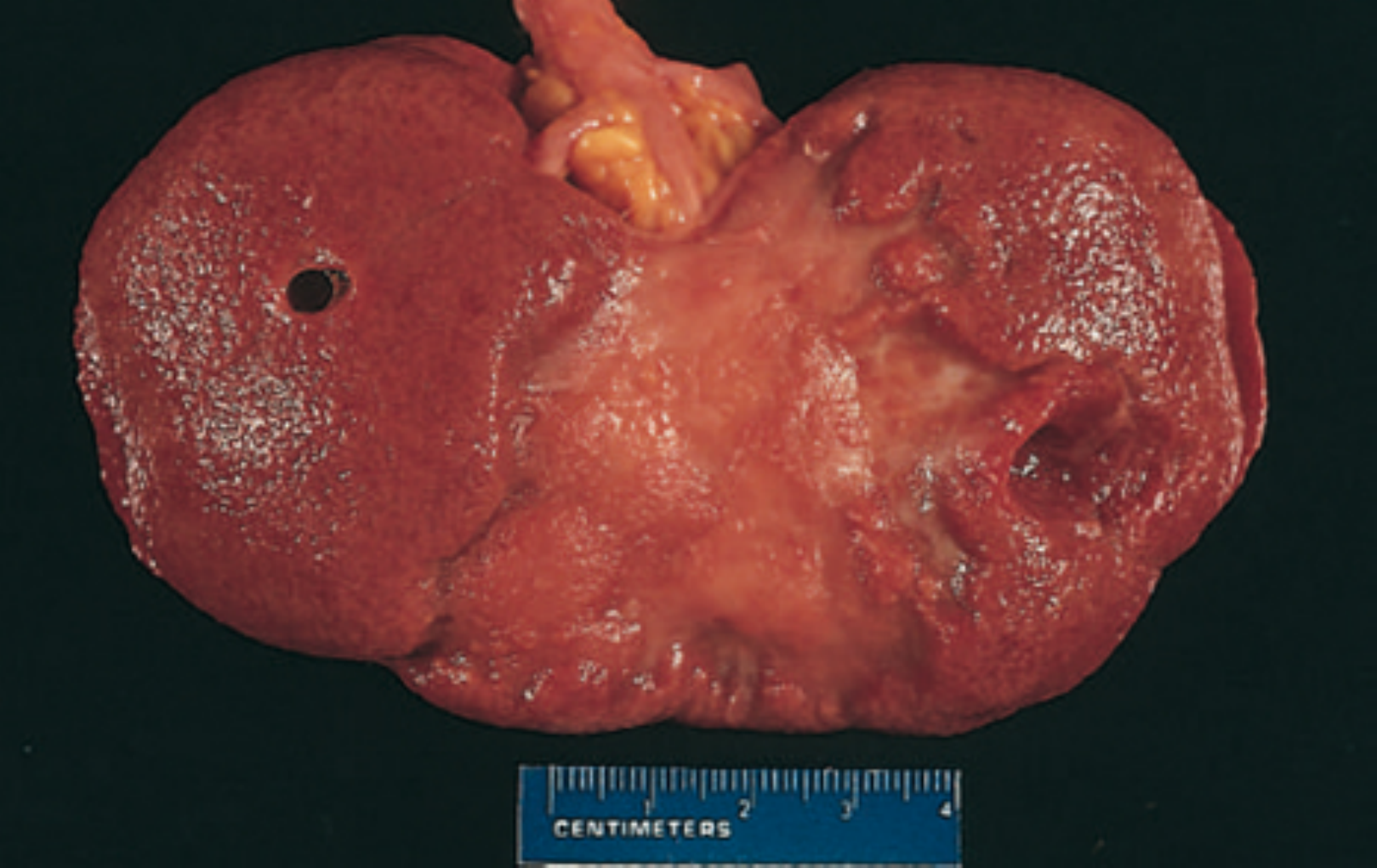}
    &\textbf{Q:} Is remote kidney infarct replaced by a large fibrotic scar?\par\textbf{A:} yes\\
    VQA-Med-2020~\cite{abacha2020overview}
    &\includegraphics[width=80pt,margin=0pt 3pt]{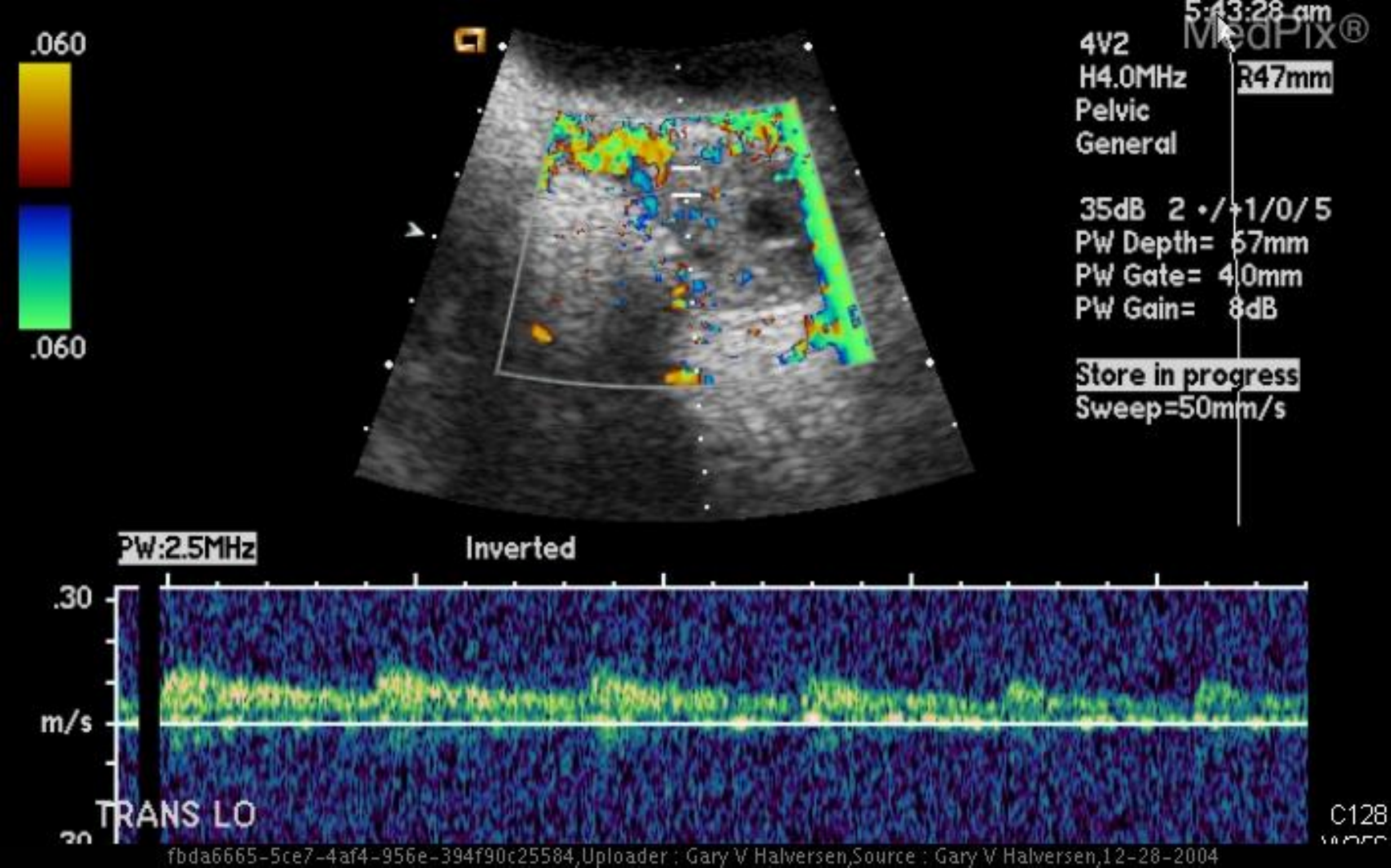}
    &\textbf{Q:} what abnormality is seen in the image?\par\textbf{A:} ovarian torsion
    &\includegraphics[width=80pt,margin=0pt 3pt]{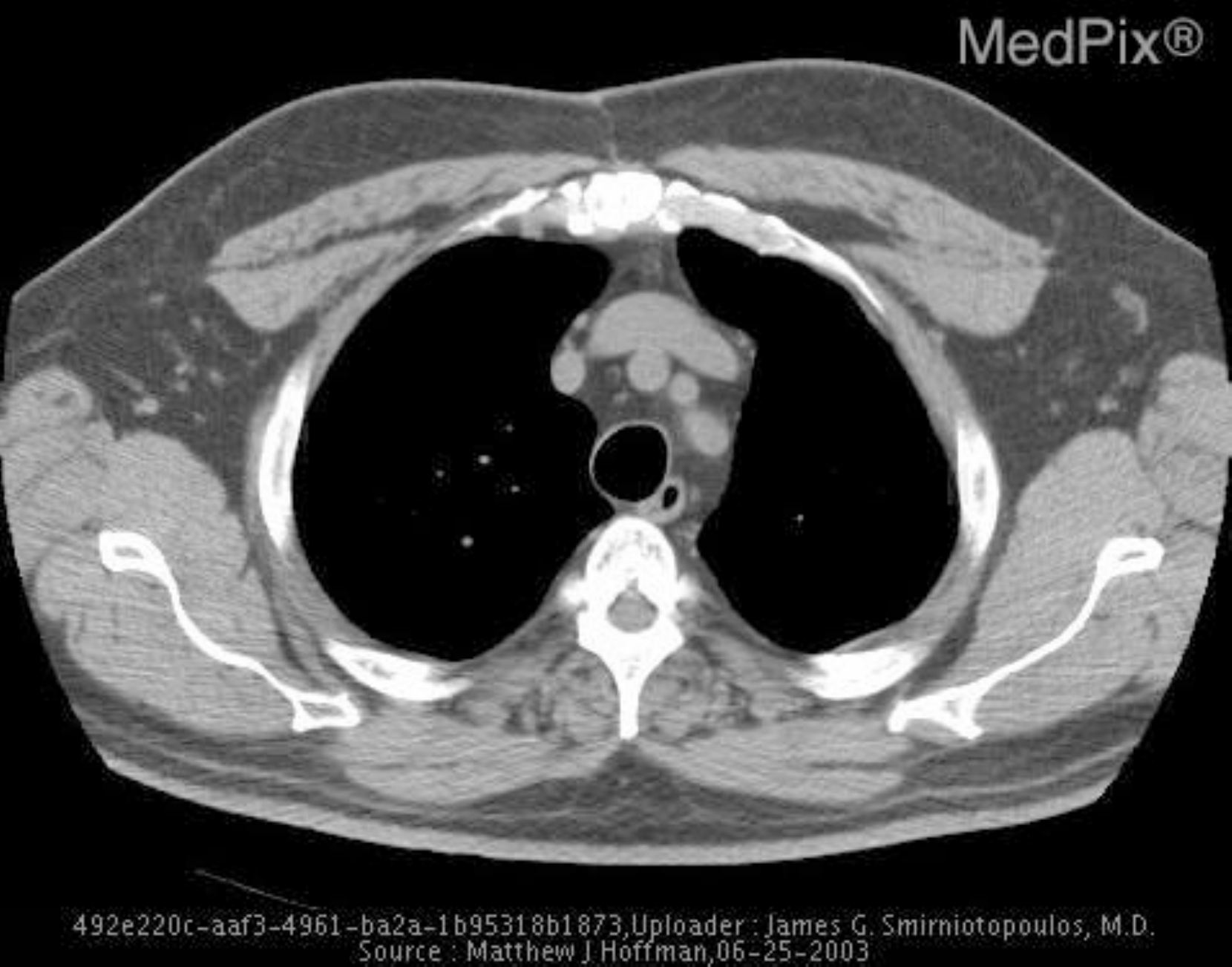}
    &\textbf{Q:} what is abnormal in the ct scan?\par\textbf{A:} partial anomalous pulmonary venous return\\
    SLAKE~\cite{liu2021slake}
    &\includegraphics[width=80pt,margin=0pt 3pt]{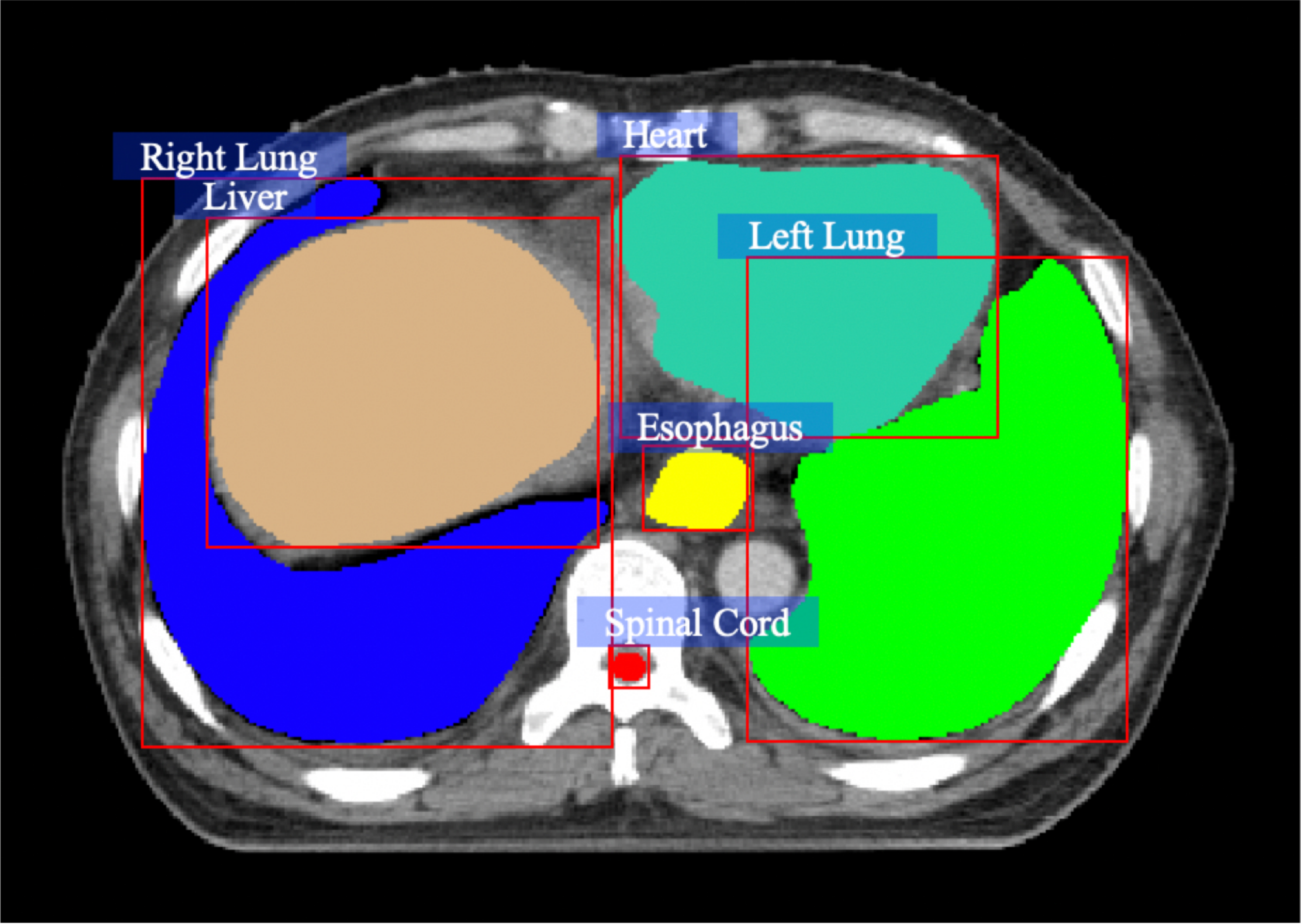}
    &\textbf{Q:} Does the image contain left lung?\par\textbf{A:} Yes
    &\begin{flushleft}\textbf{Q:} What is the function of the rightmost organ in this picture?\par\textbf{A:} Breathe\end{flushleft}
    &\\
    VQA-Med-2021~\cite{ImageCLEF-VQA-Med2021}
    &\includegraphics[height=80pt,margin=0pt 3pt]{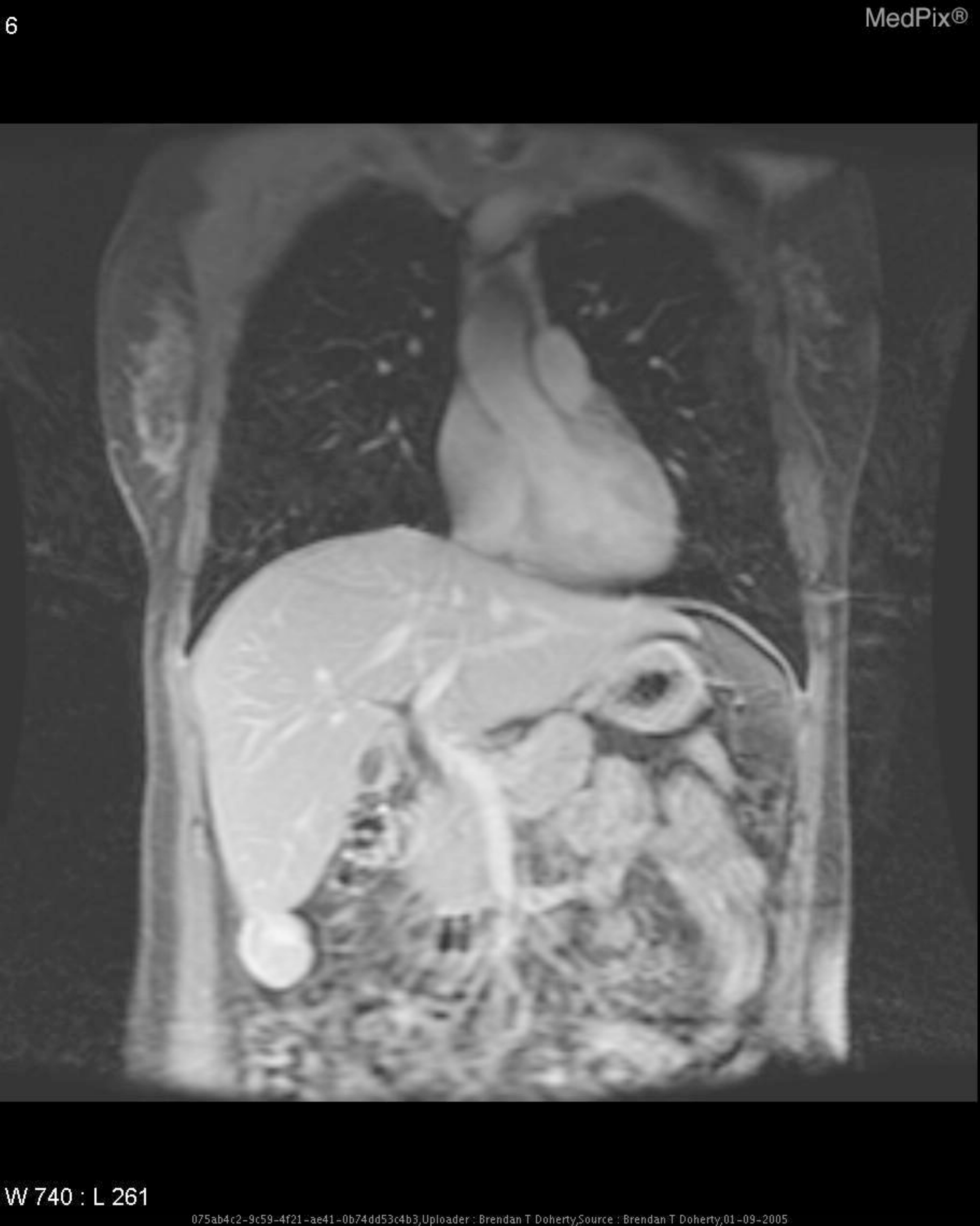}
    &\textbf{Q:} What is most alarming about this mri?\par\textbf{A:} focal nodular hyperplasia
    &\includegraphics[width=80pt,margin=0pt 3pt]{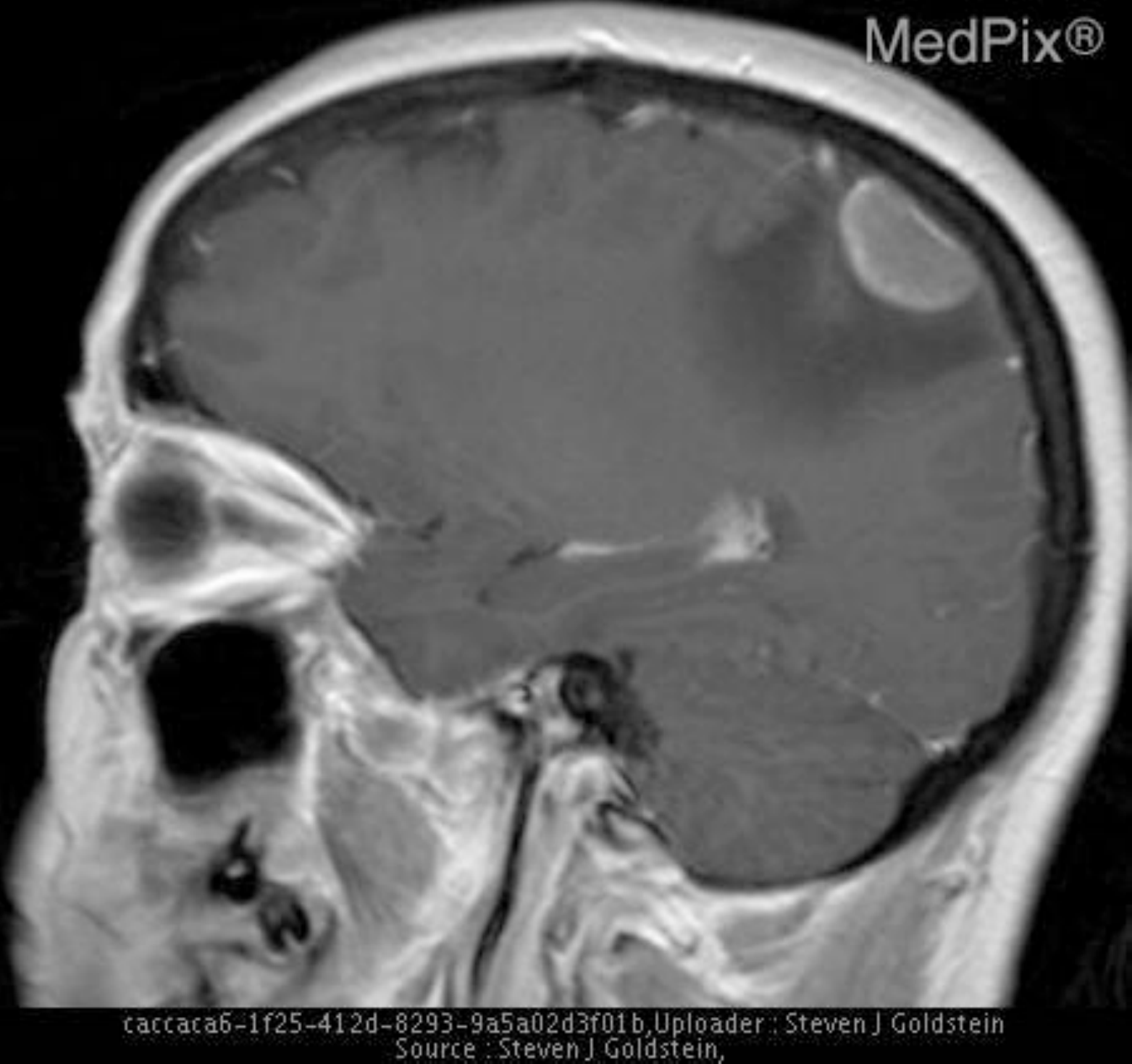}
    &\textbf{Q:} What abnormality is seen in the image?\par\textbf{A:} Enhancing lesion right parietal lobe with surrounding edema\\
    \bottomrule
    \end{tabular*}
    \label{table2}
\end{table*} 
\subsubsection{VQA-Med-2019}
VQA-Med-2019~\cite{abacha2019vqa} is the second edition of the VQA-Med and was published during the ImageCLEF 2019 challenge. Inspired by the VQA-RAD~\cite{lau2018dataset}, VQA-Med-2019 has addressed the four most frequent question categories: modality, plane, organ system, and abnormality. For each category, the questions follow the patterns from hundreds of questions naturally asked and validated in the VQA-RAD~\cite{lau2018dataset}. The first three categories~(modality, plane, and organ system) can be tackled as classification tasks, while the fourth category~(abnormality) presents an answer generation problem. 
\subsubsection{RadVisDial}
RadVisDial~\cite{kovaleva2020towards} is the first publicly available dataset for visual dialog in radiology. The visual dialogue consists of multiple QA pairs and is considered a more practical and complicated task for a radiology AI system than VQA. The images are selected from MIMIC-CXR~\cite{johnson2019mimiccxrjpg}. For each image, the MIMIC-CXR has provided a well-structured relevant report with annotations for 14 labels~(13 abnormalities and one No Findings label). The RadVisDial consists of two datasets: a silver-standard dataset and a gold-standard dataset. In the silver-standard group, the dialogues are synthetically created using the plain text reports associated with each image. Each dialogue contains five questions randomly sampled from 13 possible questions. The corresponding answer is automatically extracted from the source data and limited to four choices~(yes, no, maybe, or not mentioned in the report). 
In the gold-standard group, the dialogues are collected from two expert radiologists' conversations following detailed annotation guidelines to ensure consistency. Only 100 random images are labeled with gold-standard. The RadVisDial dataset explored a real-world scene task of AI in the medical domain. Moreover, the team compared the synthetical dialogue to the real-world dialogue and conducted experiments to reflect the importance of context information. The medical history of the patient was introduced and led to better accuracy.  
\subsubsection{PathVQA}
PathVQA~\cite{he2020pathvqa} is a dataset exploring VQA for pathology. The images with captions are extracted from digital resources (electronic textbooks and online libraries). The author developed a semi-automated pipeline to transfer the captions into QA pairs, and the generated QA pairs are manually checked and revised. The question can be divided into seven categories: what, where, when, whose, how, how much/how many, and yes/no. The open-ended questions account for 50.2\% of all questions. For the close-ended ``yes/no'' questions, the answers are balanced with 8,145 ``yes'' and 8,189 ``no''. The questions are designed according to the pathologist certification examination of the American Board of Pathology (ABP). Therefore it is an exam to verify the ``AI Pathologist'' in decision support. The PathVQA dataset demonstrates that medical VQA can be applied to various scenes.
\subsubsection{VQA-Med-2020}
VQA-Med-2020~\cite{abacha2020overview} is the third edition of the VQA-Med and was published in the ImageCLEF 2020 challenge. The images are selected with the limitation that the diagnosis was made according to the image content. The questions are specifically addressing on abnormality. A list of 330 abnormality problems is selected, and each problem needs to occur at least ten times in the dataset. The QA pairs are generated by patterns created. 

In VQA-Med-2020, the visual question generation~(VQG) task is first introduced to the medical domain. The VQG task is to generate natural language questions relative to the image content. The medical VQG dataset includes 1,001 radiology images and 2,400 associated questions. The ground truth questions are generated with a rule-based approach according to the image captions and manually revised.

\subsubsection{SLAKE}
SLAKE~\cite{liu2021slake} is a comprehensive dataset with both semantic labels and a structural medical knowledge base. The images are selected from three open source datasets~\cite{simpson2019large,ChestXRay8,CHAOSdata2019} and annotated by experienced physicians. The semantic labels for images provide masks (segmentation) and bounding boxes (detection) for visual objects. 
The medical knowledge base is provided in the form of a knowledge graph. The knowledge graph is extracted from OwnThink and manually reviewed. They are in the form of triplets (e.g., \textit{<Heart, Function, Promote blood flow>}). The dataset contains 2,603 triplets in English and 2,629 triplets in Chinese. The introduction of a knowledge graph allows external knowledge-based questions such as organ function and disease prevention. The questions are collected from experienced doctors by selecting pre-defined questions or rewriting questions. Then the questions are categorized by their types and balanced to avoid bias.

\subsubsection{VQA-Med-2021}
VQA-Med-2021~\cite{abacha2020overview} is published in ImageCLEF 2021 challenge. The VQA-Med-2021 is created under the same principles as those in VQA-Med-2020. The training set is the same dataset used in VQA-Med-2020. The validate set and test set are newly collected and manually reviewed by medical professionals. 

\begin{figure}[tb]
    \centering
    \includegraphics[width=\linewidth]{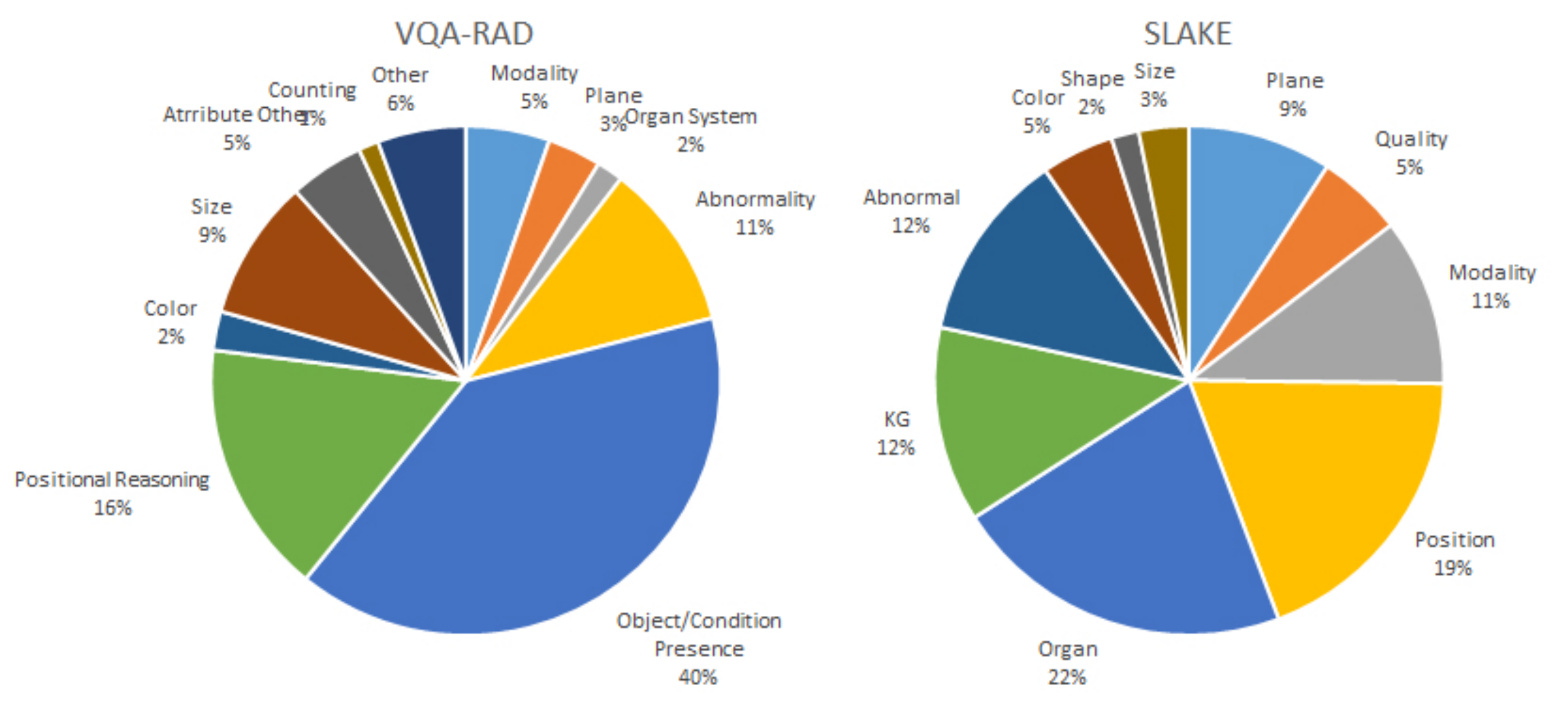}
    \caption{The question category distribution of VQA-RAD and SLAKE}
\label{fig:questiontype}
\end{figure}

\subsubsection{Discussion}
\label{data_discussion}
In the above sections, we present 8 medical VQA datasets about their quantity, data source, QA creation, and question categories. As shown in Table~\ref{table1}, we also list three general VQA datasets for comparison. The image amounts of medical VQA datasets range from 315 to 91,060, while the number of QA pairs ranges from 1 QA pair per image to 10 QA pairs per image. The imaging modality includes chest X-ray, CT, MRI, and pathology. Except for RadVisDial, the medical VQA datasets are significantly smaller than the general VQA datasets in quantity. For the QA pair creation, the medical VQA uses synthetic creation more frequently than the general domain VQA. The question categories of medical VQA and general VQA are quite different. Besides the common categories of object and attribute, the general VQA research extends their problem to objects' relationship and external knowledge, while the medical VQA research tends to image findings.

These differences between general VQA datasets and medical VQA datasets reflect the difficulties in medical dataset establishment. The three listed general VQA datasets have all utilized the Microsoft COCO\cite{lin2014microsoft}, which provides 328K images with natural language descriptions. On the medical side, there is no such large-scale data source. 
One type of data source is images with a description such as a caption and medical report. The VQA-Med-2018 is the first exploration of the medical VQA dataset. It utilizes the images in articles so that the images have a corresponding textual description. The PathVQA uses images and text from textbooks and the digital library. For these two datasets, the key problem is how to perform an accurate transformation. Another type of data source is the image with categorical attribution. The VQA-RAD, VQA-Med-2019, VQA-Med-2020, and VQA-Med-2021 are all sourced from the MedPix database, which provides images with attributions. Therefore, the key problem becomes how to acquire better questions given images and answers. The VQA-RAD starts from manual creation and collects unguided problems from clinicians. The question patterns collected from VQA-RAD leverage the construction of VQA-Med-2019, VQA-Med-2020, and VQA-Med-2021. The RadVisDial uses the data from the MIMIC-CXR dataset, which provides extracted disease labels and is the only large-scale data source in medical VQA.

Besides the data source, another difficulty is the professional knowledge required in data annotation. All three general VQA datasets adopt the Amazon Mechanical Turk workers to achieve their large-scale annotation. On the medical side, the QA creation is usually synthetical, and the manual creation is done by medical students or medical experts. The cost of manual data annotation in medical VQA is inevitably higher due to the requirement of professional knowledge.

The above difficulties in medical VQA dataset establishment have raised future challenges. With the development of technology, the existing difficulties can possibly be solved. For example, the recent Large Language Models (LLMs) have been believed good at understanding and generating natural language. The advantage of LLMs may make them ideal annotators for medical VQA. Especially for existing large-scale medical report datasets, i.e., MIMIC-CXR or FFA-IR~\cite{li2021ffa}, the LLMs can help with parsing the natural language reports and converting them into medical VQA data.

Another problem is the question categories. As shown in Fig.~\ref{fig:questiontype}, the VQA-RAD and SLAKE have different question categories distribution as they are created in different ways. Some categories such as modality and plain can help the image viewer to understand the captured information of the image, while the other categories such as abnormality and abnormality attributes can help interpret the image findings. Among all medical VQA datasets, VQA-RAD is the only one collecting the natural questions and representing a question categories distribution from medical students. In contrast, other datasets are all created with pre-defined question categories and the distribution may not represent any real-world demands. In other words, currently, there is no public dataset representing a question distribution from patients in the clinical scene. 

Furthermore, the task design and mission are also considerable problems. The recently proposed general VQA datasets have shown their special target, such as data balance, knowledge base, etc.
In the medical domain, SLAKE provides more modalities, including segmentation, detection, and knowledge graph. This feature can improve the complexity of tasks and allow more question categories. It also raises a new mission for the approach researchers as it has more modalities to expand the method's complexity. Despite the manual annotation limit its quantity, the golden standard annotation is prospective to benefit the community and future research.

As the medical VQA research is still in an early stage, the current datasets are only about radiology and pathology in data subjects. There is more field to discover, such as ophthalmology and dermatology, which are also popular in medical AI research and already has existing databases to create potential VQA task. Besides dataset works, there is also exploration addressing data collection efficiency. The MVQAS~\cite{MVQAS} builds an online system providing self-collected and annotation tools to allow users to upload data and semi-automatically generate VQA triplets.
\subsection{Performance Metrics}
The performance metrics used in the proposed medical VQA tasks can be categorized into classification-based metrics and language-based metrics. The classification-based metrics are the general metrics in classification tasks such as accuracy and F1 score. They treat the answer as a classification result and calculate the exact match accuracy, precision, recall, and e.t.c. All eight tasks in this paper use classification-based metrics as part of their performance metrics. The Language metrics are the general metrics for sentence evaluation tasks~(e.g., translation, image captioning). The tasks using language-based metrics include VQA-Med-2018, VQA-Med-2019, PathVQA, VQA-Med-2020, and VQA-Med-2021. All of those four tasks use the BLEU~\cite{papineni2002bleu}, which measures the similarity of the phrases (n-grams) between two sentences. However, the BLEU is originally a metric for machine translation and is also used in medical report generation tasks~\cite{li2021ffa}. As shown in Table~\ref{table2}, the ground truth answers in medical VQA are obviously shorter than those of machine translation or medical report generation tasks. Also, for some questions, the semantically positive or negative is more important than the word match. It suggests that BLEU may be an inappropriate metric for current medical VQA datasets. However, the BLEU can still be useful when the answer corpus of the future medical VQA becomes extensive and comprehensive sentences.

Besides the general metrics, there are also custom metrics designed for the medical VQA. For example, the WBSS (Word-based Semantic Similarity) and CBSS (Concept-based Semantic Similarity) are created in the VQA-Med-2018~\cite{hasan2018overview} as new language metrics. However, the metric is not a fixed component of the dataset and the approach. Researchers can alternatively use more suitable metrics to evaluate their results. For example, some researchers~\cite{sharma2021medfusenet} introduce the AUC-ROC~(Area under the ROC Curve) as their classification metrics to better evaluate the measure of separability.

\section{Methods}
\label{sec:method}
\begin{figure}[tb]
    \centering
    \includegraphics[width=\linewidth]{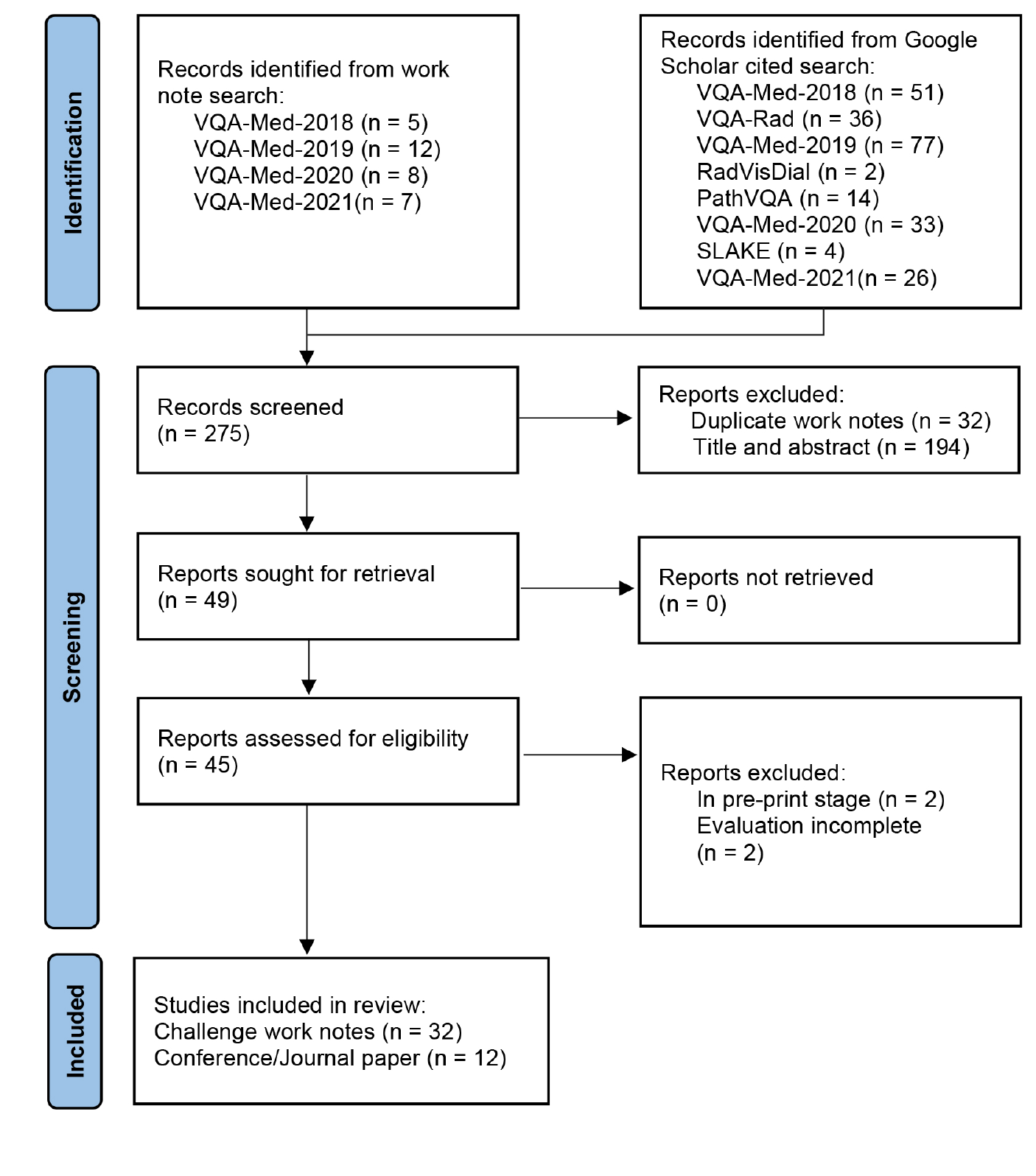}
    \caption{The number of papers included/excluded at the literature review of conference/journal papers}
\label{prisma}
\end{figure}

To investigate the feature of approaches used in the medical VQA task, we reviewed the published papers evaluated on the datasets. We search for the method papers with two strategies. For the ImageCLEF competitions, we use the corresponding overview papers~\cite{abacha2018nlm,abacha2019vqa,abacha2020overview,ImageCLEF-VQA-Med2021} to identify the participating teams and collect their work notes. This strategy helps us collect a total number of 32 papers. Then we search Google Scholar with the citation search function to find 217 papers that cite medical VQA datasets and collect a total of 12 papers. For a detailed count of the papers included and excluded at each stage, refer to Fig.~\ref{prisma}. Finally, we select 45 published papers, including 32 work notes and 13 conference/journal papers. The 45 papers describe 46 approaches, and the performance and characteristics are shown in Table~\ref{table3}. The following sub-sections discuss the medical VQA methods by the framework, components, other techniques, performance comparison, and overall discussion. 

\subsection{Framework}

\begin{figure*}[htb]
    \centering
    \includegraphics[width=\linewidth]{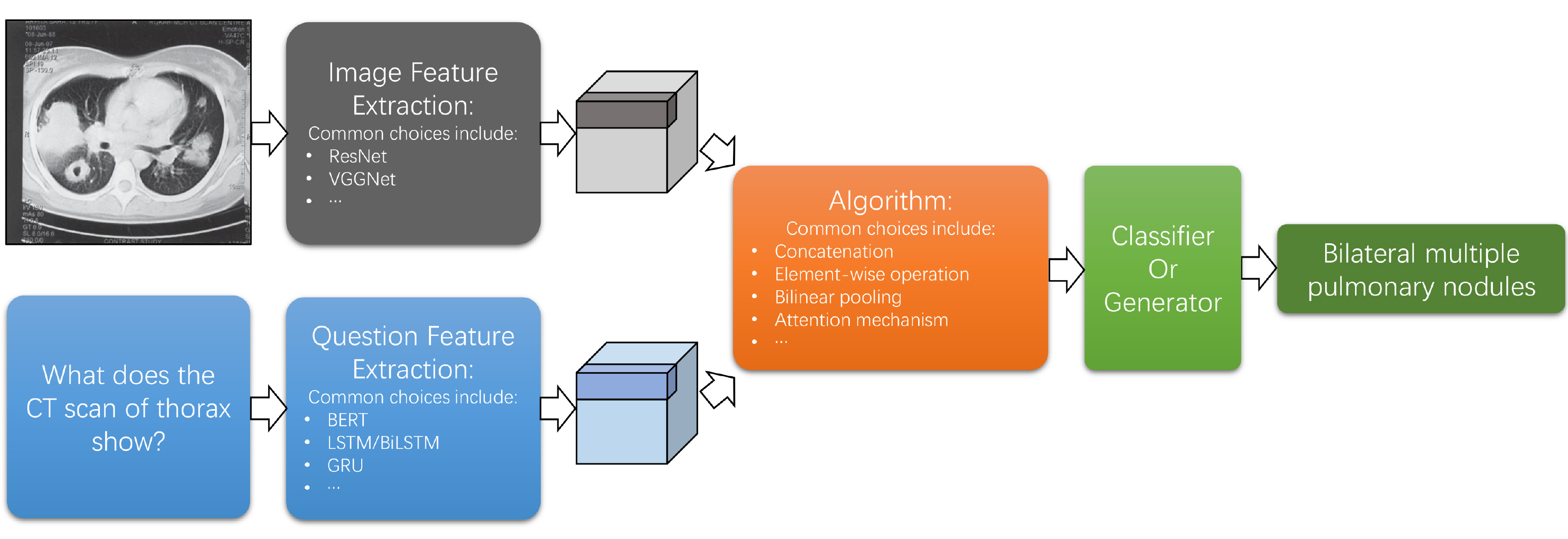}
    \caption{A schematic diagram of the mainstream medical VQA frameworks illustrating main components, including image feature extraction, question feature extraction, feature fusion, and answering component.}
\label{fig1}
\end{figure*}

Among the 46 approaches, 39 of them can be attributed to a common framework in the general VQA domain, the \textbf{joint embedding~\cite{antol2015vqa}}. It is proposed as the baseline method for the VQA v1 dataset~\cite{antol2015vqa} and referred to as the "LSTM Q+I". As illustrated in Fig.~\ref{fig1}, the framework includes four components: an image encoder, a question encoder, a feature fusing algorithm, and an answering component according to the task requirement. Respectively, the image feature extractor can be the well-developed convolution neural network (CNN) backbones such as VGG Net~\cite{simonyan2014deep} and ResNet~\cite{he2016deep}, and the question encoder can be the prevalent language encoding models such as LSTM~\cite{hochreiter1997long} and the Transformer~\cite{vaswani2017attention}. The feature encoding models are often initialized with pre-trained weights, and they can be either frozen or fine-tuned in an end-to-end manner during the training of the VQA task. The answering component is usually a neural network classifier or a recurrent neural network language generator. In the LSTM Q+I, the question features and image features are fused via element-wise multiplication. Then, researchers developed innovative fusing algorithms and introduced the popular attention mechanism into the system to further increase the performance. 

Besides the joint embedding framework, the other 7 approaches choose not to include the question feature. They find the semantic space of the questions is simply due to the data nature and use only the image feature to produce the answer. The framework has outstanding performance in VQA-Med-2020 and VQA-Med-2021, where the questions are only about abnormality and in only ``Yes/No'' or ``What'' types.

Compare to the general VQA domain, the architecture that appears in medical VQA research is less diverse. There is also other architecture in the general VQA domain, such as compositional model such as the Neural Module Network~\cite{Andreas_2016_CVPR}. However, no compositional model has been adopted in current medical VQA approaches, and it is also potential to introduce other frameworks to the medical VQA.

\subsection{Image Encoder}

In terms of the image encoder in the public challenges, the VGG Net~\cite{simonyan2014deep} is the most popular choice. As shown in Fig.\ref{fig:encoder_stat}, the participants using VGG Net as the image encoder represent a significant proportion in all of the three VQA-MED challenges. Meanwhile, the pre-trained image encoder is often used for both public challenges and conference/journal works. According to Table~\ref{table3}, more than half of the teams (29 of 46) directly used a pre-trained model on the ImageNet~\cite{ILSVRC15}. However, ImageNet has different content compared with medical VQA. Using ImageNet pre-trained is a non-reasonable practice but a workable option when the low data quantity and lack of labels both limit the pre-training on medical datasets. 

Finding better pre-training methods is a popular topic in medical VQA research as well as in the  medical AI community. The image numbers of most medical VQA datasets are under 5,000. It leads to difficulty in training image representation. The solution proposed includes using other pre-trained models, using the extra dataset, contrastive learning, multi-task pre-training, and meta-learning. The LIST team~\cite{allaouzi2019encoder} utilized an image encoder pre-trained on the CheXpert~\cite{ChestXRay8} dataset. The MMBERT~\cite{MMBERT} team uses an auxiliary dataset named ROCO~\cite{pelka2018roco} (images and captions) to perform pre-training in a token-masking manner. The CPRD~\cite{CPRD-MICCAI21} team introduces contrastive learning technology to conduct pre-training with unlabeled images in a self-supervised scheme. Self-supervised training has the advantage that it does not need image labels, which are expensive to acquire for medical images. The MTPT-CSMA~\cite{MTPT-CSMA} team uses an auxiliary dataset of segmentation tasks. On the other hand, researchers also try to further digest the original data. The MEVF~\cite{nguyen2019overcoming} team proposed the Mixture of Enhanced Visual Features (MEVF) component, which utilizes the Model-Agnostic Meta-Learning (MAML) and the Convolutional Denoising Auto-Encoder (CDAE) to initialize the model weights for the image encoder to overcome the data limitation in quantity. The various exploration on image encoder is specific in medical VQA compared that in general VQA.

Notably, the auxiliary datasets selected have three different types: captioning dataset, unlabeled image, and segmentation dataset. It indicates that the pre-training on various extra data can leverage the image decoder's performance. Hence, exploring the methods to pre-train the image encoder will be a prospective task. Another feature is that all image encoders in the reviewed papers are CNN classification models than detection ones. It restricts the application of detection-based methods such as Up-Down~\cite{anderson2018bottom}, which are popular in the general VQA domain. 

\subsection{Language Encoder}
\begin{figure*}[htb]
     \centering
     \begin{subfigure}{\textwidth}
         \centering
         \includegraphics[width=\textwidth]{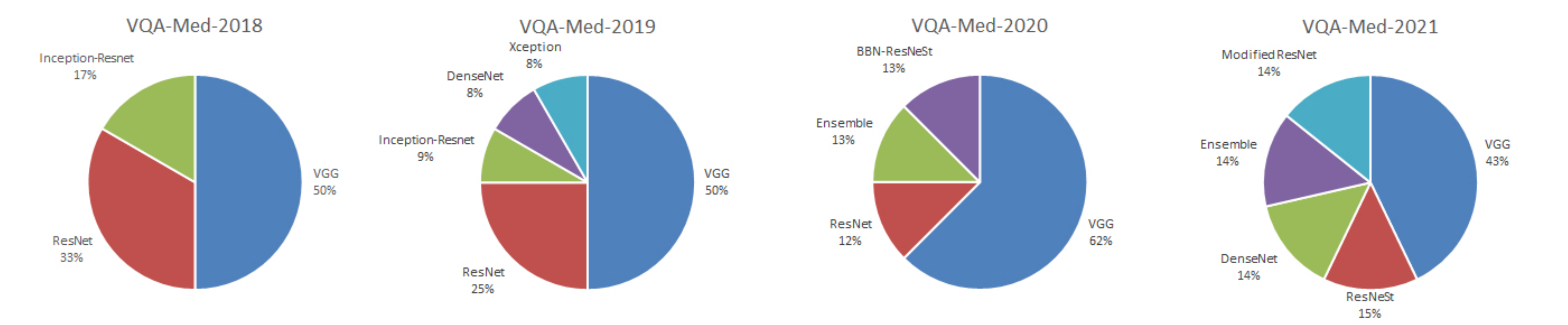}
         \caption{Composition of various network architectures as image encoders.}
     \end{subfigure}
     \hfill
     \begin{subfigure}{\textwidth}
         \raggedright
         \includegraphics[width=\textwidth]{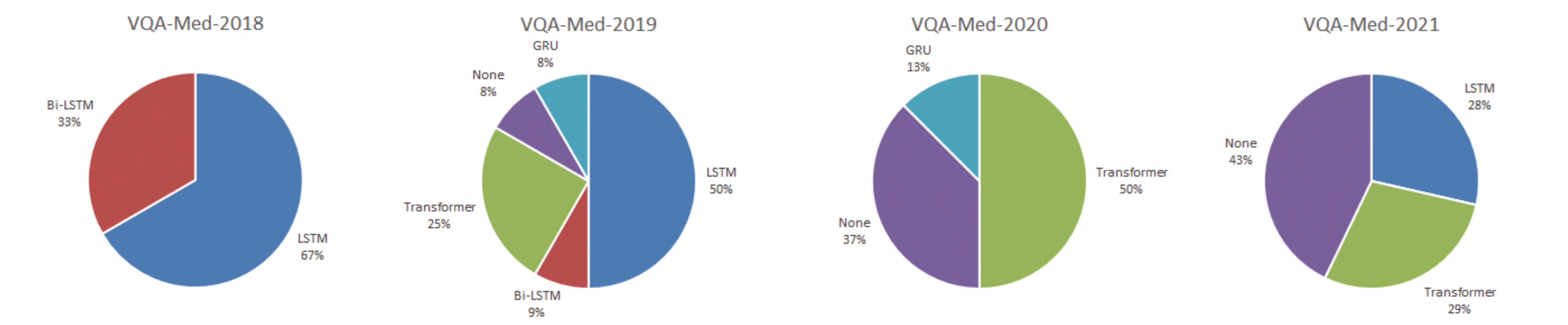}
         \caption{Composition of various network architectures as language encoders.}
     \end{subfigure}
        \caption{The encoders used in VQA-Med challenges.}
        \label{fig:encoder_stat}
\end{figure*}

The language encoders in the reviewed works include LSTM~\cite{hochreiter1997long} (18 of 46),  Bi-LSTM~\cite{schuster1997bidirectional} (5 of 46), GRU~\cite{GRU} (3 of 38), the Transformer~\cite{vaswani2017attention} (including BERT~\cite{devlin2019bert} and BioBERT~\cite{lee2020biobert}) (12 of 46) and the others (8 of 46). Notably, all the top five teams are using The BERT in the VQA-Med-2019 challenge. It is shown in Fig.\ref{fig:encoder_stat} that participating teams select BERT/BioBERT instead of LSTM/Bi-LSTM/GRU over time. This indicates the advantage of the Transformer model and the BERT pre-training despite the corpus differences between the general domain and medical domain. Meanwhile, the Recurrent Neural Networks (LSTM/Bi-LSTM/GRU) users are also adopting the pre-trained word embedding component (12 of 26). It indicates that the researchers tend to use a pre-training model to process the questions. However, only the MMBERT team~\cite{MMBERT} conducts personalized pre-training with a self-supervised method, the token masking strategy on extra data. Compared with the image encoder, the language encoder does not get much research. It is potential to develop more NLP pre-training methods or vision plus language pre-training methods for medical VQA. 

Especially, 7 teams process the questions without a deep learning model but with keyword or template matching. The reason is that keyword or template
matching has been powerful in their tasks. Hence, a light language encoder can be a practical choice in medical VQA applications as the task may not have a large number of question categories. 

\subsection{Fusion Algorithm}
The fusion stage gathers the extracted visual feature and language feature and then models the hidden relationship between the language feature and the visual feature. It is the core component of VQA methods. The typical fusion algorithm includes the attention mechanism and the pooling module. 

\textbf{Attention mechanism} is widely used in vision and language tasks. Among the Medical VQA approaches, 23 of 46 apply attention mechanisms in the fusion stage. Stacked Attention Networks (SAN)~\cite{yang2015stacked} is a typical attention algorithm. It uses the question feature as a query to rank the answer-related image regions. With a multiple-layer structure, the SAN can query an image several times to infer the answer progressively. The SAN introduced an incursive attention mechanism and is used as a baseline for many datasets. Besides the SAN, some other works employ the attention mechanism, such as the Bilinear Attention Networks (BAN)~\cite{kim2018bilinear}, the Hierarchical Question-Image Co-Attention (HieCoAtt)~\cite{lu2016hierarchical}, e.t.c. Notably, the popular multi-head attention methods(e.g., the Transformer~\cite{vaswani2017attention}, the Modular Co-Attention Network (MCAN)~\cite{yu2019deep}) are seldom applied to medical VQA. 

\onecolumn
\begin{landscape}
\begin{small}

\begin{longtable}[t]{p{55pt}p{50pt}p{45pt}p{45pt}p{45pt}p{30pt}p{50pt}p{60pt}p{100pt}p{40pt}p{40pt}}
\caption{The Results and Characteristics of Approaches in medical VQA Tasks.\label{table3}}\\
    \toprule
    Team/Method&
    Image Encoder&Pre-trained (Image)&
    Language Encoder&Pre-trained (Language)&
    Attention (Fusion)&Fusion&Output Mode&
    Other Technique(s)&
    Language Score(BLEU)&Classification Accuracy\\
    \midrule
    \endfirsthead
    \toprule
    Team/Method&
    Image\par Encoder&Pre-trained\par (Image)&
    Language\par Encoder&Pre-trained\par (Language)&
    Attention\par (Fusion)&Fusion&Output Mode&
    Other Technology(s)&
    Language Score\par (BLEU)&Classification Accuracy\\
    \midrule
    \endhead
    \multicolumn{11}{l}{\textbf{VQA-MED-2018}}\\
    \midrule
    Chakri*~\cite{chakri}&VGG16&ImageNet&GRU&None&No&Element-wise multiplication&Generation (GRU)&&0.188&\\
    UMMS~\cite{peng2018umass}&ResNet-152&ImageNet&LSTM&Pre-trained Word Embedding&Yes&MFB with Co-attention&Classification&Embedding based topic model&0.162&\\
    TU~\cite{zhou2018employing}&Inception-Resnet-v2&ImageNet&Bi-LSTM&Not mentioned&Yes&Attention mechanism&Classification&Output rules&0.135&\\
    HQS*~\cite{GUPTA-HQS}&Inception-Resnet-v2&ImageNet&Bi-LSTM&Pre-trained Word Embedding&No&Concatenation&Classification&Question segregation&0.132&\\
    NLM~\cite{abacha2018nlm}&VGG16&ImageNet&LSTM&None&Yes&SAN&Classification&&0.121&\\
    NLM&ResNet-152&ImageNet&LSTM&None&No&MCB&Classification&Pre-training with extra data&0.085&\\
    JUST~\cite{talafha2018just}&VGGNet&ImageNet&LSTM&Not mentioned&No&Concatenation&Generation (LSTM)&&0.061&\\
    FSTT~\cite{18fstt}&VGG16&ImageNet&Bi-LSTM&Not mentioned&No&Concatenation&Classification&Decision Tree Classifier&0.054&\\
    \midrule
    \multicolumn{11}{l}{\textbf{VQA-MED-2019}}\\
    \midrule
    KEML*~\cite{zheng2020KEML}&VGG16&ImageNet&Transformer&BERT&No&BLOCK&Classification&Global Average Pooling,\par Meta-learning,\par Gated Graph Neural Networks,\par Knowledge-Based Representation Learning&0.912&0.938\\
    MedFuseNet*~\cite{sharma2021medfusenet}&ResNet-152&ImageNet&LSTM&BERT Embedding&Yes&MedFuseNet&Classification/ Generation (LSTM)&Image Attention,\par Image-Question Co-Attention&0.27\par (Subset)&0.789\par (Subset)\\
    MMBERT*~\cite{MMBERT}&ResNet-152&ROCO\cite{pelka2018roco}&Transformer&Yes&Yes&Multi-head attention (Transformer)&Generation (Transformer)&Pre-training with extra data&0.69&0.672\\
    CGMVQA*~\cite{ren2020cgmvqa}&ResNet-152&ImageNet&Transformer&BERT&Yes&Concatenation&Classification Generation (Transformer)&Global Average Pooling&0.659&0.64\\
    Hanlin~\cite{yan2019zhejiang}&VGG16&ImageNet&Transformer&BERT&Yes&MFB with Co-attention&Classification&Global Average Pooling&0.644&0.62\\
    minhvu~\cite{vu2019ensemble}&ResNet-152&ImageNet&Transformer&BERT&Yes&MLB, MUTAN with attention&Classification&Ensemble,\par Skip-thought&0.634&0.616\\
    TUA1~\cite{zhou2019tua1}&Inception-Resnet-v2&ImageNet&Transformer&BERT&No&Concatenation&Classification Generation (LSTM)&Sub-models,\par Question classifier&0.633&0.606\\
    QC-MLB*~\cite{Minh-QCMLB}&ResNet-152&ImageNet&Skip-thought vectors&Yes&Yes&QC-MLB&Classification&Question-centric fusion&&0.603\\
    UMMS~\cite{shi2019deep}&ResNet-152&ImageNet&Bi-LSTM&Pre-trained Word Embedding&Yes&MFH with Co-attention&Classification&SVM Question classifier&0.593&0.566\\
    IBM Research AI~\cite{kornuta2019leveraging}&VGG16&ImageNet&LSTM&Pre-trained Word Embedding&Yes&Attention mechanism Classification&Question classifier&Image size encoder&0.582&0.558\\
    LIST~\cite{allaouzi2019encoder}&DenseNet-121&CheXpert&LSTM&Pre-trained Word Embedding&No&Concatenation&Generation (LSTM)&&0.583&0.556\\
    Turner.JCE~\cite{turner2019lstm}&VGG19&Not mentioned&LSTM&Not mentioned&No&Concatenation&Classification&	Sub-models,\par Question classifier&0.572&0.536\\
    JUST19~\cite{al2019just}&VGG16&ImageNet&None&None&No&None&Classification/ Generation (LSTM)&Sub-models,\par Question classifier&0.591&	0.534\\
    Team$\_$PwC$\_$-\par Med~\cite{bansal2019medical}&ResNet-50&ImageNet&LSTM&Pre-trained Word Embedding&Yes&Attention mechanism&Classification/ Generation (LSTM)&Sub-models,\par Question classifier&0.534&0.488\\
    Techno~\cite{task2019tlemcen}&VGG16&Not mentioned&LSTM&Not mentioned&No&Concatenation&Classification&&0.486&0.462\\
    Gasmi*~\cite{Gasmi2021}&EfficientNet&ImageNet&Bi-LSTM&Not mentioned&No&Concatenation&Classification&Optimal parameter selection&0.42&0.391\\
    Dear stranger~\cite{liu2019xception}&Xception&Not mentioned&GRU&Not mentioned&Yes&Attention mechanism&Classification&&0.393&0.21\\
    abhishek-\par thanki~\cite{thanki2019manipal}&VGG19\par DenseNet-121&ImageNet&LSTM&Pre-trained Word Embedding&No&Element-wise multiplication&Generation (LSTM)&&0.462&0.16\\
    \midrule
    \multicolumn{11}{l}{\textbf{VQA-MED-2020}}\\
    \midrule
    AIML~\cite{aiml2020mvqa}&Ensemble CNNs&Not mentioned&None&Not mentioned&No&None&Classification&Multi-scale and multi-architecture ensemble&0.542&0.496\\
    TheInception-\par Team~\cite{inception2020mvqa}&VGG16&Not mentioned&None&None&No&None&Classification&Sub-models&0.511&0.48\\
    bumjun$\_$-\par jung~\cite{jung2020mvqa}&VGG16&ImageNet&Transformer&BioBERT&Yes&MFH with Co-attention&Classification&Global Average Pooling&0.502&0.466\\
    HCP-MIC~\cite{hcp2020mvqa}&BBN-ResNeSt-50&Not mentioned&Transformer&BioBERT&No&None&Classification&Bilateral-Branch Network&0.462&0.426\\
    NLM~\cite{nlm2020mvqa}&ResNet-50&ImageNet&None&Not mentioned&No&None&Classification&&0.441&0.4\\
    HARENDRA-\par KV~\cite{harendrakv2020mvqa}&VGG16&Not mentioned&Transformer&BERT&Yes&MFB&Generation (LSTM)&&0.439&0.378\\
    Shengyan~\cite{shengyan2020mvqa}&VGG16&ImageNet&GRU&Not mentioned&No&None&Generation (GRU)&&0.412&0.376\\
    kdevqa~\cite{kdevqa2020mvqa}&VGG16&Not mentioned&Transformer&BERT&No&GLU&Classification&&0.35&0.314\\
    \midrule
    \multicolumn{11}{l}{\textbf{VQA-MED-2021}}\\
    \midrule
    SYSU-HCP~\cite{gong2021sysu}&ResNets, VGGNet, and plus HAGAP&ImageNet&None&None&No&None&Classification&Hierarchically adaptive global average pooling,\par Model ensemble,\par Mixup Augment,\par Curriculum learning,\par Label smoothing&0.416&0.382\\
    Yunnan University~\cite{xiao2021yunnan}&VGG16&ImageNet&Transformer&BioBERT&Yes&MFH with Co-attention&Classification&Global Average Pooling&0.402&0.362\\
    TeamS~\cite{eslami2021teams}&ResNeSt50&Not mentioned&None&None&No&None&Classification&Bilateral-Branch Networks&0.391&0.348\\
    Lijie~\cite{li2021lijie}&VGG8&ImageNet&Transformer&BioBERT&Yes&MFH with Co-attention&Classification&&0.352&0.316\\
    IALab$\_$PUC~\cite{schilling2021puc}&DenseNet-121&ImageNet&None&None&No&None&Classification&&0.276&0.236\\
    TAM~\cite{li2021tam}&Modified ResNet-34&Not mentioned&LSTM&GLOVE word embeddings&Yes&MFB with a co-attention&Classification&&0.255&0.222\\
    Sheerin~\cite{sitarassn}&VGGNet&ImageNet&LSTM&Not mentioned&No&Element-wise multiplication&Generation(LSTM)&&0.227&0.196\\
    \midrule
    \multicolumn{11}{l}{\textbf{VQA-RAD}}\\
    \midrule
    MTPT-CMSA~\cite{MTPT-CSMA}&Multi-ResNet-34&Multi-task&LSTM&Pre-trained Word Embedding&Yes&CSMA&Classification&Cross-modal self-attention,\par Multi-task pre-training with extra data&&0.732\\
    CPRD~\cite{CPRD-MICCAI21}&ResNet-8&Contrastive&LSTM&Pre-trained Word Embedding&Yes&BAN&Classification&Constrastive Learning,\par Knowledge Distillation&&0.727\\
    MMBERT~\cite{MMBERT}&ResNet-152&ROCO&Transformer&Yes&Yes&Multi-head attention (Transformer)&Generation (Transformer)&Pretraining with extra data&&0.72\\
    QCR~\cite{zhan2020medical}&MEVF&None&LSTM&Not mentioned&Yes&BAN/SAN&Classification&Question-Conditioned Reasoning,\par Type-Conditioned Reasoning&&0.716\\
    MMQ~\cite{MMQ-MICCAI21}&MMQ&None&LSTM&Pre-trained Word Embedding&Yes&BAN/SAN&Classification&Multiple Meta-model Quantifying&&0.67\\
    MEVF~\cite{nguyen2019overcoming}&MEVF&None&LSTM&Not mentioned&Yes&BAN/SAN&Classification&Model-Agnostic Meta-Learning,\par Convolutional Denoising Auto-Encoder&&0.439/0.751 (0.627)\\
    HQS~\cite{GUPTA-HQS}&Inception-Resnet-v2&ImageNet&Bi-LSTM&Pre-trained Word Embedding&No&Concatenation&Classification&Question Segregation&0.411&\\
    \midrule
    \multicolumn{11}{l}{\textbf{PathVQA}}\\
    \midrule
    MedFuseNet~\cite{sharma2021medfusenet}&ResNet-152&ImageNet&LSTM&BERT Embedding&Yes&MedFuseNet&Classification/ Generation (LSTM)&Image Attention,\par Image-Question Co-Attention&0.605 (Subset)&0.636 (Subset)\\
    MMQ~\cite{MMQ-MICCAI21}&MMQ&None&LSTM&Pre-trained Word Embedding&Yes&BAN/SAN&Classification&Multiple Meta-model Quantifying&&0.488\\
    \midrule
    \multicolumn{11}{l}{\textbf{SLAKE}}\\
    \midrule
    CPRD~\cite{CPRD-MICCAI21}&ResNet-8&Contrastive&LSTM&Pre-trained Word Embedding&Yes&BAN&Classification&Constrastive Learning,\par Pre-training with extra data,\par Knowledge Distillation&&0.821\\
    \bottomrule
    \multicolumn{11}{l}{The ``*'' means the approach is not proposed during the public challenge period.}\\
\end{longtable}

\end{small}
\end{landscape}
\twocolumn

\textbf{Multi-modal pooling} is another important technique used for fusing visual and language features. The basic practice includes concatenation, sum, and element-wise product. In the reviewed papers, direct concatenation is the most widely used fusion method (10 of 46) but shows average performance. 
As the outer product may be computation-cost expensive when input vectors are with high dimensionality, researchers have proposed more efficient pooling methods. Multi-modal Compact Bilinear (MCB) pooling~\cite{fukui2016multimodal} is a typical method for pooling. It aggregates visual and text features by embedding the image and text features into higher-dimensional vectors and then convolving them with multiplications in the Fourier space for efficiency. The attention mechanism can also be used in the pooling module. Regarding the multi-modal pooling family, there are some other works such as the Multi-modal Factorized High-Order (MFH) pooling~\cite{yu2018beyond}, the Multi-modal Factorized Bilinear (MFB) pooling~\cite{yu2017multi}, e.t.c. The pooling with attention method is the second adopted method (8 of 46). Both the winners in VQA-Med-2018 and VQA-Med-2019 are using MFB pooling with attention solutions. 

Among the 46 approaches, the QC-MLB~\cite{Minh-QCMLB} and MedFuseNet~\cite{sharma2021medfusenet} are the only two works that proposed an innovative fusion algorithm. The QC-MLB uses a multi-glimpse attention mechanism to ensure that the question feature selects the proper image regions to answer. The MedFuseNet uses two attention modules: Image Attention and Image-Question Co-Attention to let the image feature and question feature interact twice. The QC-MLB and MedFuseNet all show performance improvement. However, the medical VQA research focusing on fusion schemes is still insufficient at this stage.

\subsection{Answering Component}
Among the 46 reviewed approaches, 33 approaches choose the classification mode for output, and 8 methods choose the generation mode. Some methods~\cite{al2019just,ren2020cgmvqa,bansal2019medical,zhou2019tua1,sharma2021medfusenet} use a switching strategy to adopt both classification and generation. The selection of the classification method or generation method reflects the length distribution of ground-truth answers belonging to a single phrase. The classification mode will have an advantage within a small answer space. However, it can become difficult if the answer candidates are longer and with more complex information, such as lesion descriptions. 

\subsection{Other Techniques}

Other than research in the basic components, some task-specific strategies are also applied to improve performance. Sub-task strategy is the approach that divides the overall task into several sub-tasks and assigns branch models. In these cases, an extra task classification module is applied to select the corresponding model by a particular question category or a type of image modality. It is often used~\cite{kornuta2019leveraging,al2019just,zhou2019tua1,turner2019lstm,shi2019deep,aiml2020mvqa,inception2020mvqa,hcp2020mvqa} especially in VQA-Med-2019 (questions are only in four categories) and VQA-Med-2020(questions are all in one category but in two types). The reason is that the questions have only distinct categories to be divided easily. These multiple model approaches may improve the single model effectiveness but lead to another risk that they may suffer from inappropriate model allocation due to the erroneous results from the sub-task classification module. Another problem is that these approaches cannot be generalized to other tasks if the question categories are different.

Another frequently applied technique is Global Average Pooling(5 of 46). Global Average Pooling~\cite{lin2013network} is using the average of feature maps to replace the last fully connected layers. It is believed to produce better image representation. Other techniques include Embedding-based Topic Model, Question-Conditioned Reasoning, and Image Size encoder.

\subsection{Performance Comparison}
As the performance shown in Table~\ref{table3}, the keys of state-of-the-art approaches on each dataset are quite different. Among the VQA-Med-2018 approaches, the top team Chakri~\cite{chakri} is the only team that uses generative output. For the VQA-Med-2019, the key factor is the language encoder model. The BERT users significantly achieve a higher performance than the LSTM users. In VQA-Med-2020 and VQA-Med-2021, the top teams AIML~\cite{aiml2020mvqa} and SYSU-HCP~\cite{gong2021sysu} all removed the language encoders and adopted ensemble strategy for image encoder, as the question categories were reduced. For VQA-RAD, the top three teams all have extra pre-training on their image encoders. The above comparison shows the enhancement of the image encoder is the essential ingredient of achieving state-of-the-art performance in most medical VQA datasets. 

Another finding is that over most datasets, the approaches with attention-based fusion algorithms are averagely outperforming the ones without attention. For example in VQA-Med-2019, the approaches with attention-based fusion have an average accuracy of 0.576, while the others have an average accuracy of 0.522. A similar conclusion can also be drawn for other datasets. It suggests that attention-based fusion algorithms are suitable for medical VQA datasets.
\subsection{Overall Discussion}
\label{sec_method_overall}
In the method survey, we reviewed 45 papers on medical VQA approaches, including 32 challenges work notes and 13 conference/journal papers. Although medical VQA research just started in 2018, there have already been various methods proposed and explored. Since the challenges only have restricted time for problem-solving, the work notes tend to make direct applications of well-verified deep learning models. They have a high proportion of using pre-trained components. On the other hand, they also develop some task-oriented techniques according to the intrinsic data property. Notably, there are two teams~\cite{hcp2020mvqa,eslami2021teams} observing the long-tailed distribution in their tasks and introducing the Bilateral-Branch Network (BBN)~\cite{zhou2020BBN} to deal with the imbalanced data. Despite insufficient innovations, the challenge teams' observations and strategies are valuable and inspiring.

The conference/journal papers explore further problem solutions. According to the 13 conference/journal papers, the current research is more focused on the image encoder than the other component, which is quite different from the general VQA research. Researchers introduce popular CV field ideas such as meta-learning and contrastive learning to enhance the image encoder. Pre-training is commonly applied in both the image encoder and the language decoder for answering. Auxiliary data is also a simple but efficient solution. Current research shows that acquiring a generalized image encoder is a high-priority and specific task in medical VQA research.

Another research topic that appeared in those technical papers is the generalizability of proposed methods. Among the 13 papers, only 5 papers are aware of showing their generalizability and evaluating their approaches on multiple datasets. And only one team~\cite{Minh-QCMLB} evaluate their approach not only on a medical VQA dataset but also on general VQA datasets. It will be a future standard that an approach should be evaluated on multiple datasets. 

Interpretability is also a demand in medical AI research. Among the 13 papers, 5 papers have illustrated the model visualization with techniques such as GradCAM and attention visualization. Although the SLAKE dataset has the annotation for semantic segmentation and object bounding box, none of them use the SLAKE to evaluate their visualization. 

\section{Medical VQA v.s. General VQA}
\label{sec_med_vs_gen}
Medical VQA and general VQA are two approaches that harness the power of visual content to answer questions but in distinct domains and contexts. While both aim to interpret and provide meaningful responses based on visual information, they differ significantly in their applications, objective, datasets, methods, domain knowledge, and evaluation. By understanding the unique characteristics and requirements of each approach, we can gain insights into their respective contributions and impact in their respective fields. In this section, we undertake a comparative analysis of medical VQA and general VQA across various dimensions.

\textbf{Application} Commonly, general VQA and medical VQA have potential applications of human-computer interaction such as image interpreting and education. The difference is the general VQA can be embedded into information retrieval such as search engines and virtual assistants. Also, it has potential applications in further human-computer interaction such as navigation and robotics. Medical VQA focuses on specific areas such as clinical decision support, telemedicine, and patient empowerment. Overall, the modes of interaction can be more diverse in general VQA applications.

\textbf{Objective} While both medical VQA and general VQA aim to bridge the gap between visual perception and natural language understanding, they differ in their functions and the specific information they focus on. medical VQA is designed to assist in medical diagnosis, treatment, and decision-making by leveraging visual medical data and providing accurate answers to medical-related questions. Hence, the objective of medical VQA systems is to gain ability in medical image understanding, abnormality locating, and terminology communication. 
On the other hand, the objective of general VQA is to enable machines to understand and respond to questions about general visual content, such as everyday images or videos. General VQA systems aim to comprehend the visual scene, recognize objects, infer relationships and attributes, and generate accurate answers in natural language.

\textbf{Datasets} As discussed in Sec.~\ref{data_discussion}, the medical VQA datasets have a smaller quantity because of the limit of data source and high knowledge requirement for annotation. Also, most of the current medical VQA datasets are less diverse in the question category. Besides, each dataset is designed with specific objectives in mind, which can vary depending on the dataset's focus and intended applications. The general VQA datasets have already focused on specific problems such as answer balance, while most of the medical VQA datasets are at the stage of expanding topics and application scenes.

\textbf{Method} The development of methods has been hindered by the complexity of the dataset. General VQA methods have expanded to encompass various topics, such as multi-task learning, logical reasoning, and interaction with the environment. In the general domain, the research on image and language understanding has primarily been undertaken by the upstream community, namely computer vision and NLP. General methods often rely on universal pre-trained models from the upstream community, with a primary focus on multi-modal fusion and reasoning. Conversely, the medical community lacks such universal resources, resulting in medical VQA works placing less emphasis on fusion algorithms and instead prioritizing image encoder pre-training as discussed in Sec.~\ref{sec_method_overall}.

\textbf{Domain knowledge} In general VQA, domain knowledge is instrumental in designing effective models and algorithms. For example, different strategies may be needed for object recognition, scene understanding, or reasoning about relationships between objects in images. In the case of medical VQA, domain knowledge is crucial for understanding the specific medical concepts, terminology, and context involved in the questions and images. Understanding the unique characteristics and challenges of medical imaging data, such as different modalities (e.g., X-ray, MRI), image quality variations, and anatomical complexities, is essential for developing robust and accurate medical VQA systems. It is challenging to integrate medical domain knowledge into medical VQA design.

\textbf{Evaluation} Due to the critical nature of medical information, medical VQA systems are expected to provide highly accurate and reliable answers. Evaluation metrics for Medical VQA might emphasize medical relevance, precision, and recall, considering the importance of accurate medical information. General VQA systems may prioritize a wider range of plausible answers rather than strictly aiming for accuracy. General VQA evaluation metrics may prioritize overall answer correctness, language understanding, and image understanding. Especially, medical datasets are naturally imbalanced distribution and long-tail distribution. Traditional metrics may not fully capture the performance and provide misleading results. The medical VQA can introduce metrics such as Average Precision (AP) for bias problems, while the general VQA pursues balanced distribution.

\section{Challenge and Future Works}

After reviewing medical VQA datasets and approaches, we identify some existing problems. Compared to the general domain VQA, the specific requirement and practical implementation scene also lead to unique and new challenges. Besides, the works on general-domain VQA provide some inspiration for future research direction.

\subsection{Question Diversity}
Question diversity is one of the most significant challenges of medical VQA. The VQA-RAD~\cite{lau2018dataset} investigates the natural questions in clinical conversation. The questions can be categorized into modality, plane, organ system, abnormality, object/condition presence, positional reasoning, color, size, attribute other, counting, and other. In other datasets such as VQA-Med-2019~\cite{abacha2019vqa} and PathVQA~\cite{he2020pathvqa}, the question categories tend to be less diverse than the VQA-RAD dataset. In the RadVisDial~\cite{kovaleva2020towards}, VQA-Med-2020~\cite{abacha2020overview}, and VQA-Med-2021~\cite{ImageCLEF-VQA-Med2021}, the question category is reduced to the abnormality presence only. However, most questions about an abnormality in existing datasets are about presence without further inquiry, like the location of tumors or tumor size. Therefore, questions remain to be added to diversify the medical VQA dataset.

To create better medical VQA and benefit clinical workflow integration, future research should be conducted about identifying the useful question categories in practical requirements. For example, the imaging modality and examined organ are totally not required because they are already in the study record information. This kind of information may be delivered to the end users outside the VQA system. First, the data source can be expanded. The synthesized QA pairs in current medical VQA datasets are usually created from image captions and medical reports. However, the information in captions and reports is restricted to specific topics. Therefore, more data sources, such as the textbook, should be considered to provide more textural corpus. Second, collecting real-world clinical conversations, especially with patients, will help researchers better understand the practical requirements. Third, the restriction that the question must be relevant to the image content should be broken. However, in a realistic clinical scene, the conversation content often exceeds the presented image content, for example, explaining the future risks of the abnormality or predicting disease progression. Overall, the question diversity is an essential consideration in dataset design to empower medical VQA with comprehensive coverage, real-world relevance, and user engagement.

Besides dataset design, question diversity raises a challenge in method development. To answer the diverse questions, the medical VQA systems require various reasoning abilities. For example, as a sample shown in Table~\ref{table2}, to answer question \textit{``What is the function of the rightmost organ in this picture?''}, the model should understand the region described, identify the organ in the region of interest, and finally answer the function of the organ. Besides the basic image and language understanding, medical domain knowledge is a critical ability required, which includes knowledge of anatomical structures, medical procedures, diseases, medical imaging modalities, treatment options, and clinical practices. More specially, for the question categories in Table~\ref{table1}, \textit{Modality, Plane} need knowledge about radiology examination; \textit{Organ, System, Abnormality, Object/Condition Presence} need knowledge about human anatomy and medicine; \textit{Positional Reasoning, Color, Size, Attribute Other, Counting} need general knowledge and reasoning; \textit{Knowledge Graph} need to combine the upon knowledge with the knowledge triplets given in dataset. Therefore, to address question categories correspondingly, the future medical VQA should be equipped with computer-aid diagnosis, general language understanding, reasoning, knowledge integration, and contextual understanding.

Finally, the evaluation of the model performance should also consider the question of diversity. The imbalanced question category distribution can provide misleading results under overall evaluation. To mitigate the impact of imbalanced data on overall evaluation, it is important to consider specialized evaluation strategies. This can involve using category-specific metrics or weighted overall metrics. Furthermore, incorporating language-based evaluation metrics will be a challenge with the presence of verbose answers. Both the correctness of the answers and the quality of the language used should be evaluated to encourage a more comprehensive medical VQA system.

\subsection{Integrating the Extra Medical Information}

Another challenge for the medical VQA is to integrate the extra information into the inference procedure. For example, \citeauthor{kovaleva2020towards} found that incorporating the medical history of the patient leads to better performance in answering questions~\cite{kovaleva2020towards}.
\subsubsection{EHR}
An electronic health record~(EHR) contains a patient's medical history, diagnoses, medications, treatment plans, etc. In the medical AI domain, the EHR has been proven useful for disease prediction~\cite{deepehr}. Hence it should also be helpful in medical VQA. To support the research of EHR in medical VQA, a new dataset containing the original EHR or synthetical EHR is required. Furthermore, the VQA model should also be modified because the EHR contains a lot of metric variables that should be treated as numbers other than natural language. As far as we know, no previous research has investigated combining EHR and computer vision. Using numeric variables as VQA input and the corresponding fusion algorithm also has not been researched. Therefore encoding EHR in VQA is meaningful and worth studying.
\subsubsection{Multiple Images}
In the medical scene, especially radiology, it is general that the study is based on multiple images other than a single image. Multiple images can contain different scan planes and sequential slices to support the decision-making of medical professionals. For example, in MIMIC-CXR~\cite{johnson2019mimiccxrjpg}, a radiology report is often associated with two images of postero-anterior view and lateral view. However, in RadVisDial~\cite{kovaleva2020towards}, the authors keep only the postero-anterior view, although the abandoned images are also informative. 
The reason may be that the common VQA or VisDial models only support single-image input. Therefore, the VQA datasets with multiple input images are required to support the model research. Future research should also be devoted to developing the VQA model by taking multiple images as input.
 
\subsection{Interpretability and Reliability}
Interpretability is a long-standing problem of deep learning. As the VQA models are commonly based on deep learning methods, the medical VQA has to deal with interpretability. To the medical VQA system, interpretability determines the reliability of the predicted answer, and it is more important than that in the general domain because the wrong decision may lead to catastrophic consequences. The general-domain VQA researchers have addressed this problem and investigated several directions to evaluate the inference ability of a model.
\subsubsection{Unimodal Bias of VQA Models} 
In the VQA field, the unimodal bias means that the model may answer the question based on statistical regularities from one modality without considering the other modality. Particularly, the bias to language input is also called language prior. The researchers have noticed bias since the first VQA dataset was proposed. They tested the question-only model ``LSTM Q'', which has only a slightly lower performance compared with the standard model ``LSTM Q+I''~\cite{antol2015vqa} and indicates the language prior. To better measure the effect of language prior problem of VQA models, \citeauthor{agrawal2018don} presented new splits of the VQA v1 and VQA v2 datasets with changing priors~(respectively VQA-CP v1 and VQA-CP v2). They also proposed a Grounded Visual Question Answering model~(GVQA) to prevent the model from ``cheating'' by primarily relying on priors~\cite{agrawal2018don}. Alternatively, \citeauthor{ramakrishnan2018overcoming} proposed a novel regularization scheme that poses training as an adversarial game between the VQA model and a question-only model to discourage the VQA model from capturing language biases~\cite{ramakrishnan2018overcoming}. \citeauthor{cadene2019rubi} proposed the RUBi to reduce biases in VQA models. It minimizes the importance of the most biased examples and implicitly forces the VQA model to use the two input modalities instead of relying on statistical regularities between the question and the answer~\cite{cadene2019rubi}. 

The unimodal bias in general-domain VQA has been discussed in many aspects, such as benchmark and training schemes. These provide a good starting point for further work in medical VQA. Benchmarks similar to VQA-CP are required to investigate the unimodal bias problem of models. As the quantity of medical VQA datasets is not so large as general-domain VQA datasets, the solution for unimodal bias also needs research.

\subsubsection{External Knowledge}
In VQA research, external knowledge means the model may need to incorporate an external knowledge base to infer the answer besides the image and the question representation. In general-domain VQA, an ``adult-level common sense'' is required to support the cognition and inference in question answering~\cite{wu2017visual}. \citeauthor{wang2015explicit} proposed an approach named ``Ahab'' to provide explicit knowledge-base reasoning~\cite{wang2015explicit}. They also provide a dataset KB-VQA and a protocol to evaluate the knowledge-based methods. Furthermore, \citeauthor{wang2018fvqa} proposed the FVQA dataset providing a supporting fact that is critical for answering each visual questions~\cite{wang2018fvqa}. On the other hand, to test the VQA methods' ability to retrieve relevant facts without a structured knowledge base, \citeauthor{marino2019ok} proposed the OK-VQA dataset, including only questions that require external knowledge resources. They also proposed a knowledge-based baseline named ``ArticleNet'', which retrieved some articles from Wikipedia for each question image pair and then trained a network to find the answer in the retrieved articles. 

The challenge in medical VQA is that the knowledge requirement is much higher than ``adult-level common sense''. The future research includes identifying questions that required external knowledge, exploiting the existing medical knowledge base such as \cite{muller2019open}, and building a medical VQA dataset with a structured or unstructured knowledge base. There are also additional studies required to determine whether the medical knowledge-based VQA needs specific approaches.

Moreover, external knowledge can also expand the function of medical VQA in clinic workflow. In the practical environment, the questions from the patient can start from image findings and move further to topics like possible disease, potential risk, and disease control. To answer the questions of those topics, an external knowledge base is required. A medical VQA system equipped with external knowledge will have a strong topic capability and become a powerful tool in clinic workflow.
\subsubsection{Evidence Verification}
Evidence verification is a common method to verify model interpretability. In the medical domain, illustrating evidence of the abnormality will gain the trust of medical professionals and patients. A typical method to collect model evidence is through visualization, such as feature attribution or saliency map. In terms of the multi-modal models with attention mechanism, an attention map can be visualized to show the image regions that lead to the answer~\cite{yang2015stacked}. By comparing the attention map with annotated human visual attention, the researchers can verify whether the VQA methods use the correct evidence to get the answer. In general-domain VQA, \citeauthor{vqahat} proposed the VQA-HAT dataset containing annotated human visual attention to evaluate the attention maps both qualitatively~(via visualizations) and quantitatively~(via rank-order correlation)~\cite{vqahat}. To extend the verification to textual evidence, \citeauthor{huk2018multimodal} proposed the VQA-X dataset containing human-annotated both textual explanations and visual explanations~\cite{huk2018multimodal}. To address text-based question-answer pairs in the VQA task, \citeauthor{Wang_2020_CVPR} proposed the EST-VQA dataset annotating the bounding boxes of correct text clues~\cite{Wang_2020_CVPR}. Alternatively, \citeauthor{Jiang_2020_CVPR} proposed the IQVA dataset annotating the eye-tracking data as human visual attention~\cite{Jiang_2020_CVPR}. 
 
Although those studies have established a complete scheme, including the annotated evidence, reasoning model, and evaluation metrics, several problems are to be explored when applying evidence verification in medical VQA. First, the previous studies showed different annotation modes such as bounding box and eye-tracking. It is unknown what is the most suitable mode for medical VQA. For some question categories, such as the modality, the answer evidence can be hard to explain with a certain region and may need a text-based explanation. Second, manual annotation for a medical task is expensive because it requires the professional skill of radiologists. In this situation, the existing medical multi-task datasets such as~\cite{DBW86T_2018} can be beneficial by providing comprehensive knowledge of medical terms. Therefore, the evidence verification approaches for medical VQA will need to consider generating a text-based explanation and utilizing the existing annotated medical datasets.
\subsubsection{Summarization}
These three directions examine the different reasoning procedures of VQA models and provide performance evaluation methods. The VQA models have to look into the image, learn the extra knowledge, and find the correct evidence to pass the ``exams''. To our best knowledge, there are only a few works considering interpretability in medical VQA, such as the SLAKE providing lesion annotation and the CPRD, MedFuseNet providing visualization. However, there is no work really performing quantified measurement of interpretability. Some currently feasible ways include giving the attribution of images against problems and using the SLAKE dataset to calculate the overlapping rate between saliency maps and the annotations. It remains to be done building more quantified benchmarks to evaluate medical VQA interpretability.

If the medical VQA system can be verified with its inference ability, it will become a more convincing and reliable tool. It is also helpful to present the knowledge and evidence used in inference explicitly. Hence the evaluation of inference ability should be regarded as a more important benchmark than the answer accuracy. Hopefully, the medical VQA will answer the question ``why'' in the nearest future. 
\subsection{Generalizability}
Generalizability is a universal topic in medical AI research and an inevitable issue for applications running in practical scenarios. The cause of the Generalizability problem is the practical input can be out of the distribution (OOD) of the training data. The factors can be various, such as the patient race and imaging devices. The gap between different data distributions is named \textit{domain shift}. For downstream tasks like VQA, the generalizability problem is two-fold and more comprehensive. Firstly, as discussed in Sec.~\ref{sec:method}, the VQA models usually consist of several sub-models which may have a pre-trained weight. At this stage, a domain shift between pre-train data and current training data is introduced. Secondly, after a VQA model is developed and deployed, there will be a domain shift between training data and practical data, which is usually evaluated by cross-dataset validation. 

Among the reviewed methods, several approaches have considered the domain shift in sub-model pre-training and acquiring medical pre-trained models as image encoders~\cite{CPRD-MICCAI21}. However, there is no study considering the potential domain shift in language encoders. Also, the domain shift crossing medical VQA datasets has not yet been considered and studied. With the growing number of medical VQA datasets works, there has been sufficient material for transferring learning studies such as domain adaptation and domain generalization. Measuring and improving model generalizability will be a feasible and meaningful research topic.
\subsection{Large Language Models}
Large Language Models (LLMs) are highly complex artificial intelligence systems that have the capability to learn from the vast amounts of available text data~\cite{radford2018improving}. Discriminative LLMs (e.g., BERT) have been well-studied for answering questions in a classification manner. However, the generative capability of LLMs enables them to answer diverse questions more flexibly and effectively with human-like responses. Generative LLMs have already started to show their potential and remarkable results in the field of the medical domain, where models like Open AI's GPT-3~\cite{brown2020language}, GPT-4~\cite{bubeck2023sparks} have successfully passed a part of the US medical licensing exam~\cite{nori2023capabilities}. Thus, it is interesting to explore the potential approaches of how these LLMs could be of aid in the task of VQA. As the task of medical VQA involves understanding the spatial relationships in a medical image, it is challenging for LLMs that have been primarily trained on text data. However, LLMs have the potential to be integrated into improving the QA capability of a model. Below we describe certain directions of how LLMs could be beneficial to the task of medical VQA.

As a straightforward way, LLMs can be used to process the questions and generate responses based on the extracted features from the images provided by an image-based model. For example in the recent ChatCAD framework~\cite{wang2023chatcad}, the authors converted the outputs provided by different image-based models (classification, segmentation) into textual information and provided it to ChatGPT~\cite{shao2023prompting} for refinement and understanding. Then, follow-up queries were asked by the authors about the disease condition of the patient and ChatGPT provided comprehensive answers by combining the given outputs from the image-based models and its learned knowledge from the huge corpus of data. Another recent framework Prophet~\cite{yang2022language} improved the answer generation capability of a trained VQA model by encoding answer heuristics generated by the VQA model into a prompt for GPT-3 to better comprehend the task thus making better use of the potential of GPT-3. Prophet framework was able to achieve a new state-of-the-art performance on two challenging knowledge-based VQA datasets. Some works~\cite{oikarinenlabel} have also employed LLMs to improve the interpretability of the existing image-based models by generating specific attributes that could be analyzed in an image. In addition to the above directions, LLMs can also adapt to multiple modalities, such as patient history, demographics, lab results, etc. which if combined with the current medical VQA models could lead to a more comprehensive analysis and better answering of the related questions.

Although LLMs have a huge potential to be integrated into the task of medical VQA, there are some associated potential pitfalls. First, although LLMs are trained on vast amounts of text corpus data, they might not possess the same domain-specific knowledge expertise as medical professionals in specific domains. Second, LLMs have learned biases or misinformation inherently present in their training data, which could lead to incorrect conclusions. Third, LLMs are not able to express uncertainty for their answers, and thus LLMs could confidently generate incorrect responses or hallucinations~\cite{nori2023capabilities}, which could be fatal in medical decision-making. So, careful attention and high-level collaboration are required to mitigate the above issues. Specifically, involving medical experts in the framework development and fine-tuning process can help to tailor the model responses corresponding to specific medical domains. Moreover, diverse, accurate, and representative real-world medical training data could help to minimize the above biases and generate factually correct responses.
\subsection{Integration in Medical Workflow}
The community has a long history of trying to integrate AI decision-supporting systems into the clinical flow. Integrating the medical VQA system into clinical practice will provide efficient communication and serve as an effective assistant.
\citeauthor{hekler2019superior} found that the combination of human and artificial intelligence can achieve superior results over the independent results of both of these systems in skin cancer classification~\cite{hekler2019superior}. 
Furthermore, \citeauthor{tschandl2020human} found that physicians with AI-based support outperformed either AI or physicians. Meanwhile, the least experienced clinicians gain the most from AI-based support~\cite{tschandl2020human}. \citeauthor{tschandl2020human} also experimented with the accuracy improvements of human-computer collaboration in skin cancer recognition and found multiclass probabilities outperformed either content-based image retrival~(CBIR) or malignancy probability in the mobile technolog~\cite{tschandl2020human}. These findings indicate that many factors must be considered in developing successful outcomes for medical AI-supporting system integration, such as clinicians' cognitive style, cognitive error, personality, experience, and acceptance of AI. 

Moreover, most AI decision support systems used in the clinical setting are good at addressing prepared  questions only. In contrast, the advantage of medical VQA is that it understands free-form questions and provides real-time communication. To address this advantage, the workflow can remove redundant querying and get simplified. For example, the presence of abnormality may be always asked, and it can be given before the QA session. Hence, selecting useful question categories and keeping online learning may help to maintain a functional question database. The ultimate goal of a medical VQA system is answering open-ended questions~(what, which, e.t.c.).   
This goal will introduce more complexity and uncontrolled factors under the ``human-in-the-loop'' hypothesis. For example, the medical VQA system may need to resolve a dispute when its opinion is different from the clinician's. Hence, the training corpus may also prepare for a negotiation. More efforts and studies need to be conducted in this field for a successful deployment of medical VQA into the clinical pipeline. Improving workflow efficiency and service quality is also required to study multiple aspects, such as time spent, collaborative answer accuracy, and user satisfaction.

\section{Conclusion}
This article presented a survey of the datasets and approaches of medical VQA. We collect information about 8 medical VQA datasets and 45 papers describing medical VQA approaches. We conduct a comprehensive discussion on dataset creation, approach framework, approach components, and corresponding techniques. In addition to the descriptive review, we identified some challenges worth exploring in future research. The medical VQA system mainly faces the following four challenges: first, how to get the system answers a range of more comprehensive question categories; second, how to combine medical features with the task; third, how to verify the evidence of an answer to make it more convincing; forth, how to make the system not bias to any modality; finally, how to maximize the benefit of the medical VQA in the workflow. For future work, we propose the following directions. To investigate potential implementation in a real-world scene is necessary. Furthermore, we should pay attention to the conversation between medical professionals and non-professionals. The advantage of VQA is to understand natural language questions and help non-professionals. We should also introduce meaningful ideas in the general domain, such as evidence verification, bias analysis, external knowledge database, etc. These will help us build a practicable and convincing medical VQA system toward medical AI's ultimate goal.

\section{Declaration of Interests}
We declare that we have no known competing financial interests or personal relationships that could have appeared to influence the work reported in this paper.

\bibliographystyle{cas-model2-names}
\bibliography{main}

\begin{thebibliography}{112}
\expandafter\ifx\csname natexlab\endcsname\relax\def\natexlab#1{#1}\fi
\providecommand{\url}[1]{\texttt{#1}}
\providecommand{\href}[2]{#2}
\providecommand{\path}[1]{#1}
\providecommand{\DOIprefix}{doi:}
\providecommand{\ArXivprefix}{arXiv:}
\providecommand{\URLprefix}{URL: }
\providecommand{\Pubmedprefix}{pmid:}
\providecommand{\doi}[1]{\href{http://dx.doi.org/#1}{\path{#1}}}
\providecommand{\Pubmed}[1]{\href{pmid:#1}{\path{#1}}}
\providecommand{\bibinfo}[2]{#2}
\ifx\xfnm\relax \def\xfnm[#1]{\unskip,\space#1}\fi
\bibitem[{Abacha et~al.(2018)Abacha, Gayen, Lau, Rajaraman and
  Demner-Fushman}]{abacha2018nlm}
\bibinfo{author}{Abacha, A.B.}, \bibinfo{author}{Gayen, S.},
  \bibinfo{author}{Lau, J.J.}, \bibinfo{author}{Rajaraman, S.},
  \bibinfo{author}{Demner-Fushman, D.}, \bibinfo{year}{2018}.
\newblock \bibinfo{title}{{NLM} at {ImageCLEF} 2018 visual question answering
  in the medical domain.}, in: \bibinfo{booktitle}{CLEF (Working Notes)}.
\bibitem[{Agrawal et~al.(2018)Agrawal, Batra, Parikh and
  Kembhavi}]{agrawal2018don}
\bibinfo{author}{Agrawal, A.}, \bibinfo{author}{Batra, D.},
  \bibinfo{author}{Parikh, D.}, \bibinfo{author}{Kembhavi, A.},
  \bibinfo{year}{2018}.
\newblock \bibinfo{title}{Don't just assume; look and answer: Overcoming priors
  for visual question answering}, in: \bibinfo{booktitle}{2018 IEEE/CVF
  Conference on Computer Vision and Pattern Recognition (CVPR)},
  \bibinfo{publisher}{IEEE Computer Society}, \bibinfo{address}{Los Alamitos,
  CA, USA}. pp. \bibinfo{pages}{4971--4980}.
\bibitem[{Al-Sadi et~al.(2020)Al-Sadi, Al-Theiabat and
  Al-Ayyoub}]{inception2020mvqa}
\bibinfo{author}{Al-Sadi, A.}, \bibinfo{author}{Al-Theiabat, H.},
  \bibinfo{author}{Al-Ayyoub, M.}, \bibinfo{year}{2020}.
\newblock \bibinfo{title}{The inception team at {VQA-Med} 2020: Pretrained
  {VGG} with data augmentation for medical vqa and vqg}, in:
  \bibinfo{booktitle}{CLEF 2020 Working Notes}.
\bibitem[{Al-Sadi et~al.(2019)Al-Sadi, Talafha, Al-Ayyoub, Jararweh and
  Costen}]{al2019just}
\bibinfo{author}{Al-Sadi, A.}, \bibinfo{author}{Talafha, B.},
  \bibinfo{author}{Al-Ayyoub, M.}, \bibinfo{author}{Jararweh, Y.},
  \bibinfo{author}{Costen, F.}, \bibinfo{year}{2019}.
\newblock \bibinfo{title}{{JUST} at {ImageCLEF} 2019 visual question answering
  in the medical domain.}, in: \bibinfo{booktitle}{CLEF (Working Notes)}.
\bibitem[{Allaouzi and Ahmed(2018)}]{18fstt}
\bibinfo{author}{Allaouzi, I.}, \bibinfo{author}{Ahmed, M.B.},
  \bibinfo{year}{2018}.
\newblock \bibinfo{title}{Deep neural networks and decision tree classifier for
  visual question answering in the medical domain.}, in:
  \bibinfo{booktitle}{CLEF (Working Notes)}.
\bibitem[{Allaouzi et~al.(2019)Allaouzi, Ahmed and
  Benamrou}]{allaouzi2019encoder}
\bibinfo{author}{Allaouzi, I.}, \bibinfo{author}{Ahmed, M.B.},
  \bibinfo{author}{Benamrou, B.}, \bibinfo{year}{2019}.
\newblock \bibinfo{title}{An encoder-decoder model for visual question
  answering in the medical domain.}, in: \bibinfo{booktitle}{CLEF (Working
  Notes)}.
\bibitem[{Ambati and Reddy~Dudyala(2018)}]{chakri}
\bibinfo{author}{Ambati, R.}, \bibinfo{author}{Reddy~Dudyala, C.},
  \bibinfo{year}{2018}.
\newblock \bibinfo{title}{A sequence-to-sequence model approach for imageclef
  2018 medical domain visual question answering}, in: \bibinfo{booktitle}{2018
  15th IEEE India Council International Conference (INDICON)}, pp.
  \bibinfo{pages}{1--6}.
\newblock \DOIprefix\doi{10.1109/INDICON45594.2018.8987108}.
\bibitem[{Anderson et~al.(2018)Anderson, He, Buehler, Teney, Johnson, Gould and
  Zhang}]{anderson2018bottom}
\bibinfo{author}{Anderson, P.}, \bibinfo{author}{He, X.},
  \bibinfo{author}{Buehler, C.}, \bibinfo{author}{Teney, D.},
  \bibinfo{author}{Johnson, M.}, \bibinfo{author}{Gould, S.},
  \bibinfo{author}{Zhang, L.}, \bibinfo{year}{2018}.
\newblock \bibinfo{title}{Bottom-up and top-down attention for image captioning
  and visual question answering}, in: \bibinfo{booktitle}{2018 IEEE/CVF
  Conference on Computer Vision and Pattern Recognition (CVPR)},
  \bibinfo{publisher}{IEEE Computer Society}, \bibinfo{address}{Los Alamitos,
  CA, USA}. pp. \bibinfo{pages}{6077--6086}.
\bibitem[{Andreas et~al.(2016)Andreas, Rohrbach, Darrell and
  Klein}]{Andreas_2016_CVPR}
\bibinfo{author}{Andreas, J.}, \bibinfo{author}{Rohrbach, M.},
  \bibinfo{author}{Darrell, T.}, \bibinfo{author}{Klein, D.},
  \bibinfo{year}{2016}.
\newblock \bibinfo{title}{Neural module networks}, in: \bibinfo{booktitle}{2016
  IEEE Conference on Computer Vision and Pattern Recognition (CVPR)},
  \bibinfo{publisher}{IEEE Computer Society}, \bibinfo{address}{Los Alamitos,
  CA, USA}. pp. \bibinfo{pages}{39--48}.
\bibitem[{Antol et~al.(2015)Antol, Agrawal, Lu, Mitchell, Batra, Zitnick and
  Parikh}]{antol2015vqa}
\bibinfo{author}{Antol, S.}, \bibinfo{author}{Agrawal, A.},
  \bibinfo{author}{Lu, J.}, \bibinfo{author}{Mitchell, M.},
  \bibinfo{author}{Batra, D.}, \bibinfo{author}{Zitnick, C.},
  \bibinfo{author}{Parikh, D.}, \bibinfo{year}{2015}.
\newblock \bibinfo{title}{{VQA}: Visual question answering}, in:
  \bibinfo{booktitle}{2015 IEEE International Conference on Computer Vision
  (ICCV)}, \bibinfo{publisher}{IEEE Computer Society}, \bibinfo{address}{Los
  Alamitos, CA, USA}. pp. \bibinfo{pages}{2425--2433}.
\bibitem[{Bai et~al.(2021)Bai, Shan, Huang and Wang}]{MVQAS}
\bibinfo{author}{Bai, H.}, \bibinfo{author}{Shan, X.}, \bibinfo{author}{Huang,
  Y.}, \bibinfo{author}{Wang, X.}, \bibinfo{year}{2021}.
\newblock \bibinfo{title}{{MVQAS}: A Medical Visual Question Answering System}.
  \bibinfo{publisher}{Association for Computing Machinery},
  \bibinfo{address}{New York, NY, USA}.
\newblock p. \bibinfo{pages}{4675–4679}.
\bibitem[{Bansal et~al.(2019)Bansal, Gadgil, Shah and
  Verma}]{bansal2019medical}
\bibinfo{author}{Bansal, M.}, \bibinfo{author}{Gadgil, T.},
  \bibinfo{author}{Shah, R.}, \bibinfo{author}{Verma, P.},
  \bibinfo{year}{2019}.
\newblock \bibinfo{title}{Medical visual question answering at {Image CLEF
  2019-VQA Med}}, in: \bibinfo{booktitle}{CLEF (Working Notes)}.
\bibitem[{{Ben Abacha} et~al.(2020){Ben Abacha}, Datla, Hasan, Demner-Fushman
  and M\"uller}]{abacha2020overview}
\bibinfo{author}{{Ben Abacha}, A.}, \bibinfo{author}{Datla, V.V.},
  \bibinfo{author}{Hasan, S.A.}, \bibinfo{author}{Demner-Fushman, D.},
  \bibinfo{author}{M\"uller, H.}, \bibinfo{year}{2020}.
\newblock \bibinfo{title}{Overview of the {VQA-Med} task at {ImageCLEF} 2020:
  Visual question answering and generation in the medical domain}, in:
  \bibinfo{booktitle}{CLEF 2020 Working Notes},
  \bibinfo{publisher}{CEUR-WS.org}, \bibinfo{address}{Thessaloniki, Greece}.
\bibitem[{{Ben Abacha} et~al.(2019){Ben Abacha}, Hasan, Datla, Liu,
  Demner-Fushman and M\"uller}]{abacha2019vqa}
\bibinfo{author}{{Ben Abacha}, A.}, \bibinfo{author}{Hasan, S.A.},
  \bibinfo{author}{Datla, V.V.}, \bibinfo{author}{Liu, J.},
  \bibinfo{author}{Demner-Fushman, D.}, \bibinfo{author}{M\"uller, H.},
  \bibinfo{year}{2019}.
\newblock \bibinfo{title}{{VQA-Med}: Overview of the medical visual question
  answering task at imageclef 2019}, in: \bibinfo{booktitle}{CLEF2019 Working
  Notes}, \bibinfo{publisher}{CEUR-WS.org}, \bibinfo{address}{Lugano,
  Switzerland}.
\bibitem[{{Ben Abacha} et~al.(2021){Ben Abacha}, Sarrouti, Demner-Fushman,
  Hasan and M\"uller}]{ImageCLEF-VQA-Med2021}
\bibinfo{author}{{Ben Abacha}, A.}, \bibinfo{author}{Sarrouti, M.},
  \bibinfo{author}{Demner-Fushman, D.}, \bibinfo{author}{Hasan, S.A.},
  \bibinfo{author}{M\"uller, H.}, \bibinfo{year}{2021}.
\newblock \bibinfo{title}{Overview of the {VQA-Med} task at {ImageCLEF} 2021:
  Visual question answering and generation in the medical domain}, in:
  \bibinfo{booktitle}{CLEF 2021 Working Notes},
  \bibinfo{publisher}{CEUR-WS.org}, \bibinfo{address}{Bucharest, Romania}.
\bibitem[{Bounaama and Abderrahim(2019)}]{task2019tlemcen}
\bibinfo{author}{Bounaama, R.}, \bibinfo{author}{Abderrahim, M.E.A.},
  \bibinfo{year}{2019}.
\newblock \bibinfo{title}{{Tlemcen University} at {ImageCLEF} 2019 visual
  question answering task.}, in: \bibinfo{booktitle}{CLEF (Working Notes)}.
\bibitem[{Brown et~al.(2020)Brown, Mann, Ryder, Subbiah, Kaplan, Dhariwal,
  Neelakantan, Shyam, Sastry, Askell et~al.}]{brown2020language}
\bibinfo{author}{Brown, T.}, \bibinfo{author}{Mann, B.},
  \bibinfo{author}{Ryder, N.}, \bibinfo{author}{Subbiah, M.},
  \bibinfo{author}{Kaplan, J.D.}, \bibinfo{author}{Dhariwal, P.},
  \bibinfo{author}{Neelakantan, A.}, \bibinfo{author}{Shyam, P.},
  \bibinfo{author}{Sastry, G.}, \bibinfo{author}{Askell, A.}, et~al.,
  \bibinfo{year}{2020}.
\newblock \bibinfo{title}{Language models are few-shot learners}.
\newblock \bibinfo{journal}{Advances in neural information processing systems}
  \bibinfo{volume}{33}, \bibinfo{pages}{1877--1901}.
\bibitem[{Bubeck et~al.(2023)Bubeck, Chandrasekaran, Eldan, Gehrke, Horvitz,
  Kamar, Lee, Lee, Li, Lundberg et~al.}]{bubeck2023sparks}
\bibinfo{author}{Bubeck, S.}, \bibinfo{author}{Chandrasekaran, V.},
  \bibinfo{author}{Eldan, R.}, \bibinfo{author}{Gehrke, J.},
  \bibinfo{author}{Horvitz, E.}, \bibinfo{author}{Kamar, E.},
  \bibinfo{author}{Lee, P.}, \bibinfo{author}{Lee, Y.T.}, \bibinfo{author}{Li,
  Y.}, \bibinfo{author}{Lundberg, S.}, et~al., \bibinfo{year}{2023}.
\newblock \bibinfo{title}{Sparks of artificial general intelligence: Early
  experiments with gpt-4}.
\newblock \bibinfo{journal}{arXiv preprint arXiv:2303.12712} .
\bibitem[{Cadene et~al.(2019)Cadene, Dancette, Cord, Parikh
  et~al.}]{cadene2019rubi}
\bibinfo{author}{Cadene, R.}, \bibinfo{author}{Dancette, C.},
  \bibinfo{author}{Cord, M.}, \bibinfo{author}{Parikh, D.}, et~al.,
  \bibinfo{year}{2019}.
\newblock \bibinfo{title}{{RUBi}: Reducing unimodal biases for visual question
  answering}.
\newblock \bibinfo{journal}{Advances in Neural Information Processing Systems}
  \bibinfo{volume}{32}, \bibinfo{pages}{841--852}.
\bibitem[{Chen et~al.(2020)Chen, Gong and Li}]{hcp2020mvqa}
\bibinfo{author}{Chen, G.}, \bibinfo{author}{Gong, H.}, \bibinfo{author}{Li,
  G.}, \bibinfo{year}{2020}.
\newblock \bibinfo{title}{{HCP-MIC} at {VQA-Med} 2020: Effective visual
  representation for medical visual question answering}, in:
  \bibinfo{booktitle}{CLEF 2020 Working Notes}.
\bibitem[{Cho et~al.(2014)Cho, van Merri{\"e}nboer, Gulcehre, Bahdanau,
  Bougares, Schwenk and Bengio}]{GRU}
\bibinfo{author}{Cho, K.}, \bibinfo{author}{van Merri{\"e}nboer, B.},
  \bibinfo{author}{Gulcehre, C.}, \bibinfo{author}{Bahdanau, D.},
  \bibinfo{author}{Bougares, F.}, \bibinfo{author}{Schwenk, H.},
  \bibinfo{author}{Bengio, Y.}, \bibinfo{year}{2014}.
\newblock \bibinfo{title}{Learning phrase representations using {RNN}
  encoder{--}decoder for statistical machine translation}, in:
  \bibinfo{booktitle}{Proceedings of the 2014 Conference on Empirical Methods
  in Natural Language Processing ({EMNLP})}, \bibinfo{publisher}{Association
  for Computational Linguistics}, \bibinfo{address}{Doha, Qatar}. pp.
  \bibinfo{pages}{1724--1734}.
\newblock \DOIprefix\doi{10.3115/v1/D14-1179}.
\bibitem[{Cross et~al.(2020)Cross, Wildenberg, Liao, Novak, Bevilacqua, Chen,
  Siegelman and Cook}]{cross2020voice}
\bibinfo{author}{Cross, N.M.}, \bibinfo{author}{Wildenberg, J.},
  \bibinfo{author}{Liao, G.}, \bibinfo{author}{Novak, S.},
  \bibinfo{author}{Bevilacqua, T.}, \bibinfo{author}{Chen, J.},
  \bibinfo{author}{Siegelman, E.}, \bibinfo{author}{Cook, T.S.},
  \bibinfo{year}{2020}.
\newblock \bibinfo{title}{The voice of the radiologist: Enabling patients to
  speak directly to radiologists}.
\newblock \bibinfo{journal}{Clinical imaging} \bibinfo{volume}{61},
  \bibinfo{pages}{84--89}.
\bibitem[{Das et~al.(2016)Das, Agrawal, Zitnick, Parikh and Batra}]{vqahat}
\bibinfo{author}{Das, A.}, \bibinfo{author}{Agrawal, H.},
  \bibinfo{author}{Zitnick, C.L.}, \bibinfo{author}{Parikh, D.},
  \bibinfo{author}{Batra, D.}, \bibinfo{year}{2016}.
\newblock \bibinfo{title}{Human attention in visual question answering: Do
  humans and deep networks look at the same regions?}, in:
  \bibinfo{booktitle}{Conference on Empirical Methods in Natural Language
  Processing}.
\bibitem[{Devlin et~al.(2019)Devlin, Chang, Lee and Toutanova}]{devlin2019bert}
\bibinfo{author}{Devlin, J.}, \bibinfo{author}{Chang, M.W.},
  \bibinfo{author}{Lee, K.}, \bibinfo{author}{Toutanova, K.},
  \bibinfo{year}{2019}.
\newblock \bibinfo{title}{{BERT}: Pre-training of deep bidirectional
  transformers for language understanding}, in: \bibinfo{booktitle}{Proceedings
  of the 2019 Conference of the North American Chapter of the Association for
  Computational Linguistics: Human Language Technologies, Volume 1 (Long and
  Short Papers)}, pp. \bibinfo{pages}{4171--4186}.
\bibitem[{Do et~al.(2021)Do, Nguyen, Tjiputra, Tran, Tran and
  Nguyen}]{MMQ-MICCAI21}
\bibinfo{author}{Do, T.}, \bibinfo{author}{Nguyen, B.X.},
  \bibinfo{author}{Tjiputra, E.}, \bibinfo{author}{Tran, M.},
  \bibinfo{author}{Tran, Q.D.}, \bibinfo{author}{Nguyen, A.},
  \bibinfo{year}{2021}.
\newblock \bibinfo{title}{Multiple meta-model quantifying for medical visual
  question answering}, in: \bibinfo{editor}{de~Bruijne, M.},
  \bibinfo{editor}{Cattin, P.C.}, \bibinfo{editor}{Cotin, S.},
  \bibinfo{editor}{Padoy, N.}, \bibinfo{editor}{Speidel, S.},
  \bibinfo{editor}{Zheng, Y.}, \bibinfo{editor}{Essert, C.} (Eds.),
  \bibinfo{booktitle}{Medical Image Computing and Computer Assisted
  Intervention -- MICCAI 2021}, \bibinfo{publisher}{Springer International
  Publishing}, \bibinfo{address}{Cham}. pp. \bibinfo{pages}{64--74}.
\bibitem[{Eslami et~al.(2021)Eslami, de~Melo and Meinel}]{eslami2021teams}
\bibinfo{author}{Eslami, S.}, \bibinfo{author}{de~Melo, G.},
  \bibinfo{author}{Meinel, C.}, \bibinfo{year}{2021}.
\newblock \bibinfo{title}{{TeamS} at {VQA-Med} 2021: {BBN-Orchestra} for
  long-tailed medical visual question answering}.
\newblock \bibinfo{journal}{Working Notes of CLEF} \bibinfo{volume}{201}.
\bibitem[{Fukui et~al.(2016)Fukui, Park, Yang, Rohrbach, Darrell and
  Rohrbach}]{fukui2016multimodal}
\bibinfo{author}{Fukui, A.}, \bibinfo{author}{Park, D.H.},
  \bibinfo{author}{Yang, D.}, \bibinfo{author}{Rohrbach, A.},
  \bibinfo{author}{Darrell, T.}, \bibinfo{author}{Rohrbach, M.},
  \bibinfo{year}{2016}.
\newblock \bibinfo{title}{Multimodal compact bilinear pooling for visual
  question answering and visual grounding}, in: \bibinfo{booktitle}{Proceedings
  of the 2016 Conference on Empirical Methods in Natural Language Processing},
  \bibinfo{publisher}{Association for Computational Linguistics},
  \bibinfo{address}{Austin, Texas}. pp. \bibinfo{pages}{457--468}.
\bibitem[{Gasmi et~al.(2021)Gasmi, Ltaifa, Lejeune, Alshammari, Ammar and
  Mahmood}]{Gasmi2021}
\bibinfo{author}{Gasmi, K.}, \bibinfo{author}{Ltaifa, I.B.},
  \bibinfo{author}{Lejeune, G.}, \bibinfo{author}{Alshammari, H.},
  \bibinfo{author}{Ammar, L.B.}, \bibinfo{author}{Mahmood, M.A.},
  \bibinfo{year}{2021}.
\newblock \bibinfo{title}{Optimal deep neural network-based model for answering
  visual medical question}.
\newblock \bibinfo{journal}{Cybernetics and Systems} \bibinfo{volume}{0},
  \bibinfo{pages}{1--22}.
\newblock \DOIprefix\doi{10.1080/01969722.2021.2018543}.
\bibitem[{Gong et~al.(2021a)Gong, Chen, Liu, Yu and Li}]{MTPT-CSMA}
\bibinfo{author}{Gong, H.}, \bibinfo{author}{Chen, G.}, \bibinfo{author}{Liu,
  S.}, \bibinfo{author}{Yu, Y.}, \bibinfo{author}{Li, G.},
  \bibinfo{year}{2021}a.
\newblock \bibinfo{title}{Cross-modal self-attention with multi-task
  pre-training for medical visual question answering}, in:
  \bibinfo{booktitle}{Proceedings of the 2021 International Conference on
  Multimedia Retrieval}, \bibinfo{publisher}{Association for Computing
  Machinery}, \bibinfo{address}{New York, NY, USA}. p.
  \bibinfo{pages}{456–460}.
\newblock \DOIprefix\doi{10.1145/3460426.3463584}.
\bibitem[{Gong et~al.(2021b)Gong, Huang, Chen and Li}]{gong2021sysu}
\bibinfo{author}{Gong, H.}, \bibinfo{author}{Huang, R.}, \bibinfo{author}{Chen,
  G.}, \bibinfo{author}{Li, G.}, \bibinfo{year}{2021}b.
\newblock \bibinfo{title}{{SYSU-HCP at VQA-Med} 2021: A data-centric model with
  efficient training methodology for medical visual question answering}.
\newblock \bibinfo{journal}{Working Notes of CLEF} \bibinfo{volume}{201}.
\bibitem[{Goyal et~al.(2017)Goyal, Khot, Summers{-}Stay, Batra and
  Parikh}]{balanced_vqa_v2}
\bibinfo{author}{Goyal, Y.}, \bibinfo{author}{Khot, T.},
  \bibinfo{author}{Summers{-}Stay, D.}, \bibinfo{author}{Batra, D.},
  \bibinfo{author}{Parikh, D.}, \bibinfo{year}{2017}.
\newblock \bibinfo{title}{Making the {V} in {VQA} matter: Elevating the role of
  image understanding in {V}isual {Q}uestion {A}nswering}, in:
  \bibinfo{booktitle}{Conference on Computer Vision and Pattern Recognition
  (CVPR)}.
\bibitem[{Gupta et~al.(2021)Gupta, Suman and Ekbal}]{GUPTA-HQS}
\bibinfo{author}{Gupta, D.}, \bibinfo{author}{Suman, S.},
  \bibinfo{author}{Ekbal, A.}, \bibinfo{year}{2021}.
\newblock \bibinfo{title}{Hierarchical deep multi-modal network for medical
  visual question answering}.
\newblock \bibinfo{journal}{Expert Systems with Applications}
  \bibinfo{volume}{164}, \bibinfo{pages}{113993}.
\newblock \DOIprefix\doi{https://doi.org/10.1016/j.eswa.2020.113993}.
\bibitem[{Hasan et~al.(2018)Hasan, Ling, Farri, Liu, M{\"u}ller and
  Lungren}]{hasan2018overview}
\bibinfo{author}{Hasan, S.A.}, \bibinfo{author}{Ling, Y.},
  \bibinfo{author}{Farri, O.}, \bibinfo{author}{Liu, J.},
  \bibinfo{author}{M{\"u}ller, H.}, \bibinfo{author}{Lungren, M.P.},
  \bibinfo{year}{2018}.
\newblock \bibinfo{title}{Overview of {ImageCLEF} 2018 medical domain visual
  question answering task.}, in: \bibinfo{booktitle}{CLEF (Working Notes)}.
\bibitem[{He et~al.(2016)He, Zhang, Ren and Sun}]{he2016deep}
\bibinfo{author}{He, K.}, \bibinfo{author}{Zhang, X.}, \bibinfo{author}{Ren,
  S.}, \bibinfo{author}{Sun, J.}, \bibinfo{year}{2016}.
\newblock \bibinfo{title}{Deep residual learning for image recognition}, in:
  \bibinfo{booktitle}{2016 IEEE Conference on Computer Vision and Pattern
  Recognition (CVPR)}, \bibinfo{publisher}{IEEE Computer Society},
  \bibinfo{address}{Los Alamitos, CA, USA}. pp. \bibinfo{pages}{770--778}.
\bibitem[{He et~al.(2020)He, Zhang, Mou, Xing and Xie}]{he2020pathvqa}
\bibinfo{author}{He, X.}, \bibinfo{author}{Zhang, Y.}, \bibinfo{author}{Mou,
  L.}, \bibinfo{author}{Xing, E.}, \bibinfo{author}{Xie, P.},
  \bibinfo{year}{2020}.
\newblock \bibinfo{title}{{PathVQA}: 30000+ questions for medical visual
  question answering}.
\newblock \bibinfo{journal}{arXiv preprint arXiv:2003.10286} .
\bibitem[{Hekler et~al.(2019)Hekler, Utikal, Enk, Hauschild, Weichenthal,
  Maron, Berking, Haferkamp, Klode, Schadendorf et~al.}]{hekler2019superior}
\bibinfo{author}{Hekler, A.}, \bibinfo{author}{Utikal, J.S.},
  \bibinfo{author}{Enk, A.H.}, \bibinfo{author}{Hauschild, A.},
  \bibinfo{author}{Weichenthal, M.}, \bibinfo{author}{Maron, R.C.},
  \bibinfo{author}{Berking, C.}, \bibinfo{author}{Haferkamp, S.},
  \bibinfo{author}{Klode, J.}, \bibinfo{author}{Schadendorf, D.}, et~al.,
  \bibinfo{year}{2019}.
\newblock \bibinfo{title}{Superior skin cancer classification by the
  combination of human and artificial intelligence}.
\newblock \bibinfo{journal}{European Journal of Cancer} \bibinfo{volume}{120},
  \bibinfo{pages}{114--121}.
\bibitem[{Hochreiter and Schmidhuber(1997)}]{hochreiter1997long}
\bibinfo{author}{Hochreiter, S.}, \bibinfo{author}{Schmidhuber, J.},
  \bibinfo{year}{1997}.
\newblock \bibinfo{title}{Long short-term memory}.
\newblock \bibinfo{journal}{Neural Computation} \bibinfo{volume}{9},
  \bibinfo{pages}{1735--1780}.
\bibitem[{Jiang et~al.(2020)Jiang, Chen, Yang and Zhao}]{Jiang_2020_CVPR}
\bibinfo{author}{Jiang, M.}, \bibinfo{author}{Chen, S.}, \bibinfo{author}{Yang,
  J.}, \bibinfo{author}{Zhao, Q.}, \bibinfo{year}{2020}.
\newblock \bibinfo{title}{Fantastic answers and where to find them: Immersive
  question-directed visual attention}, in: \bibinfo{booktitle}{2020 IEEE/CVF
  Conference on Computer Vision and Pattern Recognition (CVPR)},
  \bibinfo{publisher}{IEEE Computer Society}, \bibinfo{address}{Los Alamitos,
  CA, USA}. pp. \bibinfo{pages}{2977--2986}.
\bibitem[{Johnson et~al.(2019)Johnson, Pollard, Greenbaum, Lungren, Deng, Peng,
  Lu, Mark, Berkowitz and Horng}]{johnson2019mimiccxrjpg}
\bibinfo{author}{Johnson, A.E.}, \bibinfo{author}{Pollard, T.J.},
  \bibinfo{author}{Greenbaum, N.R.}, \bibinfo{author}{Lungren, M.P.},
  \bibinfo{author}{Deng, C.y.}, \bibinfo{author}{Peng, Y.},
  \bibinfo{author}{Lu, Z.}, \bibinfo{author}{Mark, R.G.},
  \bibinfo{author}{Berkowitz, S.J.}, \bibinfo{author}{Horng, S.},
  \bibinfo{year}{2019}.
\newblock \bibinfo{title}{{MIMIC-CXR-JPG}, a large publicly available database
  of labeled chest radiographs}.
\newblock \bibinfo{journal}{arXiv preprint arXiv:1901.07042} .
\bibitem[{Jung et~al.(2020)Jung, Gu and HaradaAl-Sadi}]{jung2020mvqa}
\bibinfo{author}{Jung, B.}, \bibinfo{author}{Gu, L.},
  \bibinfo{author}{HaradaAl-Sadi, T.}, \bibinfo{year}{2020}.
\newblock \bibinfo{title}{bumjun$\_$jung at {VQA-Med} 2020: {VQA} model based
  on feature extraction and multi-modal feature fusion}, in:
  \bibinfo{booktitle}{CLEF 2020 Working Notes}.
\bibitem[{K.~Verma and Ramachandran~S.(2020)}]{harendrakv2020mvqa}
\bibinfo{author}{K.~Verma, H.}, \bibinfo{author}{Ramachandran~S., S.},
  \bibinfo{year}{2020}.
\newblock \bibinfo{title}{{HARENDRAKV} at {VQA-Med} 2020: Sequential {VQA} with
  attention for medical visual question answering}, in:
  \bibinfo{booktitle}{CLEF 2020 Working Notes}.
\bibitem[{Kavur et~al.(2019)Kavur, Selver, Dicle, Barış and
  Gezer}]{CHAOSdata2019}
\bibinfo{author}{Kavur, A.E.}, \bibinfo{author}{Selver, M.A.},
  \bibinfo{author}{Dicle, O.}, \bibinfo{author}{Barış, M.},
  \bibinfo{author}{Gezer, N.S.}, \bibinfo{year}{2019}.
\newblock \bibinfo{title}{{CHAOS - Combined (CT-MR) Healthy Abdominal Organ
  Segmentation Challenge Data}}.
\newblock \DOIprefix\doi{10.5281/zenodo.3362844}.
\bibitem[{Khare et~al.(2021)Khare, Bagal, Mathew, Devi, Priyakumar and
  Jawahar}]{MMBERT}
\bibinfo{author}{Khare, Y.}, \bibinfo{author}{Bagal, V.},
  \bibinfo{author}{Mathew, M.}, \bibinfo{author}{Devi, A.},
  \bibinfo{author}{Priyakumar, U.D.}, \bibinfo{author}{Jawahar, C.},
  \bibinfo{year}{2021}.
\newblock \bibinfo{title}{Mmbert: Multimodal bert pretraining for improved
  medical vqa}, in: \bibinfo{booktitle}{2021 IEEE 18th International Symposium
  on Biomedical Imaging (ISBI)}, pp. \bibinfo{pages}{1033--1036}.
\newblock \DOIprefix\doi{10.1109/ISBI48211.2021.9434063}.
\bibitem[{Kim et~al.(2018)Kim, Jun and Zhang}]{kim2018bilinear}
\bibinfo{author}{Kim, J.}, \bibinfo{author}{Jun, J.}, \bibinfo{author}{Zhang,
  B.}, \bibinfo{year}{2018}.
\newblock \bibinfo{title}{Bilinear attention networks}, in:
  \bibinfo{editor}{Bengio, S.}, \bibinfo{editor}{Wallach, H.M.},
  \bibinfo{editor}{Larochelle, H.}, \bibinfo{editor}{Grauman, K.},
  \bibinfo{editor}{Cesa{-}Bianchi, N.}, \bibinfo{editor}{Garnett, R.} (Eds.),
  \bibinfo{booktitle}{Advances in Neural Information Processing Systems},
  \bibinfo{address}{Montr{\'{e}}al, Canada}. pp. \bibinfo{pages}{1571--1581}.
\bibitem[{Kornuta et~al.(2019)Kornuta, Rajan, Shivade, Asseman and
  Ozcan}]{kornuta2019leveraging}
\bibinfo{author}{Kornuta, T.}, \bibinfo{author}{Rajan, D.},
  \bibinfo{author}{Shivade, C.}, \bibinfo{author}{Asseman, A.},
  \bibinfo{author}{Ozcan, A.S.}, \bibinfo{year}{2019}.
\newblock \bibinfo{title}{Leveraging medical visual question answering with
  supporting facts}, in: \bibinfo{booktitle}{CLEF (Working Notes)}.
\bibitem[{Kovaleva et~al.(2020)Kovaleva, Shivade, Kashyap, Kanjaria, Wu,
  Ballah, Coy, Karargyris, Guo, Beymer et~al.}]{kovaleva2020towards}
\bibinfo{author}{Kovaleva, O.}, \bibinfo{author}{Shivade, C.},
  \bibinfo{author}{Kashyap, S.}, \bibinfo{author}{Kanjaria, K.},
  \bibinfo{author}{Wu, J.}, \bibinfo{author}{Ballah, D.}, \bibinfo{author}{Coy,
  A.}, \bibinfo{author}{Karargyris, A.}, \bibinfo{author}{Guo, Y.},
  \bibinfo{author}{Beymer, D.B.}, et~al., \bibinfo{year}{2020}.
\newblock \bibinfo{title}{Towards visual dialog for radiology}, in:
  \bibinfo{booktitle}{Proceedings of the 19th SIGBioMed Workshop on Biomedical
  Language Processing}, pp. \bibinfo{pages}{60--69}.
\bibitem[{Krishna et~al.(2017)Krishna, Zhu, Groth, Johnson, Hata, Kravitz,
  Chen, Kalantidis, Li, Shamma et~al.}]{krishna2017visual}
\bibinfo{author}{Krishna, R.}, \bibinfo{author}{Zhu, Y.},
  \bibinfo{author}{Groth, O.}, \bibinfo{author}{Johnson, J.},
  \bibinfo{author}{Hata, K.}, \bibinfo{author}{Kravitz, J.},
  \bibinfo{author}{Chen, S.}, \bibinfo{author}{Kalantidis, Y.},
  \bibinfo{author}{Li, L.J.}, \bibinfo{author}{Shamma, D.A.}, et~al.,
  \bibinfo{year}{2017}.
\newblock \bibinfo{title}{Visual genome: Connecting language and vision using
  crowdsourced dense image annotations}.
\newblock \bibinfo{journal}{International journal of computer vision}
  \bibinfo{volume}{123}, \bibinfo{pages}{32--73}.
\bibitem[{Lau et~al.(2018)Lau, Gayen, Abacha and
  Demner-Fushman}]{lau2018dataset}
\bibinfo{author}{Lau, J.J.}, \bibinfo{author}{Gayen, S.},
  \bibinfo{author}{Abacha, A.B.}, \bibinfo{author}{Demner-Fushman, D.},
  \bibinfo{year}{2018}.
\newblock \bibinfo{title}{A dataset of clinically generated visual questions
  and answers about radiology images}.
\newblock \bibinfo{journal}{Scientific Data} \bibinfo{volume}{5},
  \bibinfo{pages}{1--10}.
\bibitem[{Lee et~al.(2020)Lee, Yoon, Kim, Kim, Kim, So and
  Kang}]{lee2020biobert}
\bibinfo{author}{Lee, J.}, \bibinfo{author}{Yoon, W.}, \bibinfo{author}{Kim,
  S.}, \bibinfo{author}{Kim, D.}, \bibinfo{author}{Kim, S.},
  \bibinfo{author}{So, C.H.}, \bibinfo{author}{Kang, J.}, \bibinfo{year}{2020}.
\newblock \bibinfo{title}{{BioBERT}: a pre-trained biomedical language
  representation model for biomedical text mining}.
\newblock \bibinfo{journal}{Bioinformatics} \bibinfo{volume}{36},
  \bibinfo{pages}{1234--1240}.
\bibitem[{Li and Liu(2021)}]{li2021lijie}
\bibinfo{author}{Li, J.}, \bibinfo{author}{Liu, S.}, \bibinfo{year}{2021}.
\newblock \bibinfo{title}{Lijie at {ImageCLEFmed} {VQA-Med} 2021: Attention
  model based on efficient interaction between multimodality}.
\newblock \bibinfo{journal}{Working Notes of CLEF} \bibinfo{volume}{201}.
\bibitem[{Li et~al.(2021a)Li, Cai, Liu, Weng, Zhao, Wang, Chen, Liu, Pan, Li
  et~al.}]{li2021ffa}
\bibinfo{author}{Li, M.}, \bibinfo{author}{Cai, W.}, \bibinfo{author}{Liu, R.},
  \bibinfo{author}{Weng, Y.}, \bibinfo{author}{Zhao, X.},
  \bibinfo{author}{Wang, C.}, \bibinfo{author}{Chen, X.}, \bibinfo{author}{Liu,
  Z.}, \bibinfo{author}{Pan, C.}, \bibinfo{author}{Li, M.}, et~al.,
  \bibinfo{year}{2021}a.
\newblock \bibinfo{title}{{FFA-IR}: Towards an explainable and reliable medical
  report generation benchmark}, in: \bibinfo{booktitle}{Thirty-fifth Conference
  on Neural Information Processing Systems Datasets and Benchmarks Track (Round
  2)}.
\bibitem[{Li et~al.(2021b)Li, Yang and Hao}]{li2021tam}
\bibinfo{author}{Li, Y.}, \bibinfo{author}{Yang, Z.}, \bibinfo{author}{Hao,
  T.}, \bibinfo{year}{2021}b.
\newblock \bibinfo{title}{{TAM} at {VQA-Med} 2021: A hybrid model with feature
  extraction and fusion for medical visual question answering}.
\newblock \bibinfo{journal}{Working Notes of CLEF} \bibinfo{volume}{201}.
\bibitem[{Liao et~al.(2020)Liao, Wu, Shen, van~den Hengel and
  Verjans}]{aiml2020mvqa}
\bibinfo{author}{Liao, Z.}, \bibinfo{author}{Wu, Q.}, \bibinfo{author}{Shen,
  C.}, \bibinfo{author}{van~den Hengel, A.}, \bibinfo{author}{Verjans, J.},
  \bibinfo{year}{2020}.
\newblock \bibinfo{title}{{AIML} at {VQA-Med} 2020: Knowledge inference via a
  skeleton-based sentence mapping approach for medical domain visual question
  answering}, in: \bibinfo{booktitle}{CLEF 2020 Working Notes}.
\bibitem[{Lin et~al.(2013)Lin, Chen and Yan}]{lin2013network}
\bibinfo{author}{Lin, M.}, \bibinfo{author}{Chen, Q.}, \bibinfo{author}{Yan,
  S.}, \bibinfo{year}{2013}.
\newblock \bibinfo{title}{Network in network}.
\newblock \bibinfo{journal}{arXiv preprint arXiv:1312.4400} .
\bibitem[{Lin et~al.(2014)Lin, Maire, Belongie, Hays, Perona, Ramanan,
  Doll{\'a}r and Zitnick}]{lin2014microsoft}
\bibinfo{author}{Lin, T.Y.}, \bibinfo{author}{Maire, M.},
  \bibinfo{author}{Belongie, S.}, \bibinfo{author}{Hays, J.},
  \bibinfo{author}{Perona, P.}, \bibinfo{author}{Ramanan, D.},
  \bibinfo{author}{Doll{\'a}r, P.}, \bibinfo{author}{Zitnick, C.L.},
  \bibinfo{year}{2014}.
\newblock \bibinfo{title}{Microsoft {COCO}: Common objects in context}, in:
  \bibinfo{booktitle}{European conference on computer vision},
  \bibinfo{organization}{Springer}. pp. \bibinfo{pages}{740--755}.
\bibitem[{Liu et~al.(2021a)Liu, Zhan and Wu}]{CPRD-MICCAI21}
\bibinfo{author}{Liu, B.}, \bibinfo{author}{Zhan, L.M.}, \bibinfo{author}{Wu,
  X.M.}, \bibinfo{year}{2021}a.
\newblock \bibinfo{title}{Contrastive pre-training and representation
  distillation for medical visual question answering based on radiology
  images}, in: \bibinfo{editor}{de~Bruijne, M.}, \bibinfo{editor}{Cattin,
  P.C.}, \bibinfo{editor}{Cotin, S.}, \bibinfo{editor}{Padoy, N.},
  \bibinfo{editor}{Speidel, S.}, \bibinfo{editor}{Zheng, Y.},
  \bibinfo{editor}{Essert, C.} (Eds.), \bibinfo{booktitle}{Medical Image
  Computing and Computer Assisted Intervention -- MICCAI 2021},
  \bibinfo{publisher}{Springer International Publishing},
  \bibinfo{address}{Cham}. pp. \bibinfo{pages}{210--220}.
\bibitem[{Liu et~al.(2021b)Liu, Zhan, Xu, Ma, Yang and Wu}]{liu2021slake}
\bibinfo{author}{Liu, B.}, \bibinfo{author}{Zhan, L.M.}, \bibinfo{author}{Xu,
  L.}, \bibinfo{author}{Ma, L.}, \bibinfo{author}{Yang, Y.},
  \bibinfo{author}{Wu, X.M.}, \bibinfo{year}{2021}b.
\newblock \bibinfo{title}{Slake: A semantically-labeled knowledge-enhanced
  dataset for medical visual question answering}.
\newblock \href{http://arxiv.org/abs/2102.09542}{\tt arXiv:2102.09542}.
\bibitem[{Liu et~al.(2020)Liu, Ding and Zhou}]{shengyan2020mvqa}
\bibinfo{author}{Liu, S.}, \bibinfo{author}{Ding, H.}, \bibinfo{author}{Zhou,
  X.}, \bibinfo{year}{2020}.
\newblock \bibinfo{title}{Shengyan at {VQA-Med} 2020: An encoder-decoder model
  for medical domain visual question answering task}, in:
  \bibinfo{booktitle}{CLEF 2020 Working Notes}.
\bibitem[{Liu et~al.(2019)Liu, Ou, Che, Zhou and Ding}]{liu2019xception}
\bibinfo{author}{Liu, S.}, \bibinfo{author}{Ou, X.}, \bibinfo{author}{Che, J.},
  \bibinfo{author}{Zhou, X.}, \bibinfo{author}{Ding, H.}, \bibinfo{year}{2019}.
\newblock \bibinfo{title}{An {Xception-GRU} model for visual question answering
  in the medical domain.}, in: \bibinfo{booktitle}{CLEF (Working Notes)}.
\bibitem[{Lu et~al.(2016)Lu, Yang, Batra and Parikh}]{lu2016hierarchical}
\bibinfo{author}{Lu, J.}, \bibinfo{author}{Yang, J.}, \bibinfo{author}{Batra,
  D.}, \bibinfo{author}{Parikh, D.}, \bibinfo{year}{2016}.
\newblock \bibinfo{title}{Hierarchical question-image co-attention for visual
  question answering}, in: \bibinfo{booktitle}{Advances in Neural Information
  Processing Systems}, pp. \bibinfo{pages}{289--297}.
\bibitem[{Marino et~al.(2019)Marino, Rastegari, Farhadi and
  Mottaghi}]{marino2019ok}
\bibinfo{author}{Marino, K.}, \bibinfo{author}{Rastegari, M.},
  \bibinfo{author}{Farhadi, A.}, \bibinfo{author}{Mottaghi, R.},
  \bibinfo{year}{2019}.
\newblock \bibinfo{title}{{OK-VQA}: A visual question answering benchmark
  requiring external knowledge}, in: \bibinfo{booktitle}{2019 IEEE/CVF
  Conference on Computer Vision and Pattern Recognition (CVPR)},
  \bibinfo{publisher}{IEEE Computer Society}, \bibinfo{address}{Los Alamitos,
  CA, USA}. pp. \bibinfo{pages}{3190--3199}.
\bibitem[{McDonald et~al.(2015)McDonald, Schwartz, Eckel, Diehn, Hunt,
  Bartholmai, Erickson and Kallmes}]{mcdonald2015effects}
\bibinfo{author}{McDonald, R.J.}, \bibinfo{author}{Schwartz, K.M.},
  \bibinfo{author}{Eckel, L.J.}, \bibinfo{author}{Diehn, F.E.},
  \bibinfo{author}{Hunt, C.H.}, \bibinfo{author}{Bartholmai, B.J.},
  \bibinfo{author}{Erickson, B.J.}, \bibinfo{author}{Kallmes, D.F.},
  \bibinfo{year}{2015}.
\newblock \bibinfo{title}{The effects of changes in utilization and
  technological advancements of cross-sectional imaging on radiologist
  workload}.
\newblock \bibinfo{journal}{Academic radiology} \bibinfo{volume}{22},
  \bibinfo{pages}{1191--1198}.
\bibitem[{M{\"u}ller et~al.(2019)M{\"u}ller, Gangadharaiah, Klein, Perry,
  Bernstein, Nurkse, Wailes, Graham, El-Kareh, Mehta et~al.}]{muller2019open}
\bibinfo{author}{M{\"u}ller, L.}, \bibinfo{author}{Gangadharaiah, R.},
  \bibinfo{author}{Klein, S.C.}, \bibinfo{author}{Perry, J.},
  \bibinfo{author}{Bernstein, G.}, \bibinfo{author}{Nurkse, D.},
  \bibinfo{author}{Wailes, D.}, \bibinfo{author}{Graham, R.},
  \bibinfo{author}{El-Kareh, R.}, \bibinfo{author}{Mehta, S.}, et~al.,
  \bibinfo{year}{2019}.
\newblock \bibinfo{title}{An open access medical knowledge base for community
  driven diagnostic decision support system development}.
\newblock \bibinfo{journal}{BMC medical informatics and decision making}
  \bibinfo{volume}{19}, \bibinfo{pages}{93}.
\bibitem[{Nguyen et~al.(2019)Nguyen, Do, Nguyen, Do, Tjiputra and
  Tran}]{nguyen2019overcoming}
\bibinfo{author}{Nguyen, B.D.}, \bibinfo{author}{Do, T.T.},
  \bibinfo{author}{Nguyen, B.X.}, \bibinfo{author}{Do, T.},
  \bibinfo{author}{Tjiputra, E.}, \bibinfo{author}{Tran, Q.D.},
  \bibinfo{year}{2019}.
\newblock \bibinfo{title}{Overcoming data limitation in medical visual question
  answering}, in: \bibinfo{editor}{Shen, D.}, \bibinfo{editor}{Liu, T.},
  \bibinfo{editor}{Peters, T.M.}, \bibinfo{editor}{Staib, L.H.},
  \bibinfo{editor}{Essert, C.}, \bibinfo{editor}{Zhou, S.},
  \bibinfo{editor}{Yap, P.T.}, \bibinfo{editor}{Khan, A.} (Eds.),
  \bibinfo{booktitle}{Medical Image Computing and Computer Assisted
  Intervention -- MICCAI 2019}, \bibinfo{publisher}{Springer International
  Publishing}, \bibinfo{address}{Cham}. pp. \bibinfo{pages}{522--530}.
\bibitem[{Nori et~al.(2023)Nori, King, McKinney, Carignan and
  Horvitz}]{nori2023capabilities}
\bibinfo{author}{Nori, H.}, \bibinfo{author}{King, N.},
  \bibinfo{author}{McKinney, S.M.}, \bibinfo{author}{Carignan, D.},
  \bibinfo{author}{Horvitz, E.}, \bibinfo{year}{2023}.
\newblock \bibinfo{title}{Capabilities of gpt-4 on medical challenge problems}.
\newblock \bibinfo{journal}{arXiv preprint arXiv:2303.13375} .
\bibitem[{Oikarinen et~al.()Oikarinen, Das, Nguyen and Weng}]{oikarinenlabel}
\bibinfo{author}{Oikarinen, T.}, \bibinfo{author}{Das, S.},
  \bibinfo{author}{Nguyen, L.M.}, \bibinfo{author}{Weng, T.W.}, .
\newblock \bibinfo{title}{Label-free concept bottleneck models}, in:
  \bibinfo{booktitle}{International Conference on Learning Representations}.
\bibitem[{Papineni et~al.(2002)Papineni, Roukos, Ward and
  Zhu}]{papineni2002bleu}
\bibinfo{author}{Papineni, K.}, \bibinfo{author}{Roukos, S.},
  \bibinfo{author}{Ward, T.}, \bibinfo{author}{Zhu, W.J.},
  \bibinfo{year}{2002}.
\newblock \bibinfo{title}{{BLEU}: a method for automatic evaluation of machine
  translation}, in: \bibinfo{booktitle}{Proceedings of the 40th Annual Meeting
  of the Association for Computational Linguistics},
  \bibinfo{publisher}{Association for Computational Linguistics},
  \bibinfo{address}{Philadelphia, Pennsylvania, USA}. pp.
  \bibinfo{pages}{311--318}.
\bibitem[{Park et~al.(2018)Park, Hendricks, Akata, Rohrbach, Schiele, Darrell
  and Rohrbach}]{huk2018multimodal}
\bibinfo{author}{Park, D.}, \bibinfo{author}{Hendricks, L.},
  \bibinfo{author}{Akata, Z.}, \bibinfo{author}{Rohrbach, A.},
  \bibinfo{author}{Schiele, B.}, \bibinfo{author}{Darrell, T.},
  \bibinfo{author}{Rohrbach, M.}, \bibinfo{year}{2018}.
\newblock \bibinfo{title}{Multimodal explanations: Justifying decisions and
  pointing to the evidence}, in: \bibinfo{booktitle}{2018 IEEE/CVF Conference
  on Computer Vision and Pattern Recognition (CVPR)}, \bibinfo{publisher}{IEEE
  Computer Society}, \bibinfo{address}{Los Alamitos, CA, USA}. pp.
  \bibinfo{pages}{8779--8788}.
\bibitem[{Pelka et~al.(2018)Pelka, Koitka, R{\"u}ckert, Nensa and
  Friedrich}]{pelka2018roco}
\bibinfo{author}{Pelka, O.}, \bibinfo{author}{Koitka, S.},
  \bibinfo{author}{R{\"u}ckert, J.}, \bibinfo{author}{Nensa, F.},
  \bibinfo{author}{Friedrich, C.M.}, \bibinfo{year}{2018}.
\newblock \bibinfo{title}{Radiology objects in {COntext (ROCO)}: a multimodal
  image dataset}, in: \bibinfo{booktitle}{Intravascular Imaging and Computer
  Assisted Stenting and Large-Scale Annotation of Biomedical Data and Expert
  Label Synthesis}. \bibinfo{publisher}{Springer}, pp.
  \bibinfo{pages}{180--189}.
\bibitem[{Peng et~al.(2018)Peng, Liu and Rosen}]{peng2018umass}
\bibinfo{author}{Peng, Y.}, \bibinfo{author}{Liu, F.}, \bibinfo{author}{Rosen,
  M.P.}, \bibinfo{year}{2018}.
\newblock \bibinfo{title}{{UMass} at {ImageCLEF} medical visual question
  answering ({Med-VQA}) 2018 task}, in: \bibinfo{booktitle}{CLEF (Working
  Notes)}.
\bibitem[{Radford et~al.(2018)Radford, Narasimhan, Salimans, Sutskever
  et~al.}]{radford2018improving}
\bibinfo{author}{Radford, A.}, \bibinfo{author}{Narasimhan, K.},
  \bibinfo{author}{Salimans, T.}, \bibinfo{author}{Sutskever, I.}, et~al.,
  \bibinfo{year}{2018}.
\newblock \bibinfo{title}{Improving language understanding by generative
  pre-training} .
\bibitem[{Ramakrishnan et~al.(2018)Ramakrishnan, Agrawal and
  Lee}]{ramakrishnan2018overcoming}
\bibinfo{author}{Ramakrishnan, S.}, \bibinfo{author}{Agrawal, A.},
  \bibinfo{author}{Lee, S.}, \bibinfo{year}{2018}.
\newblock \bibinfo{title}{Overcoming language priors in visual question
  answering with adversarial regularization}, in: \bibinfo{booktitle}{Advances
  in Neural Information Processing Systems}, pp. \bibinfo{pages}{1541--1551}.
\bibitem[{Ren and Zhou(2020)}]{ren2020cgmvqa}
\bibinfo{author}{Ren, F.}, \bibinfo{author}{Zhou, Y.}, \bibinfo{year}{2020}.
\newblock \bibinfo{title}{{CGMVQA}: A new classification and generative model
  for medical visual question answering}.
\newblock \bibinfo{journal}{IEEE Access} \bibinfo{volume}{8},
  \bibinfo{pages}{50626--50636}.
\bibitem[{Russakovsky et~al.(2015)Russakovsky, Deng, Su, Krause, Satheesh, Ma,
  Huang, Karpathy, Khosla, Bernstein et~al.}]{ILSVRC15}
\bibinfo{author}{Russakovsky, O.}, \bibinfo{author}{Deng, J.},
  \bibinfo{author}{Su, H.}, \bibinfo{author}{Krause, J.},
  \bibinfo{author}{Satheesh, S.}, \bibinfo{author}{Ma, S.},
  \bibinfo{author}{Huang, Z.}, \bibinfo{author}{Karpathy, A.},
  \bibinfo{author}{Khosla, A.}, \bibinfo{author}{Bernstein, M.}, et~al.,
  \bibinfo{year}{2015}.
\newblock \bibinfo{title}{{ImageNet} large scale visual recognition challenge}.
\newblock \bibinfo{journal}{International Journal of Computer Vision}
  \bibinfo{volume}{115}, \bibinfo{pages}{211--252}.
\bibitem[{Sarrouti(2020)}]{nlm2020mvqa}
\bibinfo{author}{Sarrouti, M.}, \bibinfo{year}{2020}.
\newblock \bibinfo{title}{{NLM} at {VQA-Med} 2020: Visual question answering
  and generation in the medical domain}, in: \bibinfo{booktitle}{CLEF 2020
  Working Notes}.
\bibitem[{Schilling et~al.(2021)Schilling, Messina, Parra and
  Lobel}]{schilling2021puc}
\bibinfo{author}{Schilling, R.}, \bibinfo{author}{Messina, P.},
  \bibinfo{author}{Parra, D.}, \bibinfo{author}{Lobel, H.},
  \bibinfo{year}{2021}.
\newblock \bibinfo{title}{{PUC} chile team at {VQA-Med} 2021: approaching vqa
  as a classfication task via fine-tuning a pretrained cnn}.
\newblock \bibinfo{journal}{Working Notes of CLEF} \bibinfo{volume}{201}.
\bibitem[{Schuster and Paliwal(1997)}]{schuster1997bidirectional}
\bibinfo{author}{Schuster, M.}, \bibinfo{author}{Paliwal, K.K.},
  \bibinfo{year}{1997}.
\newblock \bibinfo{title}{Bidirectional recurrent neural networks}.
\newblock \bibinfo{journal}{IEEE transactions on Signal Processing}
  \bibinfo{volume}{45}, \bibinfo{pages}{2673--2681}.
\bibitem[{Shao et~al.(2023)Shao, Yu, Wang and Yu}]{shao2023prompting}
\bibinfo{author}{Shao, Z.}, \bibinfo{author}{Yu, Z.}, \bibinfo{author}{Wang,
  M.}, \bibinfo{author}{Yu, J.}, \bibinfo{year}{2023}.
\newblock \bibinfo{title}{Prompting large language models with answer
  heuristics for knowledge-based visual question answering}.
\newblock \bibinfo{journal}{arXiv preprint arXiv:2303.01903} .
\bibitem[{Sharma et~al.(2021)Sharma, Purushotham and
  Reddy}]{sharma2021medfusenet}
\bibinfo{author}{Sharma, D.}, \bibinfo{author}{Purushotham, S.},
  \bibinfo{author}{Reddy, C.K.}, \bibinfo{year}{2021}.
\newblock \bibinfo{title}{Medfusenet: An attention-based multimodal deep
  learning model for visual question answering in the medical domain}.
\newblock \bibinfo{journal}{Scientific Reports} \bibinfo{volume}{11},
  \bibinfo{pages}{1--18}.
\bibitem[{Shi et~al.(2019)Shi, Liu and Rosen}]{shi2019deep}
\bibinfo{author}{Shi, L.}, \bibinfo{author}{Liu, F.}, \bibinfo{author}{Rosen,
  M.P.}, \bibinfo{year}{2019}.
\newblock \bibinfo{title}{Deep multimodal learning for medical visual question
  answering.}, in: \bibinfo{booktitle}{CLEF (Working Notes)}.
\bibitem[{Shickel et~al.(2018)Shickel, Tighe, Bihorac and Rashidi}]{deepehr}
\bibinfo{author}{Shickel, B.}, \bibinfo{author}{Tighe, P.J.},
  \bibinfo{author}{Bihorac, A.}, \bibinfo{author}{Rashidi, P.},
  \bibinfo{year}{2018}.
\newblock \bibinfo{title}{Deep {EHR}: A survey of recent advances in deep
  learning techniques for electronic health record ({EHR}) analysis}.
\newblock \bibinfo{journal}{IEEE Journal of Biomedical and Health Informatics}
  \bibinfo{volume}{22}, \bibinfo{pages}{1589--1604}.
\newblock \DOIprefix\doi{10.1109/JBHI.2017.2767063}.
\bibitem[{Simonyan and Zisserman(2015)}]{simonyan2014deep}
\bibinfo{author}{Simonyan, K.}, \bibinfo{author}{Zisserman, A.},
  \bibinfo{year}{2015}.
\newblock \bibinfo{title}{Very deep convolutional networks for large-scale
  image recognition}, in: \bibinfo{booktitle}{Proceedings of the 3rd
  International Conference on Learning Representations}.
\bibitem[{Simpson et~al.(2019)Simpson, Antonelli, Bakas, Bilello, Farahani, van
  Ginneken, Kopp-Schneider, Landman, Litjens, Menze, Ronneberger, Summers,
  Bilic, Christ, Do, Gollub, Golia-Pernicka, Heckers, Jarnagin, McHugo, Napel,
  Vorontsov, Maier-Hein and Cardoso}]{simpson2019large}
\bibinfo{author}{Simpson, A.L.}, \bibinfo{author}{Antonelli, M.},
  \bibinfo{author}{Bakas, S.}, \bibinfo{author}{Bilello, M.},
  \bibinfo{author}{Farahani, K.}, \bibinfo{author}{van Ginneken, B.},
  \bibinfo{author}{Kopp-Schneider, A.}, \bibinfo{author}{Landman, B.A.},
  \bibinfo{author}{Litjens, G.}, \bibinfo{author}{Menze, B.},
  \bibinfo{author}{Ronneberger, O.}, \bibinfo{author}{Summers, R.M.},
  \bibinfo{author}{Bilic, P.}, \bibinfo{author}{Christ, P.F.},
  \bibinfo{author}{Do, R.K.G.}, \bibinfo{author}{Gollub, M.},
  \bibinfo{author}{Golia-Pernicka, J.}, \bibinfo{author}{Heckers, S.H.},
  \bibinfo{author}{Jarnagin, W.R.}, \bibinfo{author}{McHugo, M.K.},
  \bibinfo{author}{Napel, S.}, \bibinfo{author}{Vorontsov, E.},
  \bibinfo{author}{Maier-Hein, L.}, \bibinfo{author}{Cardoso, M.J.},
  \bibinfo{year}{2019}.
\newblock \bibinfo{title}{A large annotated medical image dataset for the
  development and evaluation of segmentation algorithms}.
\newblock \href{http://arxiv.org/abs/1902.09063}{\tt arXiv:1902.09063}.
\bibitem[{Sitara and Kavitha(2021)}]{sitarassn}
\bibinfo{author}{Sitara, N.M.S.}, \bibinfo{author}{Kavitha, S.},
  \bibinfo{year}{2021}.
\newblock \bibinfo{title}{{SSN MLRG} at {VQA-Med} 2021: An approach for {VQA}
  to solve abnormality related queries using improved datasets}.
\newblock \bibinfo{journal}{Working Notes of CLEF} \bibinfo{volume}{201}.
\bibitem[{Talafha and Al-Ayyoub(2018)}]{talafha2018just}
\bibinfo{author}{Talafha, B.}, \bibinfo{author}{Al-Ayyoub, M.},
  \bibinfo{year}{2018}.
\newblock \bibinfo{title}{{JUST} at {VQA-Med}: A {VGG-Seq2Seq} model}, in:
  \bibinfo{booktitle}{CLEF (Working Notes)}.
\bibitem[{Thanki and Makkithaya(2019)}]{thanki2019manipal}
\bibinfo{author}{Thanki, A.}, \bibinfo{author}{Makkithaya, K.},
  \bibinfo{year}{2019}.
\newblock \bibinfo{title}{{MIT Manipal} at {ImageCLEF} 2019 visual question
  answering in medical domain}, in: \bibinfo{booktitle}{CLEF (Working Notes)}.
\bibitem[{Thomee et~al.(2016)Thomee, Shamma, Friedland, Elizalde, Ni, Poland,
  Borth and Li}]{thomee2016yfcc100m}
\bibinfo{author}{Thomee, B.}, \bibinfo{author}{Shamma, D.A.},
  \bibinfo{author}{Friedland, G.}, \bibinfo{author}{Elizalde, B.},
  \bibinfo{author}{Ni, K.}, \bibinfo{author}{Poland, D.},
  \bibinfo{author}{Borth, D.}, \bibinfo{author}{Li, L.J.},
  \bibinfo{year}{2016}.
\newblock \bibinfo{title}{{YFCC100M}: The new data in multimedia research}.
\newblock \bibinfo{journal}{Communications of the ACM} \bibinfo{volume}{59},
  \bibinfo{pages}{64--73}.
\bibitem[{Tschandl et~al.(2020)Tschandl, Rinner, Apalla, Argenziano, Codella,
  Halpern, Janda, Lallas, Longo, Malvehy et~al.}]{tschandl2020human}
\bibinfo{author}{Tschandl, P.}, \bibinfo{author}{Rinner, C.},
  \bibinfo{author}{Apalla, Z.}, \bibinfo{author}{Argenziano, G.},
  \bibinfo{author}{Codella, N.}, \bibinfo{author}{Halpern, A.},
  \bibinfo{author}{Janda, M.}, \bibinfo{author}{Lallas, A.},
  \bibinfo{author}{Longo, C.}, \bibinfo{author}{Malvehy, J.}, et~al.,
  \bibinfo{year}{2020}.
\newblock \bibinfo{title}{Human-computer collaboration for skin cancer
  recognition}.
\newblock \bibinfo{journal}{Nature Medicine} \bibinfo{volume}{26},
  \bibinfo{pages}{1229--1234}.
\bibitem[{Tschandl et~al.(2018)Tschandl, Rosendahl and Kittler}]{DBW86T_2018}
\bibinfo{author}{Tschandl, P.}, \bibinfo{author}{Rosendahl, C.},
  \bibinfo{author}{Kittler, H.}, \bibinfo{year}{2018}.
\newblock \bibinfo{title}{The {HAM}10000 dataset, a large collection of
  multi-source dermatoscopic images of common pigmented skin lesions}.
\newblock \bibinfo{journal}{Scientific Data} \bibinfo{volume}{5}.
\newblock \DOIprefix\doi{10.1038/sdata.2018.161}.
\bibitem[{Turner and Spanier(2019)}]{turner2019lstm}
\bibinfo{author}{Turner, A.}, \bibinfo{author}{Spanier, A.},
  \bibinfo{year}{2019}.
\newblock \bibinfo{title}{{LSTM} in {VQA-Med}, is it really needed? {JCE} study
  on the {ImageCLEF} 2019 dataset}, in: \bibinfo{booktitle}{CLEF (Working
  Notes)}.
\bibitem[{Umada and Aono(2020)}]{kdevqa2020mvqa}
\bibinfo{author}{Umada, H.}, \bibinfo{author}{Aono, M.}, \bibinfo{year}{2020}.
\newblock \bibinfo{title}{kdevqa at {VQA-Med} 2020: focusing on {GLU}-based
  classification}, in: \bibinfo{booktitle}{CLEF 2020 Working Notes}.
\bibitem[{Vaswani et~al.(2017)Vaswani, Shazeer, Parmar, Uszkoreit, Jones,
  Gomez, Kaiser and Polosukhin}]{vaswani2017attention}
\bibinfo{author}{Vaswani, A.}, \bibinfo{author}{Shazeer, N.},
  \bibinfo{author}{Parmar, N.}, \bibinfo{author}{Uszkoreit, J.},
  \bibinfo{author}{Jones, L.}, \bibinfo{author}{Gomez, A.N.},
  \bibinfo{author}{Kaiser, {\L}.}, \bibinfo{author}{Polosukhin, I.},
  \bibinfo{year}{2017}.
\newblock \bibinfo{title}{Attention is all you need}, in:
  \bibinfo{booktitle}{Advances in Neural Information Processing Systems}, pp.
  \bibinfo{pages}{5998--6008}.
\bibitem[{Vu et~al.(2019)Vu, Sznitman, Nyholm and
  L{\"o}fstedt}]{vu2019ensemble}
\bibinfo{author}{Vu, M.}, \bibinfo{author}{Sznitman, R.},
  \bibinfo{author}{Nyholm, T.}, \bibinfo{author}{L{\"o}fstedt, T.},
  \bibinfo{year}{2019}.
\newblock \bibinfo{title}{Ensemble of streamlined bilinear visual question
  answering models for the {ImageCLEF} 2019 challenge in the medical domain},
  in: \bibinfo{booktitle}{CLEF 2019}.
\bibitem[{Vu et~al.(2020)Vu, Löfstedt, Nyholm and Sznitman}]{Minh-QCMLB}
\bibinfo{author}{Vu, M.H.}, \bibinfo{author}{Löfstedt, T.},
  \bibinfo{author}{Nyholm, T.}, \bibinfo{author}{Sznitman, R.},
  \bibinfo{year}{2020}.
\newblock \bibinfo{title}{A question-centric model for visual question
  answering in medical imaging}.
\newblock \bibinfo{journal}{IEEE Transactions on Medical Imaging}
  \bibinfo{volume}{39}, \bibinfo{pages}{2856--2868}.
\newblock \DOIprefix\doi{10.1109/TMI.2020.2978284}.
\bibitem[{Wang et~al.(2017a)Wang, Wu, Shen, Dick and van~den
  Hengel}]{wang2015explicit}
\bibinfo{author}{Wang, P.}, \bibinfo{author}{Wu, Q.}, \bibinfo{author}{Shen,
  C.}, \bibinfo{author}{Dick, A.}, \bibinfo{author}{van~den Hengel, A.},
  \bibinfo{year}{2017}a.
\newblock \bibinfo{title}{Explicit knowledge-based reasoning for visual
  question answering}, in: \bibinfo{booktitle}{Proceedings of the Twenty-Sixth
  International Joint Conference on Artificial Intelligence, {IJCAI-17}}, pp.
  \bibinfo{pages}{1290--1296}.
\bibitem[{Wang et~al.(2018)Wang, Wu, Shen, Dick and van~den
  Hengel}]{wang2018fvqa}
\bibinfo{author}{Wang, P.}, \bibinfo{author}{Wu, Q.}, \bibinfo{author}{Shen,
  C.}, \bibinfo{author}{Dick, A.}, \bibinfo{author}{van~den Hengel, A.},
  \bibinfo{year}{2018}.
\newblock \bibinfo{title}{{FVQA}: Fact-based visual question answering}.
\newblock \bibinfo{journal}{IEEE Transactions on Pattern Analysis \& Machine
  Intelligence} \bibinfo{volume}{40}, \bibinfo{pages}{2413--2427}.
\bibitem[{Wang et~al.(2023)Wang, Zhao, Ouyang, Wang and Shen}]{wang2023chatcad}
\bibinfo{author}{Wang, S.}, \bibinfo{author}{Zhao, Z.},
  \bibinfo{author}{Ouyang, X.}, \bibinfo{author}{Wang, Q.},
  \bibinfo{author}{Shen, D.}, \bibinfo{year}{2023}.
\newblock \bibinfo{title}{Chat{CAD}: Interactive computer-aided diagnosis on
  medical image using large language models}.
\newblock \bibinfo{journal}{arXiv preprint arXiv:2302.07257} .
\bibitem[{Wang et~al.(2020)Wang, Liu, Shen, Ng, Luo, Jin, Chan, van~den Hengel
  and Wang}]{Wang_2020_CVPR}
\bibinfo{author}{Wang, X.}, \bibinfo{author}{Liu, Y.}, \bibinfo{author}{Shen,
  C.}, \bibinfo{author}{Ng, C.}, \bibinfo{author}{Luo, C.},
  \bibinfo{author}{Jin, L.}, \bibinfo{author}{Chan, C.},
  \bibinfo{author}{van~den Hengel, A.}, \bibinfo{author}{Wang, L.},
  \bibinfo{year}{2020}.
\newblock \bibinfo{title}{On the general value of evidence, and bilingual
  scene-text visual question answering}, in: \bibinfo{booktitle}{2020 IEEE/CVF
  Conference on Computer Vision and Pattern Recognition (CVPR)},
  \bibinfo{publisher}{IEEE Computer Society}, \bibinfo{address}{Los Alamitos,
  CA, USA}. pp. \bibinfo{pages}{10123--10132}.
\bibitem[{Wang et~al.(2017b)Wang, Peng, Lu, Lu, Bagheri and
  Summers}]{ChestXRay8}
\bibinfo{author}{Wang, X.}, \bibinfo{author}{Peng, Y.}, \bibinfo{author}{Lu,
  L.}, \bibinfo{author}{Lu, Z.}, \bibinfo{author}{Bagheri, M.},
  \bibinfo{author}{Summers, R.M.}, \bibinfo{year}{2017}b.
\newblock \bibinfo{title}{{ChestX-Ray8}: Hospital-scale chest {X-Ray} database
  and benchmarks on weakly-supervised classification and localization of common
  thorax diseases}, in: \bibinfo{booktitle}{2017 IEEE Conference on Computer
  Vision and Pattern Recognition (CVPR)}, \bibinfo{publisher}{IEEE Computer
  Society}, \bibinfo{address}{Los Alamitos, CA, USA}. pp.
  \bibinfo{pages}{3462--3471}.
\newblock \DOIprefix\doi{10.1109/CVPR.2017.369}.
\bibitem[{Wu et~al.(2017)Wu, Teney, Wang, Shen, Dick and {van den
  Hengel}}]{wu2017visual}
\bibinfo{author}{Wu, Q.}, \bibinfo{author}{Teney, D.}, \bibinfo{author}{Wang,
  P.}, \bibinfo{author}{Shen, C.}, \bibinfo{author}{Dick, A.},
  \bibinfo{author}{{van den Hengel}, A.}, \bibinfo{year}{2017}.
\newblock \bibinfo{title}{Visual question answering: A survey of methods and
  datasets}.
\newblock \bibinfo{journal}{Computer Vision and Image Understanding}
  \bibinfo{volume}{163}, \bibinfo{pages}{21--40}.
\newblock \bibinfo{note}{Language in Vision}.
\bibitem[{Xiao et~al.(2021)Xiao, Zhou, Xiao and Zhao}]{xiao2021yunnan}
\bibinfo{author}{Xiao, Q.}, \bibinfo{author}{Zhou, X.}, \bibinfo{author}{Xiao,
  Y.}, \bibinfo{author}{Zhao, K.}, \bibinfo{year}{2021}.
\newblock \bibinfo{title}{Yunnan university at {VQA-Med} 2021: Pretrained
  {BioBERT} for medical domain visual question answering}.
\newblock \bibinfo{journal}{Working Notes of CLEF} \bibinfo{volume}{201}.
\bibitem[{Yan et~al.(2019)Yan, Li, Xie, Xiao and Gu}]{yan2019zhejiang}
\bibinfo{author}{Yan, X.}, \bibinfo{author}{Li, L.}, \bibinfo{author}{Xie, C.},
  \bibinfo{author}{Xiao, J.}, \bibinfo{author}{Gu, L.}, \bibinfo{year}{2019}.
\newblock \bibinfo{title}{{Zhejiang University} at {ImageCLEF} 2019 visual
  question answering in the medical domain}, in: \bibinfo{booktitle}{CLEF
  (Working Notes)}.
\bibitem[{Yang et~al.(2022)Yang, Panagopoulou, Zhou, Jin, Callison-Burch and
  Yatskar}]{yang2022language}
\bibinfo{author}{Yang, Y.}, \bibinfo{author}{Panagopoulou, A.},
  \bibinfo{author}{Zhou, S.}, \bibinfo{author}{Jin, D.},
  \bibinfo{author}{Callison-Burch, C.}, \bibinfo{author}{Yatskar, M.},
  \bibinfo{year}{2022}.
\newblock \bibinfo{title}{Language in a bottle: Language model guided concept
  bottlenecks for interpretable image classification}.
\newblock \bibinfo{journal}{arXiv preprint arXiv:2211.11158} .
\bibitem[{Yang et~al.(2016)Yang, He, Gao, Deng and Smola}]{yang2015stacked}
\bibinfo{author}{Yang, Z.}, \bibinfo{author}{He, X.}, \bibinfo{author}{Gao,
  J.}, \bibinfo{author}{Deng, L.}, \bibinfo{author}{Smola, A.},
  \bibinfo{year}{2016}.
\newblock \bibinfo{title}{Stacked attention networks for image question
  answering}, in: \bibinfo{booktitle}{2016 IEEE Conference on Computer Vision
  and Pattern Recognition (CVPR)}, \bibinfo{publisher}{IEEE Computer Society},
  \bibinfo{address}{Los Alamitos, CA, USA}. pp. \bibinfo{pages}{21--29}.
\bibitem[{Yu et~al.(2019)Yu, Yu, Cui, Tao and Tian}]{yu2019deep}
\bibinfo{author}{Yu, Z.}, \bibinfo{author}{Yu, J.}, \bibinfo{author}{Cui, Y.},
  \bibinfo{author}{Tao, D.}, \bibinfo{author}{Tian, Q.}, \bibinfo{year}{2019}.
\newblock \bibinfo{title}{Deep modular co-attention networks for visual
  question answering}, in: \bibinfo{booktitle}{2019 IEEE/CVF Conference on
  Computer Vision and Pattern Recognition (CVPR)}, \bibinfo{publisher}{IEEE
  Computer Society}, \bibinfo{address}{Los Alamitos, CA, USA}. pp.
  \bibinfo{pages}{6274--6283}.
\bibitem[{Yu et~al.(2017)Yu, Yu, Fan and Tao}]{yu2017multi}
\bibinfo{author}{Yu, Z.}, \bibinfo{author}{Yu, J.}, \bibinfo{author}{Fan, J.},
  \bibinfo{author}{Tao, D.}, \bibinfo{year}{2017}.
\newblock \bibinfo{title}{Multi-modal factorized bilinear pooling with
  co-attention learning for visual question answering}, in:
  \bibinfo{booktitle}{2017 IEEE International Conference on Computer Vision
  (ICCV)}, \bibinfo{publisher}{IEEE Computer Society}, \bibinfo{address}{Los
  Alamitos, CA, USA}. pp. \bibinfo{pages}{1839--1848}.
\bibitem[{Yu et~al.(2018)Yu, Yu, Xiang, Fan and Tao}]{yu2018beyond}
\bibinfo{author}{Yu, Z.}, \bibinfo{author}{Yu, J.}, \bibinfo{author}{Xiang,
  C.}, \bibinfo{author}{Fan, J.}, \bibinfo{author}{Tao, D.},
  \bibinfo{year}{2018}.
\newblock \bibinfo{title}{Beyond bilinear: Generalized multimodal factorized
  high-order pooling for visual question answering}.
\newblock \bibinfo{journal}{IEEE Transactions on Neural Networks and Learning
  Systems} \bibinfo{volume}{29}, \bibinfo{pages}{5947--5959}.
\newblock \DOIprefix\doi{10.1109/TNNLS.2018.2817340}.
\bibitem[{Zhan et~al.(2020)Zhan, Liu, Fan, Chen and Wu}]{zhan2020medical}
\bibinfo{author}{Zhan, L.M.}, \bibinfo{author}{Liu, B.}, \bibinfo{author}{Fan,
  L.}, \bibinfo{author}{Chen, J.}, \bibinfo{author}{Wu, X.M.},
  \bibinfo{year}{2020}.
\newblock \bibinfo{title}{Medical visual question answering via conditional
  reasoning}, in: \bibinfo{booktitle}{Proceedings of the 28th ACM International
  Conference on Multimedia (MM ’20)}, \bibinfo{organization}{ACM}.
\bibitem[{Zheng et~al.(2020)Zheng, Yan, Wang and Gou}]{zheng2020KEML}
\bibinfo{author}{Zheng, W.}, \bibinfo{author}{Yan, L.}, \bibinfo{author}{Wang,
  F.Y.}, \bibinfo{author}{Gou, C.}, \bibinfo{year}{2020}.
\newblock \bibinfo{title}{Learning from the guidance: Knowledge embedded
  meta-learning for medical visual question answering}, in:
  \bibinfo{editor}{Yang, H.}, \bibinfo{editor}{Pasupa, K.},
  \bibinfo{editor}{Leung, A.C.S.}, \bibinfo{editor}{Kwok, J.T.},
  \bibinfo{editor}{Chan, J.H.}, \bibinfo{editor}{King, I.} (Eds.),
  \bibinfo{booktitle}{Neural Information Processing},
  \bibinfo{publisher}{Springer International Publishing},
  \bibinfo{address}{Cham}. pp. \bibinfo{pages}{194--202}.
\bibitem[{Zhou et~al.(2020)Zhou, Cui, Wei and Chen}]{zhou2020BBN}
\bibinfo{author}{Zhou, B.}, \bibinfo{author}{Cui, Q.}, \bibinfo{author}{Wei,
  X.S.}, \bibinfo{author}{Chen, Z.M.}, \bibinfo{year}{2020}.
\newblock \bibinfo{title}{{BBN}: Bilateral-branch network with cumulative
  learning for long-tailed visual recognition} , \bibinfo{pages}{1--8}.
\bibitem[{Zhou et~al.(2018)Zhou, Kang and Ren}]{zhou2018employing}
\bibinfo{author}{Zhou, Y.}, \bibinfo{author}{Kang, X.}, \bibinfo{author}{Ren,
  F.}, \bibinfo{year}{2018}.
\newblock \bibinfo{title}{Employing {Inception-Resnet-v2} and {Bi-LSTM} for
  medical domain visual question answering}, in: \bibinfo{booktitle}{CLEF
  (Working Notes)}.
\bibitem[{Zhou et~al.(2019)Zhou, Kang and Ren}]{zhou2019tua1}
\bibinfo{author}{Zhou, Y.}, \bibinfo{author}{Kang, X.}, \bibinfo{author}{Ren,
  F.}, \bibinfo{year}{2019}.
\newblock \bibinfo{title}{{TUA1} at {ImageCLEF} 2019 vqa-med: a classification
  and generation model based on transfer learning}, in:
  \bibinfo{booktitle}{CLEF (Working Notes)}.

\end{thebibliography}
\end{document}